%% file: main.tex
\documentclass{article}

    \PassOptionsToPackage{numbers, compress}{natbib}

\usepackage[main, final]{neurips_2026}

\usepackage[utf8]{inputenc} 
\usepackage[T1]{fontenc}    
\usepackage{hyperref}       
\usepackage{url}            
\usepackage{booktabs}       
\usepackage{amsfonts}       
\usepackage{nicefrac}       
\usepackage{microtype}      
\usepackage{enumitem}
\usepackage{bm}
\usepackage{graphicx}
\usepackage{subcaption}
\usepackage{caption}
\usepackage{amsmath}
\usepackage{amssymb}
\usepackage{mathtools}
\usepackage{amsthm}
\usepackage{multirow}
\usepackage[table]{xcolor}
\usepackage{booktabs}
\usepackage{wrapfig}
\definecolor{vanillacolor}{RGB}{245,245,245} 
\definecolor{isgecolor}{RGB}{240,255,240} 
\usepackage[capitalize,noabbrev]{cleveref}

\definecolor{isgecolor1}{HTML}{4C78A8}
\definecolor{vanillacolor1}{HTML}{F28E2B}
\definecolor{oraclecolor1}{HTML}{D95F5F}
\theoremstyle{plain}
\newtheorem{theorem}{Theorem}[section]
\newtheorem{proposition}[theorem]{Proposition}
\newtheorem{lemma}[theorem]{Lemma}

\theoremstyle{definition}
\newtheorem{definition}[theorem]{Definition}
\newtheorem{assumption}[theorem]{Assumption}
\theoremstyle{remark}

\usepackage[textsize=tiny]{todonotes}

\definecolor{StratBlue}{RGB}{225,235,245}
\definecolor{EnvGreen}{RGB}{230,245,235}

\title{Iterative Scarcity-Guided Exploration: Bootstrapping Generative Auto-bidding from Narrow Support}

\author{%
  \textbf{Qingmao Yao$^{1}$, Guangzheng Hu$^{2}$, Yusen Huo$^{4}$\thanks{This work was completed while the authors were at Alibaba.}, Zhilin Zhang$^{4}$\footnotemark[1]} \\
  \textbf{Chuan Yu$^{2}$, Jian Xu$^{2}$, Xu Chu$^{1}$\thanks{Corresponding authors}, Bo Zheng$^{2}$\footnotemark[2], Xiaotie Deng$^{1,3}$\footnotemark[2]} \\
  \normalfont $^{1}$Center on Frontiers of Computing Studies, School of Computer Science, Peking University\\
  \normalfont $^{2}$Alimama Tech, Taobao \& Tmall Group of Alibaba\\
  \normalfont $^{3}$Department of Computer Science, City University of Hong Kong\\
  \normalfont $^{4}$Independent Researcher\\
  \normalfont \texttt{qmyao25@stu.pku.edu.cn}, \texttt{huguangzheng.hgz@taobao.com}, \texttt{huoyusen1@126.com}\\
  \normalfont \texttt{421525432@qq.com}, \texttt{\{yuchuan.yc,xiyu.xj\}@taobao.com}\\
  \normalfont \texttt{bozheng@alibaba-inc.com}, \texttt{\{chu\_xu,xiaotie\}@pku.edu.cn}\\
}

\begin{document}

\maketitle

\begin{abstract}
Auto-bidding is a core algorithmic component in online advertising auctions.
In practical cold-start settings, iterative training is commonly used to compensate for low-quality, narrowly supported historical data.
Unfortunately, in iterative training, representative diffusion-based AI-Generated Bidding (AIGB) methods fail to sustain extrapolation beyond the narrow data support and stagnate at a suboptimal level.
In this paper, we theoretically attribute this stagnation to Signal-to-Noise Ratio (SNR) collapse: the weak radial return signal is overwhelmed by a curvature-induced penalty.
To break this stagnation, we propose \textbf{Iterative Scarcity-Guided Exploration (ISGE)}, which introduces a guidance handoff: as the return signal collapses, scarcity subsequently takes over as an exploratory signal to elevate the SNR above a critical threshold.
Specifically, ISGE iteratively alternates between a \emph{Judger} that assigns scarcity scores to trajectories and an \emph{Explorer} that performs guided generation of high-scarcity and high-return trajectories, thereby bootstrapping from the narrow support with such self-generated trajectories.
Extensive experiments on the industrial AuctionNet benchmark demonstrate that ISGE, starting from cold-start datasets, effectively surpasses the performance of the Full-Dataset baseline within three iterations.
Code is available at \url{https://github.com/alimama-tech/ISGE}.
\end{abstract}

\section{Introduction}
\label{sec:intro}
Auto-bidding in online advertising is a complex sequential decision-making problem~\cite{evans2009online,10.1145/2623330.2623633,USCB}.
Recently, diffusion-based AIGB methods have emerged as a powerful paradigm by formulating bidding as trajectory generation via conditional diffusion planners~\cite{aigb2024,peng2025expert,cbd2025}.
However, these models heavily rely on broad-coverage expert data. 
In practical cold-start settings for new advertisers or emerging business
sectors, historical data is often generated by conservative strategies, yielding a narrow, low-quality distribution. 
Bootstrapping beyond this poor initialization requires iterative training, yet existing diffusion planners lack efficient mechanisms to explore high-return out-of-distribution (OOD) trajectories. This raises a central question:

\emph{Can diffusion-based AIGB methods perform efficient exploration to bootstrap a near-optimal policy solely from low-quality, narrowly supported data in just a few iterations?}

\begin{figure}[t]
    \centering
    \begin{subfigure}[b]{0.32\textwidth}
        \centering
        \includegraphics[width=\textwidth]{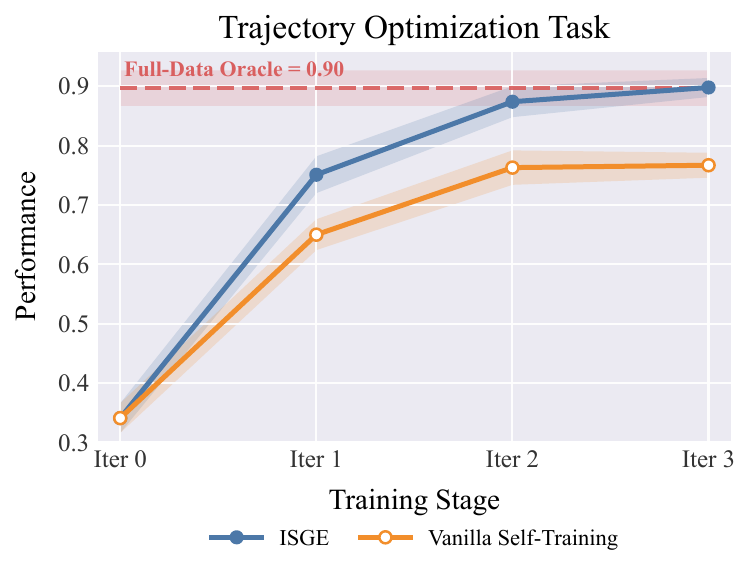}
    \end{subfigure}
    \hfill
    \begin{subfigure}[b]{0.32\textwidth}
        \centering
        \includegraphics[width=\textwidth]{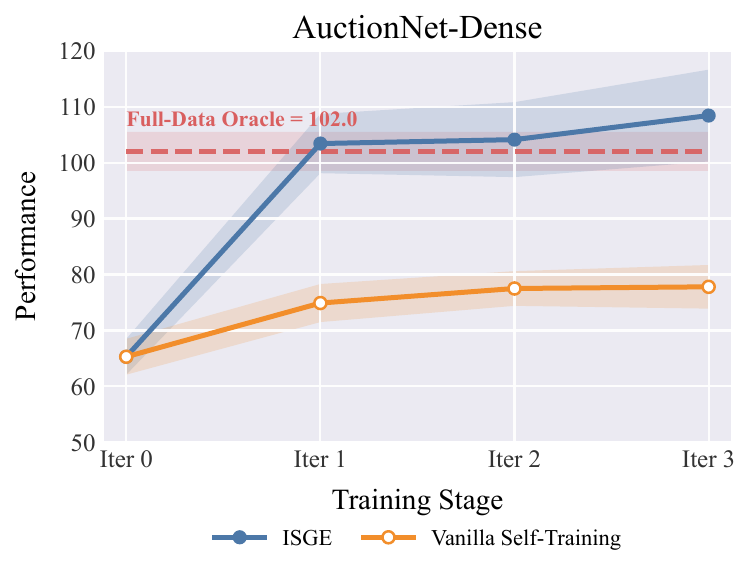}
    \end{subfigure}
    \hfill
    \begin{subfigure}[b]{0.32\textwidth}
        \centering
        \includegraphics[width=\textwidth]{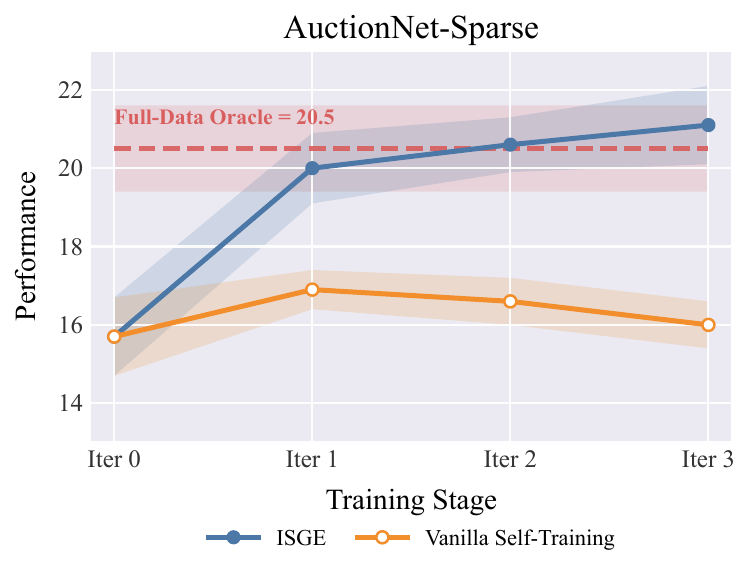}
    \end{subfigure}

    \caption{Iterative Experiments on Trajectory Optimization Task (left) and AuctionNet Benchmarks (middle, right). \textcolor{vanillacolor1}{Vanilla Self-Training (orange)} inevitably stagnates well below the optimal level during iterative training. We term this self-locking. In contrast, our proposed \textcolor{isgecolor1}{ISGE (blue)} successfully breaks self-locking and consistently reaches or surpasses the \textcolor{oraclecolor1}{Full-Data Oracle (red dashed line)} within just three iterations. Full results in \cref{subsec:motivating_example} and \cref{sec:experiments}.}
    \label{Fig:iteration_experiments}
    \vspace{-5pt}
\end{figure}

Investigating this, we observe that vanilla self-training (naively iterating diffusion planners) suffers from a bottleneck we term \emph{self-locking}. As illustrated in \cref{Fig:iteration_experiments}, despite initial extrapolation capabilities, performance inevitably stagnates well below the optimum. We theoretically attribute this to an inherent Signal-to-Noise Ratio (SNR) collapse at the boundary of the training support. As generated trajectories reach this suboptimal frontier, the return signal becomes sparse and misaligned, failing to overcome the high-dimensional curvature penalty and trapping the planner.

To break self-locking, we propose Iterative Scarcity-Guided Exploration (ISGE). ISGE relies on the core insight of a \emph{guidance handoff}, utilizing scarcity (quantified via denoising autoencoder~\cite{10.1145/1390156.1390294} reconstruction error) as a robust auxiliary signal. Geometrically, scarcity acts as an outward manifold repulsion field~\cite{alain2014regularized}. During reverse denoising, ISGE executes a guidance handoff: at early steps, return guidance navigates the known landscape to push trajectories to the suboptimal boundary. When the return signal inevitably collapses, scarcity subsequently takes over. This provides a stable outward impulse that reliably pushes the trajectory deeper into unexplored OOD space, explicitly preventing regression into the low-quality interior. By leveraging the local continuity of the return landscape, this handoff forms a highly effective exploration heuristic.

Practically, ISGE alternates between a Judger scoring trajectory scarcity and an Explorer using dual-condition guidance to generate high-return, high-scarcity ``gift samples''. Extensive experiments on the industrial AuctionNet~\cite{auctionnet2024} benchmark demonstrate that ISGE effectively bridges the data quality gap, surpassing the Full-Dataset baseline \textit{within just three iterations} from narrowly supported cold-start data.
To summarize, our main contributions are threefold:
\begin{enumerate}[leftmargin=2em]
\item We theoretically attribute return-guided self-locking to SNR collapse at the suboptimal boundary, and establish a guidance handoff mechanism where scarcity subsequently takes over to elevate the SNR and break the bottleneck.
\item We propose ISGE, an iterative framework that leverages return and scarcity guidance to generate ``gift samples'' beyond the training support.
\item  We demonstrate that ISGE bootstraps near-optimal policies from cold-start datasets, surpassing the Full-Dataset baseline on the industrial AuctionNet benchmark within three iterations.
\end{enumerate}



\section{Preliminaries}
\label{sec:preliminaries}

\paragraph{Generative Auto-Bidding}
\label{subsec:gen_bidding}
Auto-bidding is a sequential decision problem with long horizons, sparse rewards, and non-Markovian dynamics.
Following standard settings (details in Appendix~\ref{app:detailed_formulation}), an agent allocates a global budget to maximize total return while satisfying strict KPI constraints over $T$ steps.
Diffusion-based AIGB~\cite{aigb2024} formulates this as conditional trajectory generation $p(\tau | y(\tau))$, where $\tau = (s_0, \dots, s_T)$ is the state trajectory and $y(\tau)$ is the target return (KPI constraints are typically absorbed into the return via reward shaping). This conditional distribution acts as the \emph{policy}, implicitly determining a sequence of actions via an inverse dynamics model $a_t = \phi(s_t, s_{t+1})$. The planner effectively satisfies bidding objectives by generating state trajectories $\tau$ consistent with the specified condition $y$.


\paragraph{Conditional Diffusion Models and Classifier-Free Guidance}
\label{subsec:diffusion_cfg}

Conditional diffusion models transform decision-making to conditional trajectory generation.
Letting \(\bm{x}_0 = \tau\) denote the clean trajectory, the forward process gradually corrupts it into pure noise \(\bm{x}_T \sim \mathcal{N}(0, \mathbf{I})\).
The reverse process learns to iteratively denoise \(\bm{x}_t\) by predicting the added noise \(\epsilon_\theta(\bm{x}_t, y, t)\), conditioned on both the guidance signal \(y\) (e.g., target return) and the timestep \(t\).
Classifier-Free Guidance (CFG)~\cite{ho2022classifier,ajay2022conditional} avoids training an explicit classifier, making it lightweight for decision-making.
CFG jointly trains a conditional model $\epsilon_\theta(\bm{x}_t, y, t)$ and an unconditional counterpart $\epsilon_\theta(\bm{x}_t, \emptyset, t)$ via random condition dropout. During sampling, the perturbed noise is given by:
$\hat{\epsilon} = \epsilon_\theta(\bm{x}_t, \emptyset, t) + w \cdot \left( \epsilon_\theta(\bm{x}_t, y, t) - \epsilon_\theta(\bm{x}_t, \emptyset, t) \right)$, 
where $w$ is the guidance scale.

\section{Motivation and Theoretical Analysis}
\label{sec:theory}
\vspace{-3pt}
In this section, we first introduce a motivating example that empirically reveals the self-locking bottleneck. We then establish a geometric optimization framework to demonstrate how an inherent SNR collapse causes self-locking, and how scarcity breaks self-locking via guidance handoff.

\vspace{-3pt}
\subsection{Motivating Example: Self-Locking in Trajectory Optimization}
\label{subsec:motivating_example}

Generative auto-bidding can be viewed as a high-dimensional trajectory optimization problem. However, in real-world auction environments, the optimal bidding strategy is unknown and the return landscape is highly complex, strictly limiting rigorous theoretical analysis. 
To isolate the underlying optimization dynamics, we construct a tractable high-dimensional ($D=768$) trajectory optimization task: $R(\bm{x}) = e^{-\beta \|\bm{x} - \bm{x}^*\|_2}$ ($\beta>0$), where the global optimum $\bm{x}^*$ lies strictly outside the narrow training support. 
As illustrated in \cref{Fig:iteration_experiments} (left), Vanilla Self-Training exhibits a distinct dual-phase dynamic. While it initially extrapolates beyond the cold-start initialization, performance abruptly stagnates at a suboptimal boundary by Iteration 2. 
Crucially, a Full‑Data Oracle with an identical architecture readily approaches the optimum, demonstrating that this stagnation is not a model capacity limitation but a data bottleneck.
We term this ubiquitous failure mode \emph{self-locking}, motivating our systematic theoretical analysis.
\vspace{-2pt}
\subsection{How Does the Diffusion Planner Optimize Trajectories?}
\label{subsec:initial_improvement}

To uncover the origins of self-locking, we cast the CFG-guided reverse denoising process as formal trajectory optimization. By leveraging the score-matching property, CFG effectively injects an implicit likelihood gradient $\nabla_{\bm{x}_t} \log p(y|\bm{x}_t)$ into the standard diffusion update. This reformulates the sampling step into a biased Langevin dynamic (see Appendix~\ref{proof:cfg_update} for detailed derivations):
\begin{equation}
\label{eq:sampling_update}
    \bm{x}_{t-1} = \mu_\theta(\bm{x}_t) + \underbrace{\lambda_t w \cdot \nabla_{\bm{x}_t} \log p(y|\bm{x}_t)}_{\text{Guidance Drift}} + \underbrace{\sigma_t \mathbf{z}}_{\text{Noise}},
\end{equation}
where $\mu_\theta(\bm{x}_t)$ is the unconditional mean that pulls the trajectory back toward the known data manifold, $\mathbf{z} \sim \mathcal{N}(0, \mathbf{I})$ provides isotropic sampling stochasticity, and the schedule-dependent coefficient $\lambda_t = (1-\alpha_t)/\sqrt{\alpha_t}$ acts as the gradient step size.
Equation~\eqref{eq:sampling_update} explicitly isolates the two interacting forces driving the planner's initial extrapolation: the isotropic noise ($\mathbf{z}$) provides the necessary stochastic capacity to venture beyond the narrow training support, while the guidance drift steers the optimization toward high-return regions.

\vspace{-2pt}
\subsection{Why Does Return-Guided Optimization Self-Lock?}
\label{subsec:self_locking}

As iteratively generated trajectories approach the suboptimal boundary of the training support, the delicate interplay between guidance drift and isotropic noise breaks down. We characterize this failure as an inherent SNR collapse.
To formalize this, consider the flattened trajectory vector $\bm{x} \in \mathbb{R}^D$ with $D \gg 1$, and adopt an analytically tractable spherical return landscape: $R(\bm{x}) = h(d(\bm{x}))$, where $d(\bm{x}) = \|\bm{x} - \bm{x}^*\|_2$ is the Euclidean distance to the OOD global optimum $\bm{x}^*$, and $h$ is a decreasing function (as instantiated by $h(d) = e^{-\beta d}$ in the motivating example).
This abstraction yields two crucial analytical advantages: (i) the optimal improvement direction $\mathbf{n}^* = (\bm{x}^* - \bm{x}) / d(\bm{x})$ is explicitly defined as the inward radial vector, and (ii) the scalar distance $d(\bm{x})$ serves as a direct proxy for optimization progress.
As detailed in Appendix~\ref{app:geometry_generalization}, this spherical landscape establishes a \emph{conservative lower bound} for optimization difficulty in general smooth landscapes. 
Under this setting, we now demonstrate the fundamental SNR barrier.

\begin{theorem}[SNR Barrier in High-Dimensional Optimization]
\label{thm:geometric_barrier}
Consider a trajectory $\bm{x}$ at distance $d = \|\bm{x} - \bm{x}^*\|_2$ from the optimum $\bm{x}^*$, and an update $\Delta \bm{x} = \mathbf{v}_{\mathrm{det}} + \mathbf{v}_{\mathrm{noise}}$, where $\mathbf{v}_{\mathrm{det}}$ is a deterministic drift and $\mathbf{v}_{\mathrm{noise}} \sim \mathcal{N}(0, \sigma^2 \mathbf{I})$ is isotropic noise. Let $v_{\mathrm{rad}} = \langle \mathbf{v}_{\mathrm{det}}, \mathbf{n}^* \rangle$ be the radial component of the drift.
Then, the expected change in the squared distance satisfies the exact algebraic equality:
\begin{equation}
    \mathbb{E}[\Delta (d^2)] = \underbrace{-2d v_{\mathrm{rad}}}_{\text{Signal}} + \underbrace{\|\mathbf{v}_{\mathrm{det}}\|^2 + D \sigma^2}_{\text{Penalty}}.
\end{equation}

Consequently, we define the geometric Signal-to-Noise Ratio (SNR) as the ratio of the radial signal to the total displacement penalty:
\begin{equation}
    \text{SNR} = \frac{2d v_{\mathrm{rad}}}{\|\mathbf{v}_{\mathrm{det}}\|^2 + D\sigma^2}.
\end{equation}
Distance reduction ($\mathbb{E}[\Delta (d^2)] < 0$) is strictly achievable if and only if $\text{SNR} > 1$.
\end{theorem}

Theorem~\ref{thm:geometric_barrier} (proof in Appendix~\ref{proof:geometric_barrier}) reveals a fundamental trade-off: optimization succeeds only when the radial signal strongly dominates the curvature-induced penalty ($\text{SNR} > 1$). 
\emph{Since the signal magnitude $v_{\mathrm{rad}}$ typically does not scale with dimension $D$, the $D$-dependent noise penalty effortlessly overwhelms the signal as $d$ decreases.}
The following proposition formally translates this geometric barrier to the diffusion planner, demonstrating why return-guided optimization inevitably self-locks in OOD regions.

\begin{proposition}[Return Signal Collapse]
\label{prop:bottleneck}
Consider a trajectory $\bm{x}$ whose unconditional mean $\mu_\theta(\bm{x})$ is at a distance $d$ from the OOD optimum $\bm{x}^*$. Let $\Delta \bm{x}_{R} = \lambda w_R \mathbf{g}_R + \sigma \mathbf{z}$ denote the return-guided step. 
Assuming the guidance scale $w_R$ is chosen optimally to maximize the reduction of squared distance, the return-guided optimization stagnates ($\mathbb{E}[\Delta (d^2)]_R \geq 0$) when the radial signal falls below the threshold:
\begin{equation}
    \langle \mathbf{g}_R, \mathbf{n}^* \rangle \leq \frac{\sigma \sqrt{D} \, \|\mathbf{g}_R\|}{d} \quad (\text{i.e., } \text{SNR} \leq 1).
\end{equation}
Consequently, there exists a minimal achievable radius $d_{\min} > 0$ such that further reduction of $d$ is statistically impossible under a vanishing radial signal outside the suboptimal training support.
\end{proposition}


Proposition~\ref{prop:bottleneck} (proof in Appendix~\ref{proof:bottleneck}) characterizes the mechanism of self-locking. As the planner extrapolates beyond the known training manifold, the implicit return gradient $\mathbf{g}_R$ becomes increasingly sparse and misaligned. Once its radial component $\langle \mathbf{g}_R, \mathbf{n}^* \rangle$ decays below the threshold required to combat the high-dimensional displacement penalty, the expected geometric progress halts. The diffusion planner is physically locked at the suboptimal boundary, incapable of sustaining further OOD exploration.


\subsection{Breaking Self-Locking via Scarcity Guidance Handoff}
\label{subsec:scarcity_works}
To break this self-locking, the diffusion planner requires a guidance \textit{handoff}: a transition from the collapsed return signal to a robust auxiliary signal that persists in the OOD vacuum. Crucially, since the training data is confined to a suboptimal region, advancing toward the OOD optimum inherently requires a stable outward impulse moving beyond the known data manifold. We identify scarcity as the ideal geometric signal to execute this handoff.

\begin{definition}[Scarcity via Optimal DAE]
\label{def:scarcity}
Consider a denoising autoencoder (DAE) mapping $r: \mathbb{R}^D \rightarrow \mathbb{R}^D$ trained to minimize the expected squared reconstruction error under isotropic Gaussian corruption $\bm{\epsilon} \sim \mathcal{N}(0, \eta^2 \mathbf{I})$. Let $r_\eta^*(\cdot)$ denote the theoretically optimal reconstruction function. We define the \textit{scarcity score} $S(\bm{x})$ of any state $\bm{x}$ as its expected reconstruction error:
\begin{equation}
    S(\bm{x}) := \mathbb{E}_{\bm{\epsilon}} \left[ \| \bm{x} - r_\eta^*(\bm{x} + \bm{\epsilon}) \|^2 \right].
\end{equation}
\end{definition}

Intuitively, a DAE denoises a sample by pulling it back toward the high-density data manifold (a direction mathematically aligned with the score function $\nabla \log p_{\mathrm{data}}$~\cite{alain2014regularized}). 
In this sense, $S(\bm{x})$ naturally acts as a proxy for the geometric distance to the training data: it remains small within the data distribution and strictly increases as the trajectory ventures outward.
Therefore, its gradient $\nabla S(\bm{x})$ inherently points away from the known data, providing the outward repulsion field needed to escape the suboptimal boundary. Formally, we present the following properties.

\begin{proposition}[Properties of the Scarcity Gradient]
\label{prop:scarcity_field}
Let $\mathcal{M}$ be the suboptimal training manifold. Following the theory of optimal DAEs, the explicit scarcity gradient $\mathbf{v}_S(\bm{x}) := \nabla_{\bm{x}} S(\bm{x})$ satisfies two critical geometric properties in the local OOD neighborhood $\mathcal{N}(\mathcal{M}) \setminus \mathcal{M}$:
\begin{enumerate}
    \item \textbf{Normal Alignment:} $\mathbf{v}_S(\bm{x})$ is consistently aligned with the outward-pointing normal vector $\mathbf{n}_{\perp}$ of the manifold $\mathcal{M}$.
    \item \textbf{Non-Vanishing Magnitude:} There exists a constant $\delta > 0$ such that $\|\mathbf{v}_S(\bm{x})\| \geq \delta$, ensuring the signal does not collapse at the boundary.
\end{enumerate}
Furthermore, by integrating this signal via CFG, the implicit scarcity gradient $\mathbf{g}_S(\bm{x}) := \nabla_{\bm{x}} \log p(y_S | \bm{x})$ preserves these properties, acting as a non-vanishing outward impulse.
\end{proposition}

Proposition~\ref{prop:scarcity_field} (proof in Appendix~\ref{proof:scarcity_field}) formally justifies that the implicit scarcity gradient $\mathbf{g}_S$ reliably provides a non-vanishing, outward-pointing signal. This naturally raises the question: \emph{does an outward impulse facilitate optimization?} 
While moving outward is not a universal guarantee of improvement in arbitrary landscapes, it serves as a geometrically sound heuristic under the self-locking condition. 
At this self-locked frontier, all improvement directions within the known suboptimal support are exhausted.
The cone of improvement directions towards the OOD optimum $\bm{x}^*$ necessarily extends outward beyond the frontier.
Consequently, the outward-pointing scarcity gradient $\mathbf{g}_S$ exhibits a high likelihood of aligning with this cone, naturally yielding a positive inner product ($\langle \mathbf{g}_S, \mathbf{n}^* \rangle > 0$). 
We validate this positive alignment via dedicated sanity checks in Section~\ref{subsec:sanity_check}.
Assuming this alignment holds, we now demonstrate how $\mathbf{g}_S$ takes over the optimization process.

\begin{theorem}[Optimization Handoff via Scarcity]\label{thm:scarcity_revival}
Consider the trajectory $\bm{x}$ from Proposition~\ref{prop:bottleneck} at the critical self-locking boundary, where $\mathbb{E}[\Delta (d^2)]_R = 0$. 
Let $\Delta \bm{x}_{\text{dual}} = \lambda (w_R \mathbf{g}_R + w_S \mathbf{g}_S) + \sigma \mathbf{z}$ denote the dual-guided exploration step.
Assuming the scarcity gradient aligns positively with the optimal direction ($\langle \mathbf{g}_S, \mathbf{n}^* \rangle > 0$, with full assumptions detailed in Appendix \ref{proof:scarcity_revival}), the optimization handoff is successfully executed ($\mathbb{E}[\Delta (d^2)]_{\text{dual}} < 0$) if the scarcity guidance scale $w_S$ satisfies: 
\begin{equation}
    w_S \in \left( 0, \frac{2d \langle \mathbf{g}_S, \mathbf{n}^* \rangle}{\lambda \|\mathbf{g}_S\|^2} \right).
\end{equation}
Furthermore, by choosing the scarcity scale $w_S^* = {d \langle \mathbf{g}_S, \mathbf{n}^* \rangle}/{\lambda \|\mathbf{g}_S\|^2}$, the marginal SNR of the scarcity guidance is exactly 2 ($\text{SNR}_{\Delta}(w_S^*)=2$), maximizing the lower bound on optimization progress.
\end{theorem}


Theorem~\ref{thm:scarcity_revival} (proof in Appendix~\ref{proof:scarcity_revival}) formally guarantees that by injecting scarcity guidance, the planner breaks self-locking, uncovers trajectories with both high return and high scarcity. We denote such trajectories as \hyperref[def:gift_samples]{\emph{gift samples}}.
Appending these gift samples to the training dataset permanently restores the missing radial component of the return gradient (Appendix~\ref{proof:gradient_restoration}), directly motivating the iterative framework.

\vspace{-5pt}
\paragraph{Discussion.} 
Our theoretical framework relies on two primary abstractions. 
(i) We adopt a tractable spherical return landscape. Intuitively, any smooth optimum admits a quadratic expansion; this spherical abstraction establishes a conservative lower bound for analyzing general smooth landscapes (Appendix~\ref{app:geometry_generalization}).  
(ii) We assume Frontier Alignment ($\langle \mathbf{g}_S, \mathbf{n}^* \rangle > 0$). Intuitively, since a self-locked policy has exhausted in-distribution improvements, the true optimal direction must inherently point outward, which we empirically validate in Section~\ref{subsec:sanity_check}.

\vspace{-5pt}
\section{Iterative Scarcity-Guided Exploration}
\label{sec:method}

\begin{figure}[!t]
    \centering
    \includegraphics[width=0.7\linewidth]{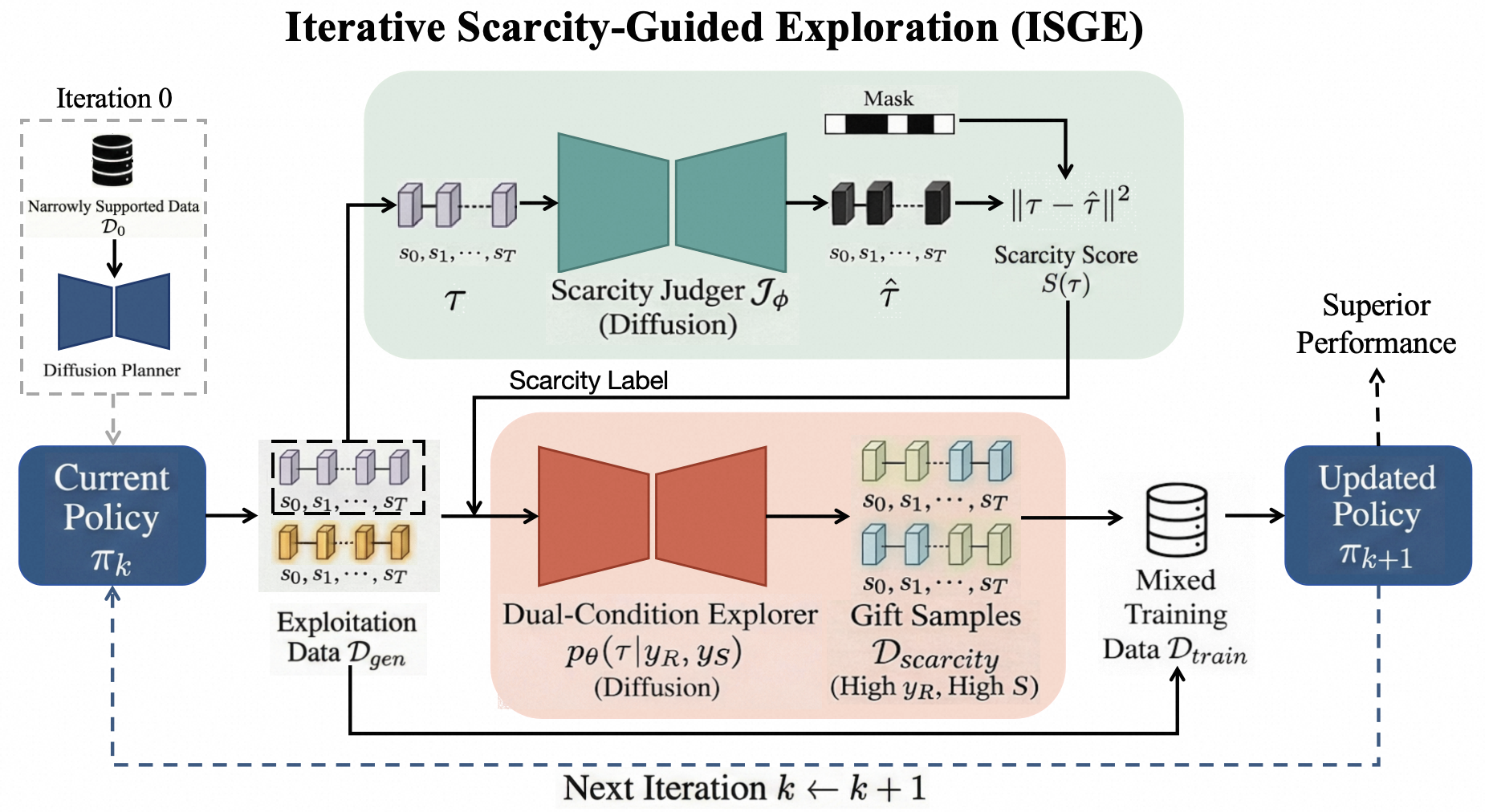}
    \caption{ISGE Framework}
    \label{fig:isge_framework}
    \vspace{-10pt}
\end{figure}

Building upon our theoretical insights, we present Iterative Scarcity-Guided Exploration (ISGE). As illustrated in Figure~\ref{fig:isge_framework}, ISGE breaks self-locking through a dynamic ``Generate-Judge-Explore-Update'' cycle, advancing the policy $\pi_k$ to $\pi_{k+1}$ by aggregating novel exploratory trajectories with conservative exploitation data. 
In contrast, vanilla self-training naively executes a ``Generate-Update'' loop by omitting the Judge (green panel) and Explore (red panel) modules.

\paragraph{Scarcity Evaluation via Reconstruction.} To operationalize Definition~\ref{def:scarcity}, we introduce a Scarcity Judger $\mathcal{J}_\phi$ (an unconditional diffusion model) trained on the current exploitation data $\mathcal{D}_{gen}$. It estimates the scarcity of a trajectory $\tau$ via the empirical reconstruction residual: $S(\tau) = \|\text{Mask}(\tau - \hat{\tau})\|^2$, where $\hat{\tau}$ is the Judger's reconstruction (see implementation details in Appendix~\ref{implementation_details}). 
Crucially, the $\text{Mask}(\cdot)$ operation restricts evaluation strictly to strategic dimensions (e.g., historical bids) while ignoring uncontrollable environmental stochasticity (e.g., traffic fluctuations). This isolates true strategic novelty from mere environmental randomness~\cite{pathak2017curiosity}.

\paragraph{Scarcity-Guided Diffusion for Exploration.} To discover high-return OOD ``gift samples'', we propose a \emph{Dual-Condition Diffusion} model acting as our Explorer, parameterized as $p_\theta(\tau | y_R, y_S)$. During sampling, we employ disentangled classifier-free guidance to execute the optimization handoff:
\begin{equation}\label{eq:dual_guidance}
\begin{aligned}
\hat{\epsilon}
&= \epsilon_\theta(\tau_t, \emptyset, t)
 + w_R \underbrace{\big( \epsilon_{\theta}(\tau_t, y_R, t)
 - \epsilon_\theta(\tau_t, \emptyset, t) \big)}_{\propto \mathbf{g}_R \text{ (Return Signal)}} 
 + w_S \underbrace{\big( \epsilon_{\theta}(\tau_t, y_S, t)
 - \epsilon_\theta(\tau_t, \emptyset, t) \big)}_{\propto \mathbf{g}_S \text{ (Scarcity Signal)}} ,
\end{aligned}
\end{equation}
where $w_R$ and $w_S$ control the respective guidance scales. The return term ($\mathbf{g}_R$) drives the trajectory to the suboptimal support boundary. At this point, the scarcity term ($\mathbf{g}_S$) takes over, injecting a robust outward impulse that steers the trajectory through the cone of improvement directions into unexplored, high-value regions.

\paragraph{Iterative Training and Dynamic Recalibration.} The overall framework ties these components together into a self-improving loop. At iteration $k$, we sample exploitation data $\mathcal{D}_{gen}$ from $\pi_k$. The Judger learns this current manifold density and scores $\mathcal{D}_{gen}$. Next, the Explorer is trained on these labeled samples and then deployed to generate a novel exploration batch $\mathcal{D}_{scarcity}$ conditioned on high return and high scarcity targets (e.g., the 99th percentile). Finally, the updated policy $\pi_{k+1}$ is trained on $\mathcal{D}_{gen} \cup \mathcal{D}_{scarcity}$. Each iteration dynamically recalibrates by specifically seeking trajectories that are scarce relative to the \textit{current} policy, ensuring efficient expansion without redundant exploration.

\section{Experiments}
\label{sec:experiments}

In this section, we evaluate ISGE on AuctionNet~\cite{auctionnet2024}, a large-scale industrial auto-bidding benchmark that served as the official environment for the NeurIPS 2024 Auto-bidding Competition~\cite{competition_ref} and remains widely adopted across the auto-bidding community~\cite{gas2025,gave2025,lei2025generative,10.1145/3774904.3792202,aigb-r1,ding2026langbp}.
To simulate the severe cold-start challenge, we construct cold-start datasets by retaining only trajectories within the $10\%$ to $30\%$ return percentile range. We evaluate our framework across two distinct settings: AuctionNet-Dense (with strict CPA constraints) and AuctionNet-Sparse (with sparse reward signals).
We aim to answer the following research questions: (1) Do the empirical dynamics in real-world auctions align with our theoretical predictions of return signal collapse and optimization handoff? (2) Can ISGE bootstrap superior policies from low-quality, narrowly supported data? (3) Is ISGE efficient and safe for practical auto-bidding? (4) How do the specific design choices contribute to the overall success?

\subsection{Sanity Check: Empirical Validation of Theoretical Abstractions}
\label{subsec:sanity_check}

To bridge our idealized geometric abstractions with the complex, non-convex AuctionNet environment, we conduct empirical sanity checks to validate the two core theoretical predictions: the return signal collapse (Proposition~\ref{prop:bottleneck}) and the optimization handoff via scarcity (\cref{thm:scarcity_revival}). 
Comprehensive experimental setups and supplementary quantitative results are deferred to Appendix~\ref{app:empirical_validation}.

\begin{figure}[t]
\centering
\captionsetup[subfigure]{font=footnotesize}
\begin{subfigure}{0.22\linewidth}
    \includegraphics[width=\linewidth]{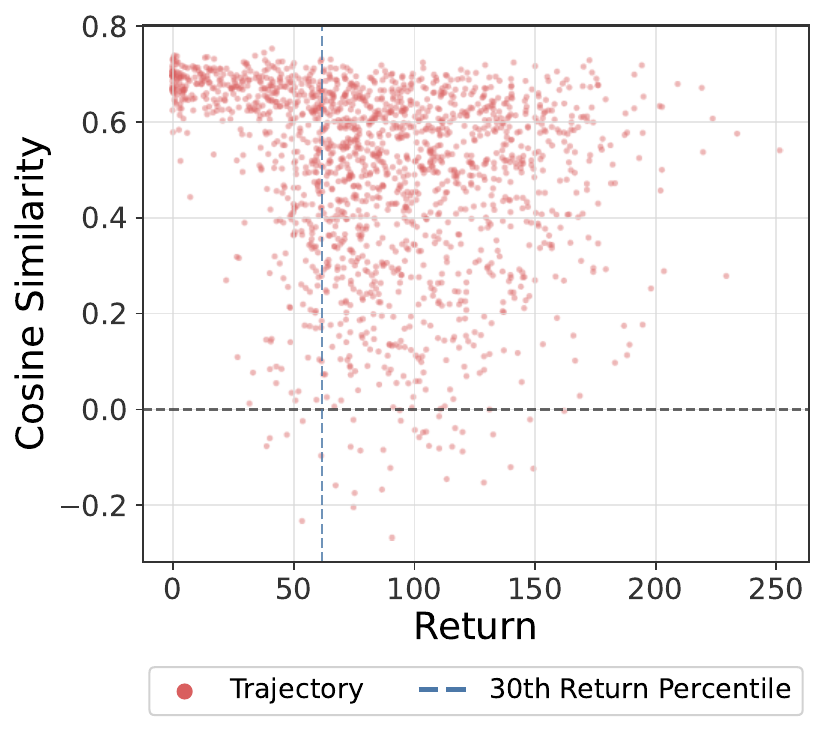}
    \caption{Step $t=15$}
\end{subfigure}
\hfill
\begin{subfigure}{0.22\linewidth}
    \includegraphics[width=\linewidth]{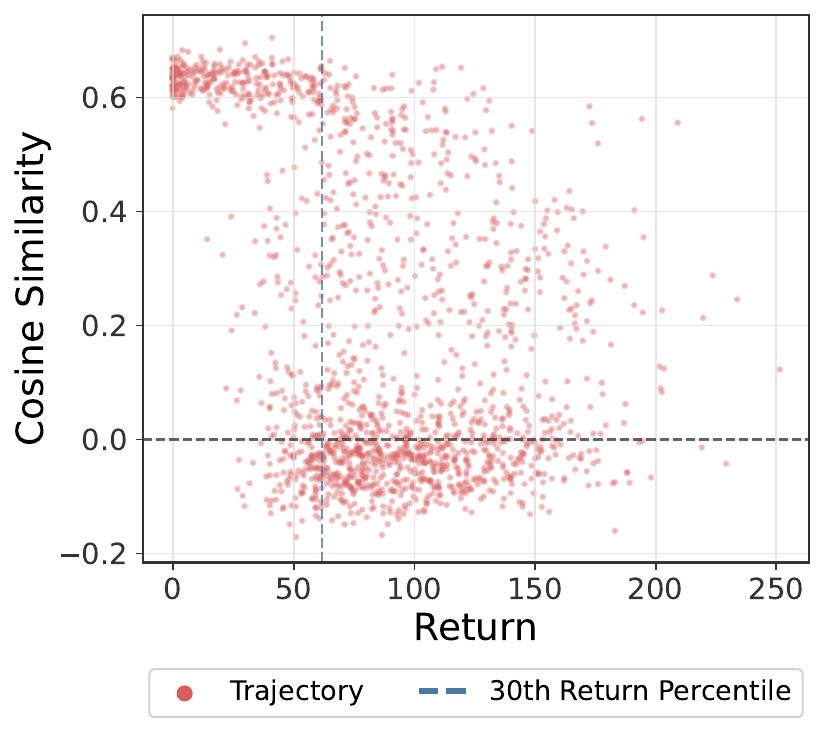}
    \caption{Step $t=5$ (Collapse)}
\end{subfigure}
\hfill
\rule{0.5pt}{3.3cm}   
\hfill
\begin{subfigure}{0.22\linewidth}
    \includegraphics[width=\linewidth]{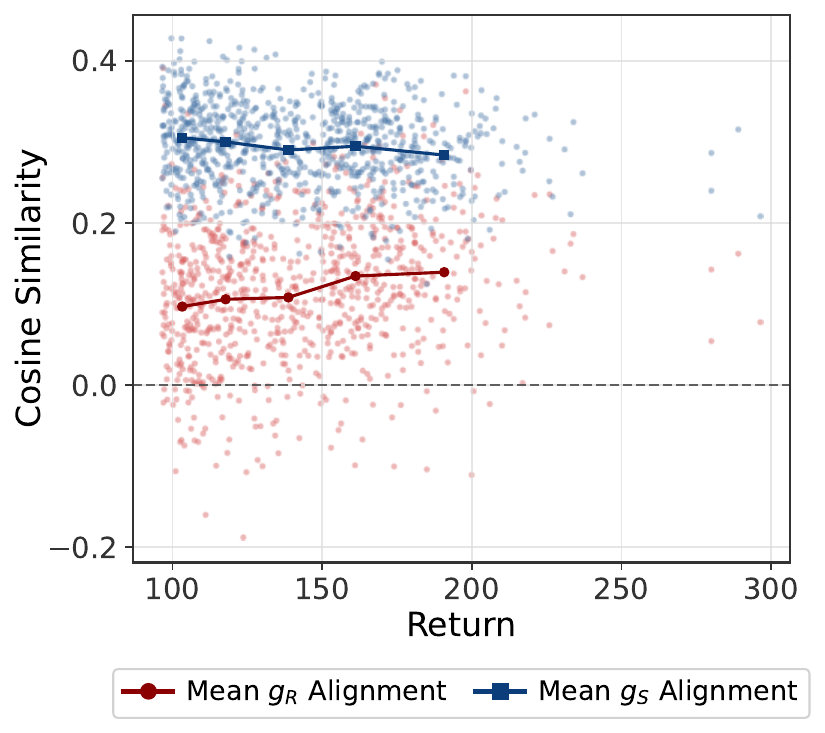}
    \caption{Step $t=15$}
\end{subfigure}
\hfill
\begin{subfigure}{0.22\linewidth}
    \includegraphics[width=\linewidth]{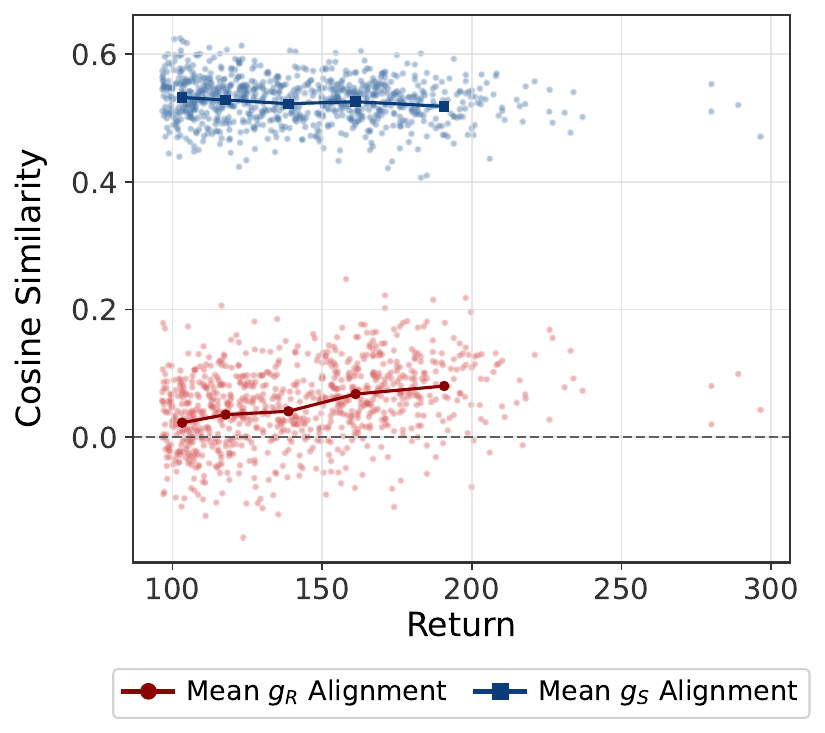}
    \caption{Step $t=5$ (Handoff)}
\end{subfigure}
\caption{
\textbf{Left:} At late denoising stage ($t=5$), return gradient $\mathbf{g}_R$ maintains alignment within training support but sharply decays toward zero in the high-return OOD region, triggering self-locking.
\textbf{Right:} Tracking high-return trajectories reveals that when return gradient $\mathbf{g}_R$ (red) collapses at $t=5$, scarcity gradient $\mathbf{g}_S$ (blue) maintains a positive alignment with the true denoising gradient $\mathbf{g}^*$.}
\label{fig:combined_collapse_handoff}
\end{figure}

\textbf{Validating Return Signal Collapse (Proposition~\ref{prop:bottleneck}).} 
To verify whether the return gradient loses its directional alignment when extrapolating into OOD regions, we evaluate the cosine similarity between the empirical return gradient $\mathbf{g}_R$ (from a cold-start planner trained only on the 10\%--30\% return quantile) and the empirical optimal direction $\tilde{\mathbf{n}}^*$ (derived from a Full-Data Oracle). 
As shown in Figure~\ref{fig:combined_collapse_handoff} (left), during the fine-grained late denoising stage ($t=5$, where the total denoising steps run from $20$ down to $0$), $\cos(\tilde{\mathbf{n}}^*, \mathbf{g}_R)$ remains robust in lower-return regions but drops sharply to $0.11$ in the high-return OOD region (>30\% return quantile). This perfectly corroborates the theoretical self-locking condition: the return guidance is physically incapable of maintaining optimal directional alignment once it crosses the suboptimal boundary.

\textbf{Validating Optimization Handoff via Scarcity (\cref{thm:scarcity_revival}).}
To justify our critical assumption that the outward-pointing scarcity gradient aligns positively with the optimal improving direction ($\langle \mathbf{g}_S, \mathbf{n}^* \rangle > 0$), we analyze the top high-return gift samples discovered by the ISGE planner. We measure the alignment of both $\mathbf{g}_R$ and the scarcity gradient $\mathbf{g}_S$ against the true denoising gradient $\mathbf{g}^*$ required for the perfect recovery of these top gift samples.
Figure~\ref{fig:combined_collapse_handoff} (right) illustrates a clear temporal handoff mechanism. At the late boundary-crossing stage ($t=5$), the return gradient's alignment collapses ($\cos \approx 0.05$). In contrast, the scarcity gradient maintains a robust positive alignment ($\cos \approx 0.53$). Because $\mathbf{g}_R$ has successfully navigated the trajectory to the high-return frontier in earlier steps, the non-vanishing, outward-pointing $\mathbf{g}_S$ subsequently takes over, providing the necessary directional impulse to break self-locking.
\input{main_result_table}
\vspace{-5pt}

\subsection{Main Results: Bootstrapping from Narrow Support}
\label{subsec:main_results}
We evaluate ISGE against a Vanilla Self-Training baseline. 
To rigorously isolate the efficacy of our iterative framework, all methods presented in Table~\ref{tab:main_results} share the identical underlying diffusion planner architecture~\cite{aigb2024}. 
Specifically, the Cold-Start Base is trained exclusively on the low-quality cold-start data, serving as the shared starting point for both Vanilla Self-Training and ISGE. 
The Full-Data Oracle serves as an empirical reference for the diffusion planner architecture, possessing global visibility into full offline data.
Performance is measured by Score (primary metric), Total Reward (total conversions), CPA Exceedance Ratio (ER), and Win Rate (WR). Complete metric definitions, implementation details, and additional comparisons against non-diffusion baselines are provided in Appendix~\ref{experiments_details}. 
Robustness analysis on more iterations and an additional cold-start setting can be found in Appendix~\ref{app:extended_results}.

Table \ref{tab:main_results} highlights a stark contrast in performance evolution. Vanilla Self-Training succumbs to self-locking: it saturates far below the Full-Data Oracle in the Dense setting ($77.8$ vs. $102.0$) and regresses in the Sparse setting (Iter $1 \to 3$: $16.9 \to 16.0$).
In contrast, ISGE effectively breaks this bottleneck. It not only outperforms baselines but surpasses the Full-Data Oracle in both settings. Notably, in AuctionNet-Dense, ISGE achieves a Score of $108.4$ (vs. Full-Data Oracle $102.0$) with significantly better constraint satisfaction (ER $54.3\%$ vs. $74.9\%$), demonstrating Pareto-optimal improvement. By Iteration 3, ISGE attains relative gains of \textbf{+66.0\%} (Dense) and \textbf{+34.4\%} (Sparse). 
This consistent growth confirms that the guidance handoff mechanism successfully breaks self-locking, continuously repairing the optimization landscape and enabling robust bootstrapping from a severely suboptimal initialization.

\begin{figure}[t]
\centering
\captionsetup[subfigure]{font=footnotesize}

\begin{minipage}[t]{0.68\linewidth}
\centering

\begin{subfigure}[t]{0.32\linewidth}
    \centering
    \includegraphics[width=\linewidth]{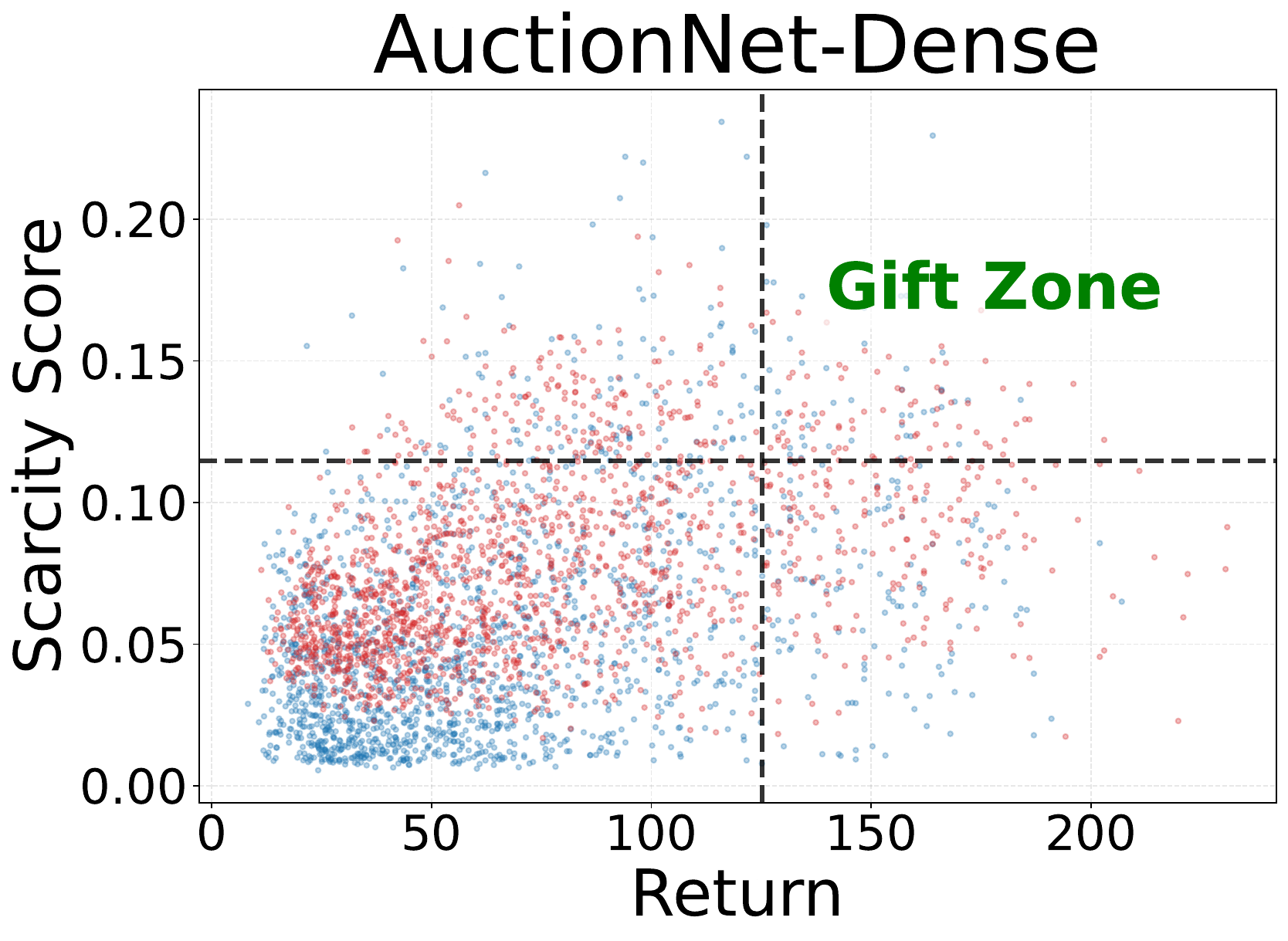}
    \caption{ASN (Gift: 4.9\%)}
\end{subfigure}
\hfill
\begin{subfigure}[t]{0.32\linewidth}
    \centering
    \includegraphics[width=\linewidth]{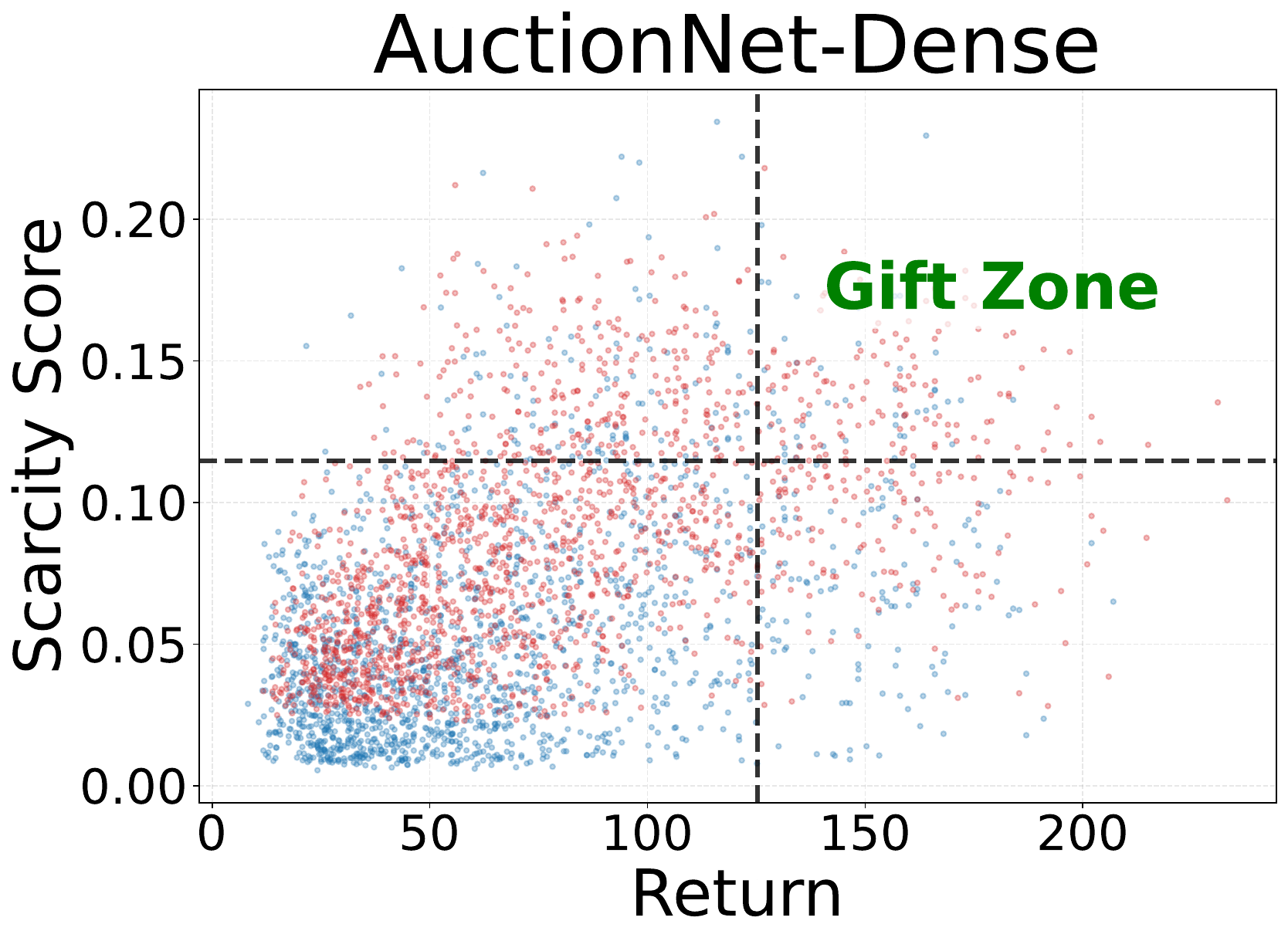}
    \caption{PSN (Gift: 7.9\%)}
\end{subfigure}
\hfill
\begin{subfigure}[t]{0.32\linewidth}
    \centering
    \includegraphics[width=\linewidth]{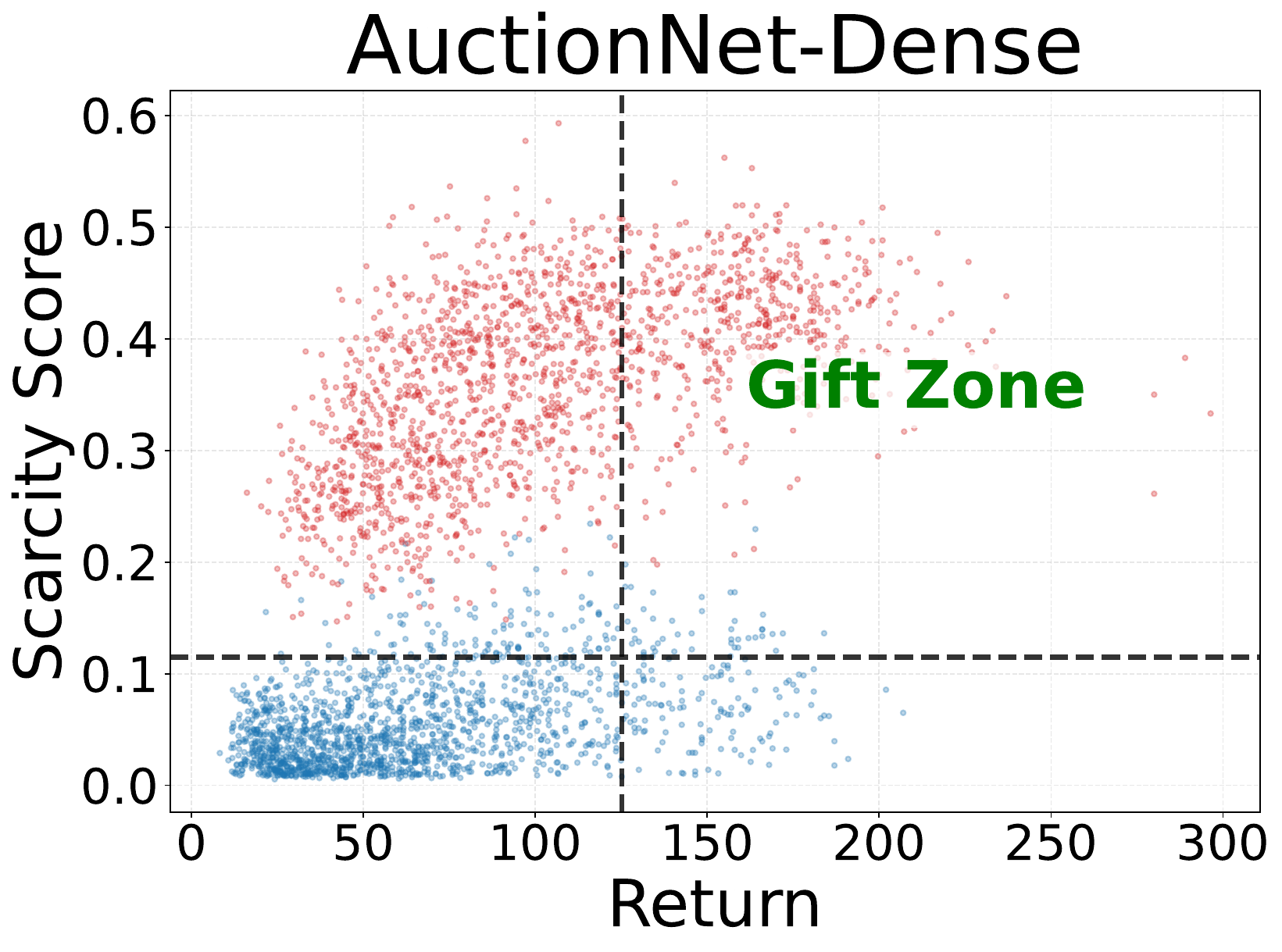}
    \caption{\textbf{ISGE (Gift: 30.4\%)}}
\end{subfigure}

\vspace{0.5em}

\begin{subfigure}[t]{0.32\linewidth}
    \centering
    \includegraphics[width=\linewidth]{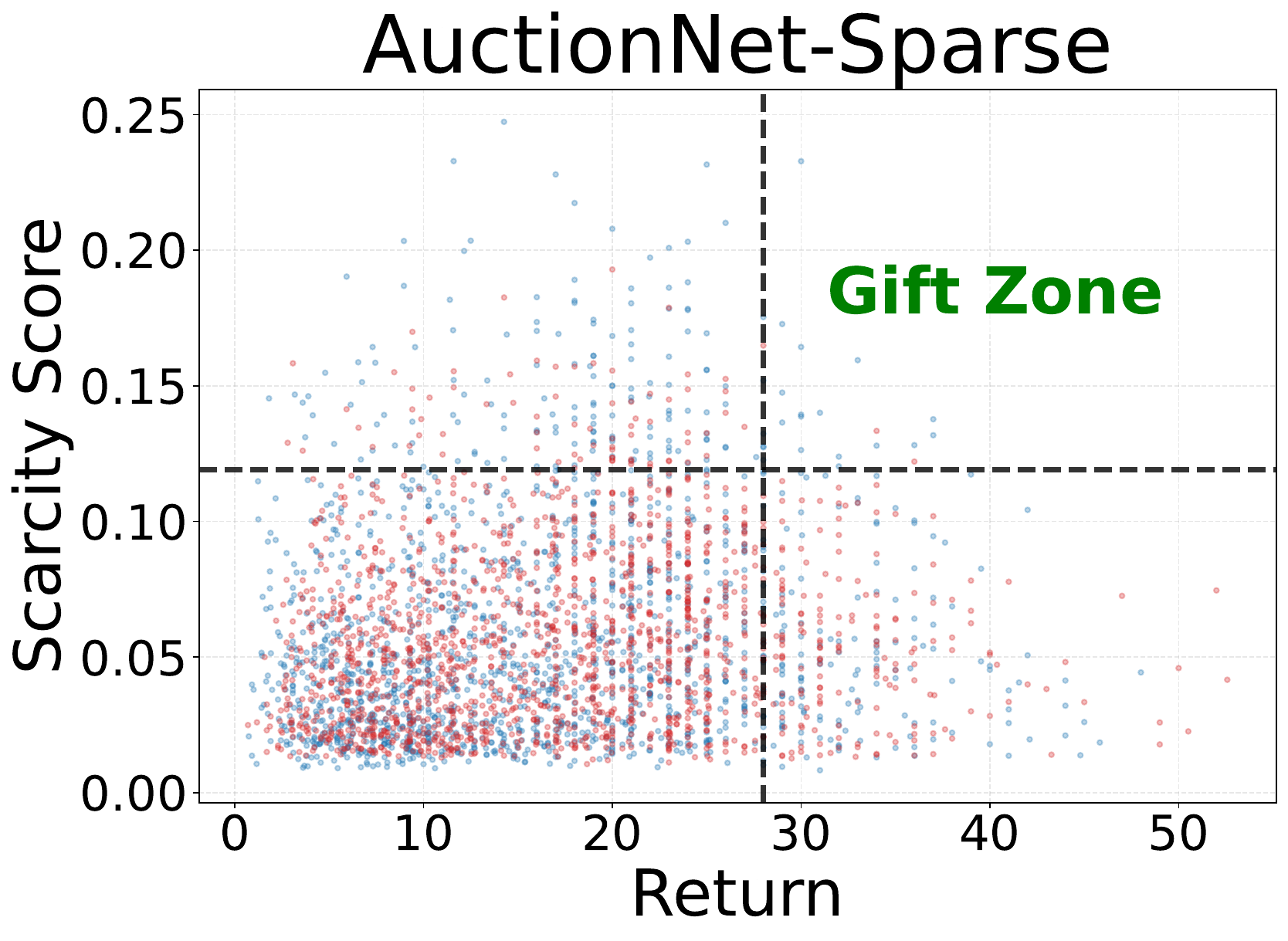}
    \caption{ASN (Gift: 0.1\%)}
\end{subfigure}
\hfill
\begin{subfigure}[t]{0.32\linewidth}
    \centering
    \includegraphics[width=\linewidth]{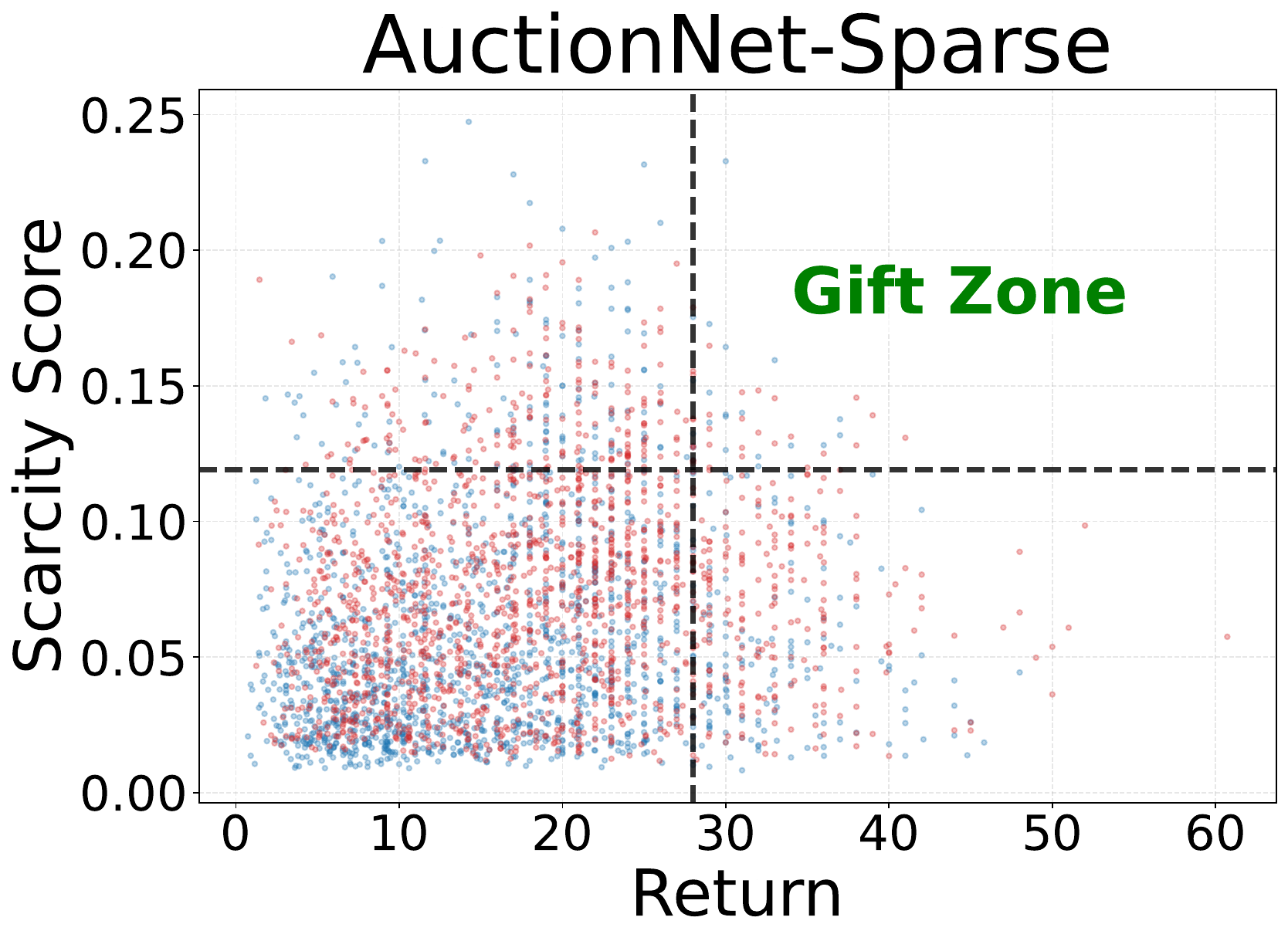}
    \caption{PSN (Gift: 1.2\%)}
\end{subfigure}
\hfill
\begin{subfigure}[t]{0.32\linewidth}
    \centering
    \includegraphics[width=\linewidth]{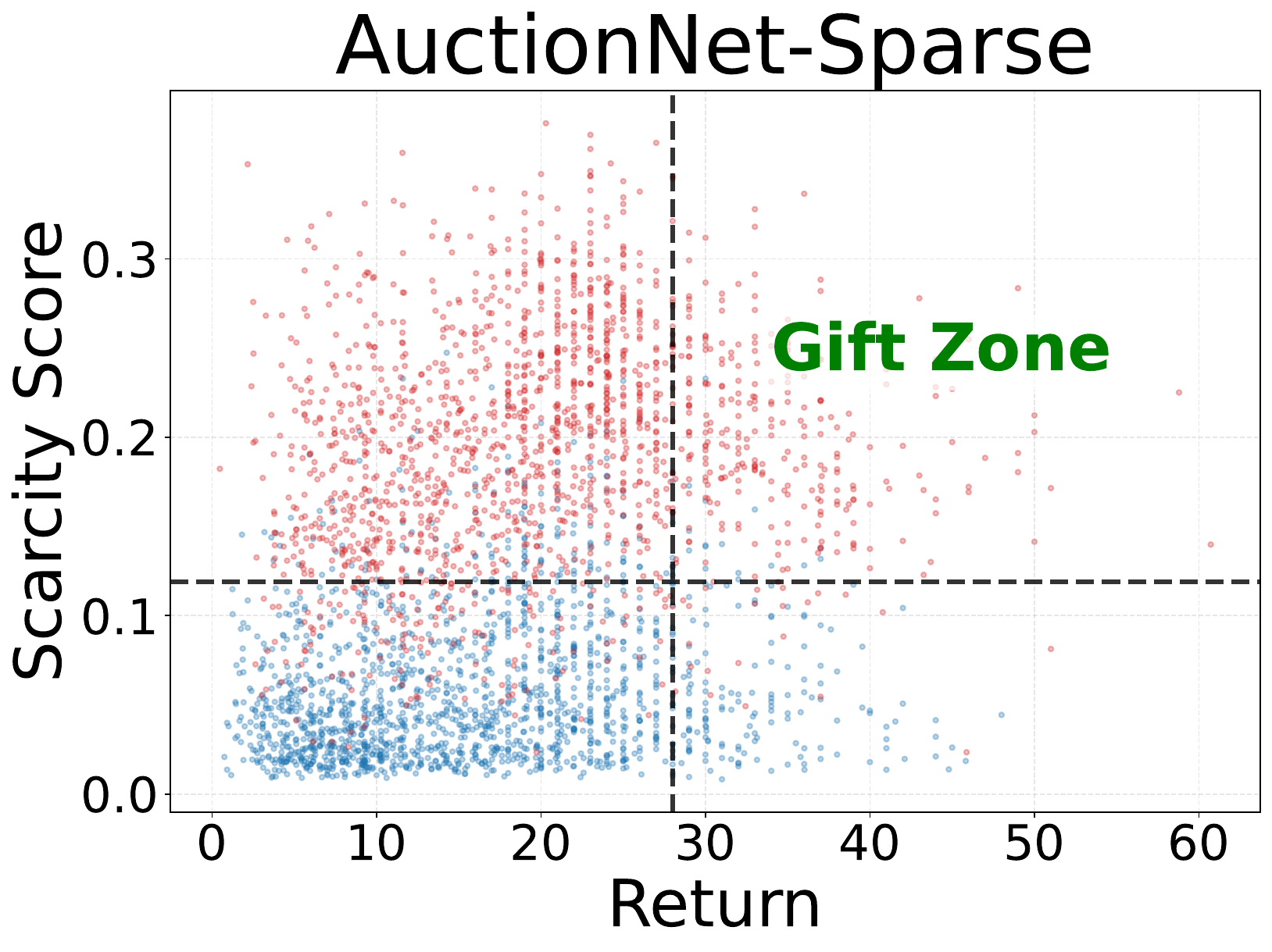}
    \caption{\textbf{ISGE (Gift: 16.2\%)}}
\end{subfigure}

\end{minipage}
\hfill
\begin{minipage}[t]{0.30\linewidth}
\centering

\begin{subfigure}[t]{\linewidth}
    \centering
    \includegraphics[width=0.95\linewidth]{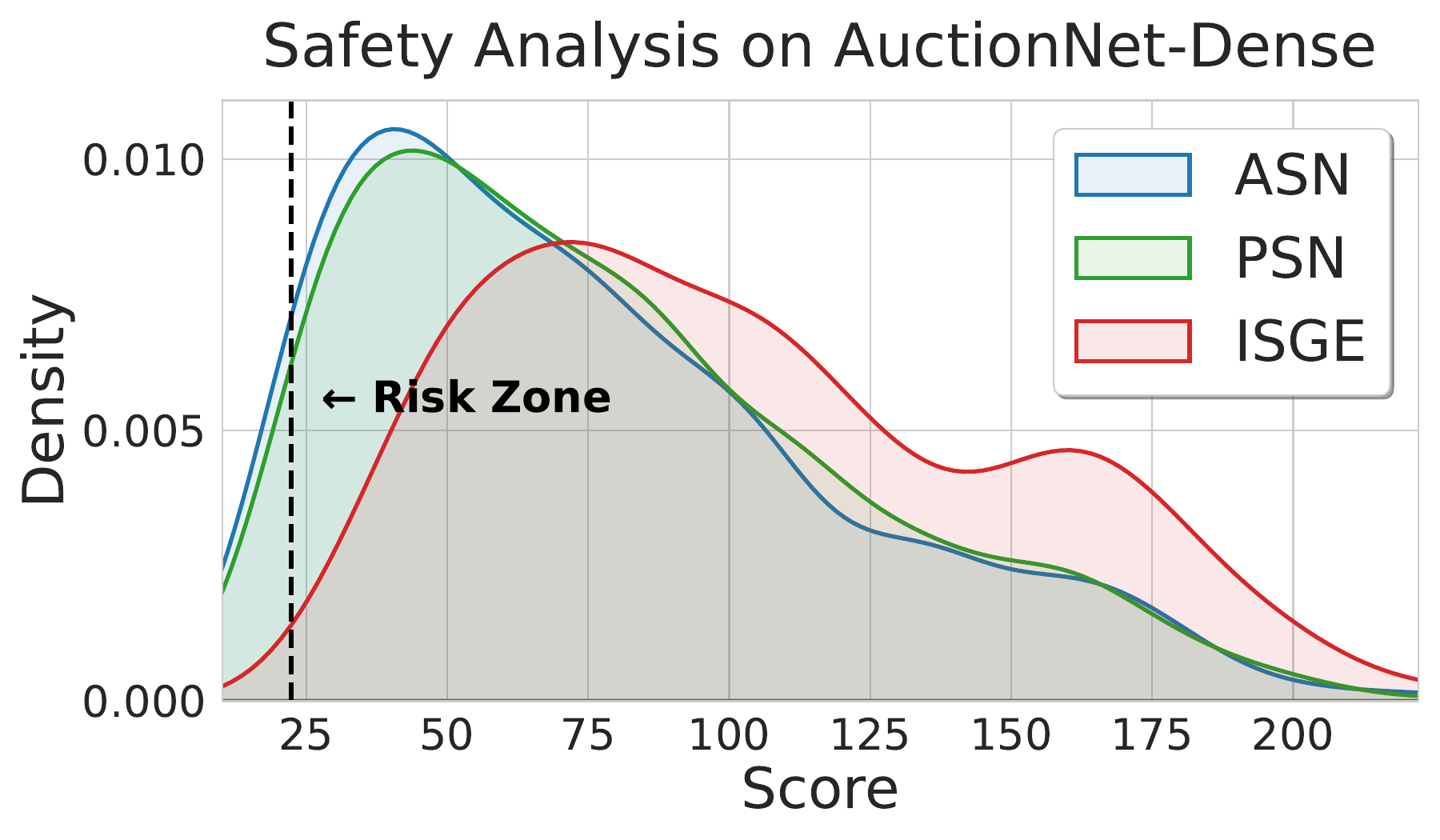}
    \caption{Risk(Dense): ASN$>$PSN$>$ISGE}
\end{subfigure}

\vspace{0.3em}
\begin{subfigure}[t]{\linewidth}
    \centering
    \includegraphics[width=0.95\linewidth]{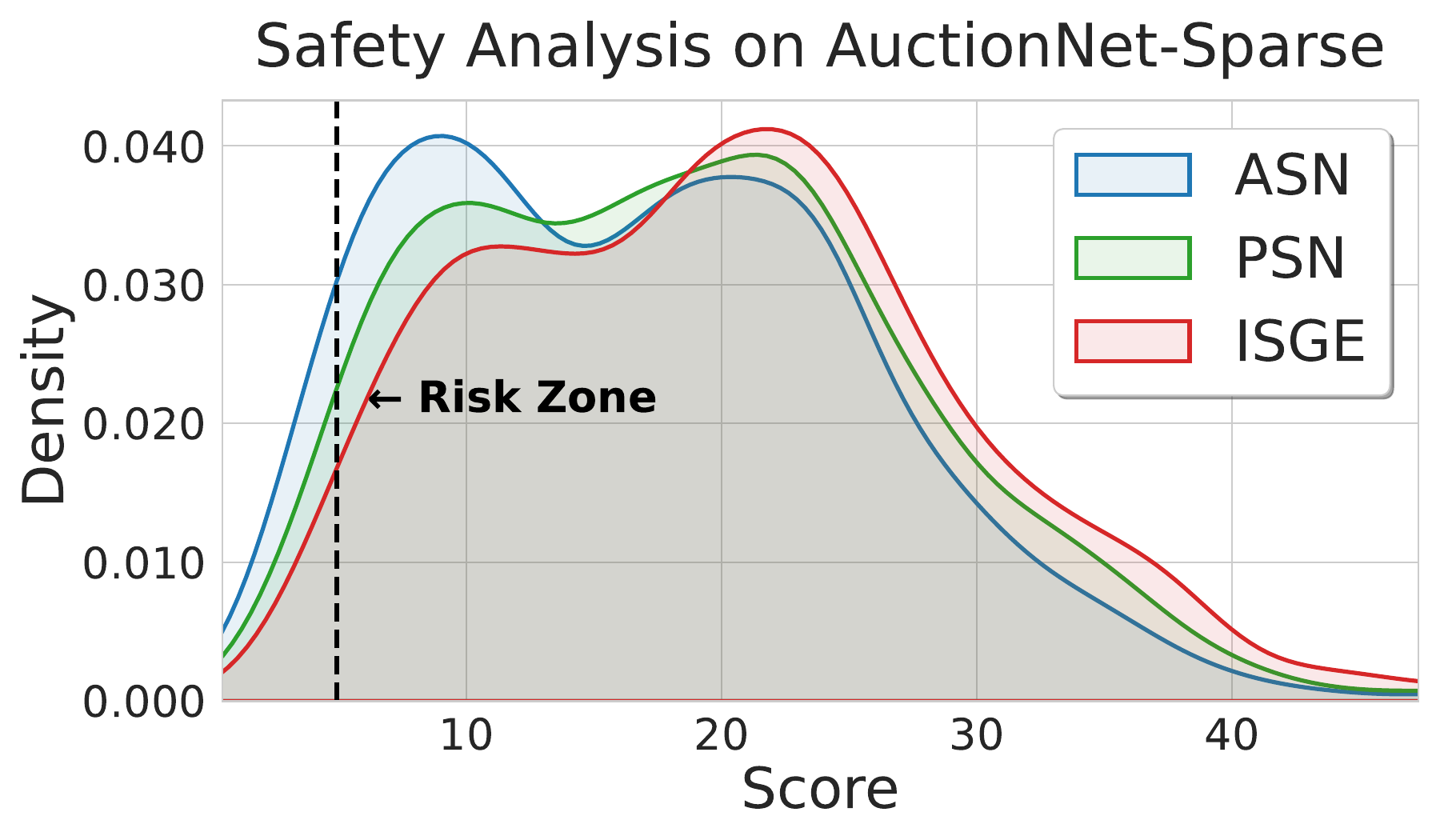}
    \caption{Risk(Sparse): ASN$>$PSN$>$ISGE}
\end{subfigure}

\end{minipage}

\caption{
\textbf{Exploration Efficiency and Safety Analysis.}
\textbf{Left~(a)-(f):} Scarcity--Return landscapes for Dense and Sparse environments. Red and blue points denote exploration and baseline (V1) trajectories, respectively. The \emph{Gift Zone} (top-right) represents high-value strategies exceeding the top 10\% thresholds of both return and scarcity computed on the baseline dataset.
\textbf{Right~(g)-(h):} Kernel density estimates of exploration score distributions for ASN (blue), PSN (green), and ISGE (red).
The \emph{Risk Zone} indicates performance falling below the 10\% quantile of the baseline (V1) distribution. Risk proportion (Dense, Sparse): ASN (4.7\%, 6.8\%), PSN (3.6\%, 4.1\%), \textbf{ISGE (0.1\%, 2.6\%)}.
}
\label{fig:efficiency_and_safety}
\end{figure}

\begin{wraptable}{r}{0.60\linewidth}
\vspace{-0.4cm}
\centering
\footnotesize
\captionsetup{font=footnotesize}
\caption{\textbf{Performance of Directed vs. Undirected Exploration.} 
Metrics are reported for Iteration 1 relative to Vanilla Self-Training (V1).
ISGE significantly outperforms undirected baselines (ASN/PSN) across all metrics.}
\label{tab:dataset_metric_comparison}
\vspace{-0.2cm}
\resizebox{\linewidth}{!}{
\begin{tabular}{l|l|cccc}
\toprule
Dataset & Metrics & Vanilla (V1) & ASN & PSN & \textbf{ISGE} \\
\midrule
\multirow{5}{*}{Dense}
& Score $\uparrow$ & $74.9 \pm 3.4$ & $81.7 \pm 4.0$ & $81.8 \pm 3.8$ & \bm{$103.4 \pm 5.3$} \\
& Reward $\uparrow$ & $192.5 \pm 2.8$ & $200.8 \pm 3.9$ & $201.0 \pm 3.9$ & \bm{$219.1 \pm 4.6$} \\
& ER $\downarrow$ & $85.2 \pm 3.2\%$ & $77.7 \pm 4.0\%$ & $77.4 \pm 4.1\%$ & \bm{$58.8 \pm 3.6\%$} \\
& WR $\uparrow$ & $5.6 \pm 0.1\%$ & $5.8 \pm 0.1\%$ & $5.8 \pm 0.1\%$ & \bm{$6.1 \pm 0.1\%$} \\
& \emph{Improve} $\uparrow$ & -- & $9.1\%$ & $9.2\%$ & \bm{$38.1\%$} \\
\midrule
\multirow{5}{*}{Sparse}
& Score $\uparrow$ & $16.9 \pm 0.5$ & $18.0 \pm 0.5$ & $18.2 \pm 0.6$ & \bm{$20.0 \pm 0.9$} \\
& Reward $\uparrow$ & $27.2 \pm 0.5$ & $28.1 \pm 0.5$ & $28.2 \pm 0.5$ & \bm{$29.6 \pm 0.6$} \\
& ER $\downarrow$ & $35.3 \pm 3.3\%$ & $30.9 \pm 2.9\%$ & $30.4 \pm 3.0\%$ & \bm{$23.7 \pm 3.3\%$} \\
& WR $\uparrow$ & $8.1 \pm 0.1\%$ & $8.3 \pm 0.1\%$ & $8.2 \pm 0.1\%$ & \bm{$8.5 \pm 0.1\%$} \\
& \emph{Improve} $\uparrow$ & -- & $6.5\%$ & $7.7\%$ & \bm{$18.3\%$} \\
\bottomrule
\end{tabular}
}

\vspace{-0.6em}
\end{wraptable}
\subsection{ISGE Achieves Efficient and Safe Exploration}
\label{subsec:mechanism}
To validate that our gains stem from scarcity guidance rather than simple randomness, we compare ISGE against two standard undirected exploration baselines in auto-bidding: Action Space Noise (ASN)~\cite{mou2022sustainable} and Parameter Space Noise (PSN)~\cite{li2024trajectory} (see Appendix~\ref{experiments_details} for details). Using Iteration 1 Vanilla (V1) as the baseline, Table \ref{tab:dataset_metric_comparison} shows that ASN and PSN yield only marginal improvements ($\sim9\%$ in Dense, $\sim8\%$ in Sparse). This stagnation supports our hypothesis that in high-dimensional manifolds, isotropic noise is statistically orthogonal to sparse directions of improvement. 
In contrast, ISGE achieves a transformative leap, improving the Score by 38.1\% in Dense and 18.3\% in Sparse settings. Crucially, ISGE dominates in both reward maximization and constraint satisfaction, successfully navigating the trade-off between aggressive exploration and strict CPA limits.

\paragraph{Mechanism Analysis: Efficiency and Safety.   }
We further investigate this success by visualizing the \textit{Scarcity-Return landscape} of the explored trajectories in Figure \ref{fig:efficiency_and_safety} (a)-(f).
We define the ``Gift Zone'' (top-right quadrant) as the target region containing high-scarcity, high-return trajectories. 
As observed in the scatter plots, undirected baselines (ASN/PSN) are inefficient, capturing negligible samples in the Gift Zone (e.g., $\le$ 1.2\% in Sparse), proving that random perturbations fail to stumble upon high-value strategies in data vacuums.
Conversely, ISGE’s dual-condition explorer functions as a directed guide, efficiently achieving capture rates of \textbf{30.4\%} (Dense) and \textbf{16.2\%} (Sparse). 
Furthermore, regarding \emph{safety} (downside risk), Figure \ref{fig:efficiency_and_safety} (g)-(h) reveals that undirected methods exhibit heavy left-tail distributions falling into the ``Risk Zone'' (up to 6.8\% risk). 
In contrast, ISGE explicitly truncates this downside risk ($\le$ 0.1\% in Dense, $\le$ 2.6\% in Sparse) via return conditioning. 
This demonstrates that ISGE robustly expands the performance upper bound without compromising the reliable lower bound.

Furthermore, to demonstrate the superiority of our trajectory-level scarcity guidance over traditional step-wise intrinsic motivation, we provide additional comparisons against transition-level curiosity-driven exploration (e.g., ICM \cite{pathak2017curiosity} and PGR \cite{wang2025prioritized}) in Appendix~\ref{app:curiosity_baseline}.

\subsection{Ablation Studies}
\label{subsec:ablation}
\begin{wraptable}{r}{0.42\linewidth}
\vspace{-0.8cm}
\centering
\small
\captionsetup[subtable]{font=footnotesize,labelformat=empty}
\begin{subtable}[t]{\linewidth}
\centering
\caption{Table 3(a): \textbf{Masking vs. No Masking.}}
\label{tab:masking_comparison}

\resizebox{0.97\linewidth}{!}{
\begin{tabular}{l|cc}
\toprule
Metrics & No Masking & \textbf{Masking} \\
\midrule
Score  &$94.7\pm4.5$ &\bm{$103.4\pm5.3$} \\
Gift \% & $19.1\%$ & \bm{$30.4\%$} \\
\bottomrule
\end{tabular}%
}
\end{subtable}

\begin{subtable}[t]{\linewidth}
\centering
\caption{Table 3(b): \textbf{Feature Scarcity Growth.}}
\label{tab:feature_scarcity}
\footnotesize
\resizebox{0.95\linewidth}{!}{
\begin{tabular}{l|l|cc}
\toprule
\multicolumn{2}{c|}{\textbf{State Feature}} & \textbf{$\mathcal{D}_{gen}$} & \textbf{$\mathcal{D}_{scarcity}$}  \\
\midrule
\multirow{2}{*}{\rotatebox{90}{Strat.}} 
  & Budget Left & 0.056 & 0.105  \\
  & History Bid & 0.025 & 0.486  \\
\midrule
\multirow{2}{*}{\rotatebox{90}{Env.}} 
  & Traffic Volume & 0.052 & 0.062  \\
  & Traffic Value & 0.229 & 0.235\\
\bottomrule
\end{tabular}}
\end{subtable}
\vspace{-0.3cm}
\end{wraptable}
\paragraph{Effect of Strategic Masking.} 
ISGE calculates scarcity solely on controllable policy states, masking out uncontrollable environmental dimensions. Removing this mask degrades performance and significantly reduces the gift sample capture rate from 30.4\% to 19.1\% (Table \ref{tab:masking_comparison}). This decline stems from a fundamental distraction: granular analysis (Table \ref{tab:feature_scarcity}) reveals that environmental features like \textit{Traffic Value} exhibit negligible scarcity growth (i.e., from $\mathcal{D}_{gen}$ to $\mathcal{D}_{scarcity}$) because the agent cannot ``invent'' market conditions. Consequently, including these immutable dimensions introduces futile noise that distracts the explorer from controllable strategic features (e.g., \textit{Budget Left}), thereby diluting the gradient signal of scarcity for valid exploration.

\paragraph{Hyperparameter Sensitivity \& Mechanism Validation.}
We analyze the impact of guidance scales and data mixture ratios (Table \ref{tab:sensitivity}) to validate the core design principles of ISGE.

\begin{wraptable}{r}{0.63\linewidth}
\vspace{-5pt}
\centering
\footnotesize
\captionsetup{font=footnotesize}
\caption{\textbf{Hyperparameter Analysis.} We study the impact of (a) Guidance Scale balance and (b) Data Mixture Ratios. The default settings are highlighted in gray.}
\label{tab:sensitivity}
\vspace{-0.2cm}
\resizebox{\linewidth}{!}{
\begin{tabular}{l|c|ccc|c}
\toprule
\multicolumn{6}{c}{\textbf{(a) CFG Guidance Balance ($w_R : w_S$)}} \\
\midrule
\multirow{2}{*}{Setting ($w_R, w_S$)} & \multirow{2}{*}{Ratio} 
& \multicolumn{3}{c|}{Exploration Data Quality} & Final Policy \\
\cmidrule(lr){3-5} \cmidrule(lr){6-6}
& & Avg Score & Avg Scarcity & \textbf{Gift \%} & \textbf{Score} \\
\midrule
Reward-Heavy ($3,1$) & 3:1 & 88.6 & 0.14 & 19.7\% & $90.2\pm3.8$ \\
\rowcolor{gray!15} Balanced ($3,3$) & 1:1 & 102.9 & 0.36 & 30.4\% & $103.4\pm5.3$ \\
Scarcity-Heavy ($3,5$) & 3:5 & \textbf{105.4} & \textbf{0.59} & \textbf{32.2\%} & \bm{$105.5\pm6.1$} \\
\midrule
\multicolumn{6}{c}{\textbf{(b) Data Mixture Ratio (Explore : Base)}} \\
\midrule
\multicolumn{3}{c|}{Mixture Ratio} & \multicolumn{3}{c}{\textbf{Score}} \\
\midrule
\multicolumn{3}{c|}{Base-Only (0 : 1)} & \multicolumn{3}{c}{$74.9\pm3.4$} \\
\multicolumn{3}{c|}{Conservative (1 : 2)} & \multicolumn{3}{c}{$103.3\pm6.9$} \\
\rowcolor{gray!15} \multicolumn{3}{c|}{\textbf{Balanced (5 : 4)}} 
& \multicolumn{3}{c}{\bm{$103.4\pm5.3$}} \\
\multicolumn{3}{c|}{Aggressive (2 : 1)} & \multicolumn{3}{c}{$101.6\pm5.3$} \\
\multicolumn{3}{c|}{Explore-Only (1 : 0)} & \multicolumn{3}{c}{$96.7\pm6.0$} \\
\bottomrule
\end{tabular}
}

\vspace{-3pt}
\end{wraptable}

\textit{(a) The Necessity of Proper Scarcity Guidance.}
Table \ref{tab:sensitivity}(a) reveals that weak scarcity guidance scale ($w_S=1$) fails to effectively capture gift samples ($19.7\%$), dropping performance to $90.2$. This confirms that scarcity identification alone is insufficient; a sufficiently large $w_S$ is required to mechanically ``push'' the trajectory away from the high-density manifold against the strong attraction of the diffusion prior. This empirically validates our theoretical analysis (Theorem \ref{thm:scarcity_revival}) that proper guidance scale $w_S$ is necessary for scarcity-guided exploration.

\textit{(b) The Necessity of Data Mixing.}
Table \ref{tab:sensitivity}(b) demonstrates that while \emph{Explore-Only} ($96.7$) significantly outperforms \emph{Base-Only} ($74.9$), it still lags behind the mixed policy ($103.4$). This highlights the complementary nature of the datasets: exploration data fixes policy ``blind spots'' (high-value/sparse regions) but sacrifices global coverage due to its targeted bias. Consequently, mixing is essential to combine the evolutionary signal of exploration with the structural stability of the base distribution.

\section{Related Work}\label{related_work}
\paragraph{Auto-bidding and Generative Auto-bidding}
Traditionally, auto-bidding has been formulated as a constrained optimization problem, typically addressed through PID controllers~\cite{zhang2016feedback,yang2019bid} or primal-dual frameworks~\cite{balseiro2021landscape,aggarwal2024auto}.
To handle dynamic market environments, Reinforcement Learning (RL) methods were subsequently introduced to maximize cumulative returns~\cite{cai2017real,wu2018budget,USCB,wen2022cooperative,rtbsurvey2023}. However, they struggle with ineffective credit assignment in non-Markovian long-horizon scenarios. 
Recently, the field has undergone a paradigm shift toward AIGB, which reformulates auto-bidding as a generative sequence modeling task.
Pioneered by DiffBid~\cite{aigb2024}, this paradigm utilizes conditional diffusion models to capture the joint distribution of bidding trajectories~\cite{cbd2025}, while subsequent Transformer-based approaches like GAVE~\cite{gave2025}, GAS~\cite{gas2025} and GRAD~\cite{lei2025generative} employ autoregressive planning with return-to-go conditioning.
To mitigate the limitations of static offline datasets, recent research has focused on offline improvement mechanisms~\cite{peng2025expert,aigbpearl2025,jiang2025optimal}.
However, these methods typically rely on high-quality expert demonstrations or broad data coverage.
In contrast, our work targets the iterative cold-start setting, focusing on bootstrapping policies purely from low-quality, narrowly supported data without expert priors.

\paragraph{Decision Making and Exploration with Diffusion Models}
Diffusion models have emerged as a versatile paradigm for decision-making. 
Diffuser~\cite{janner2022planning}, Decision Diffuser~\cite{ajay2022conditional}, and DiffuserLite~\cite{dong2024diffuserlite}, directly treat planning as a conditional trajectory generation problem. Alternatively, methods such as Diffusion-QL~\cite{wang2022diffusion}, DTQL~\cite{chen2024diffusion}, and MetaDiffuser~\cite{ni2023metadiffuser} leverage diffusion for policy representation.
Beyond policy fitting, recent studies investigate diffusion-based data augmentation~\cite{lu2023synthetic} and exploration.
Approaches like PGR~\cite{wang2025prioritized}, ICM~\cite{pathak2017curiosity}, and ExDM~\cite{ying2025exploratory} rely on novelty or uncertainty scores to guide exploration.
These methods predominantly operate at the \emph{transition level} ($s_t, a_t \to s_{t+1}$) for RL policies, utilizing forward prediction error or diffusion noise prediction error as local signals. 
However, auto-bidding is a strictly non-Markovian problem governed by global constraints: the budget spent at step $t$ limits the feasible actions at step $T$, thereby transition-level metrics are insufficient for capturing the long-horizon strategic structure.
In contrast, ISGE adopts trajectory reconstruction error to perform \emph{trajectory-level} exploration, which inherently aligns with the sequential modeling nature of diffusion planners.
(We provide an empirical comparison demonstrating the advantage of our trajectory-level scarcity over transition-level curiosity in Appendix \ref{app:curiosity_baseline}.)
Furthermore, regarding the utilization of exploration signals, AdaptDiffuser~\cite{liang2023adaptdiffuser} achieves self-evolution by filtering generated plans via a learned discriminator, but lacks exploration beyond the support of the initial generation.
Conversely, ISGE actively injects the scarcity signal as a condition during the diffusion sampling process via CFG.
This mechanism allows our explorer to extrapolate beyond the current support, purposefully seeking ``gift samples'' in high-scarcity regions that filtering methods would rarely encounter.

\section{Conclusion}
\label{sec:conclusion}
In this paper, we addressed the critical challenge of bootstrapping auto-bidding policies from narrowly supported data. 
We revealed that diffusion planners suffer from a self-locking bottleneck due to an inherent SNR collapse in OOD regions. 
To break this limit, we proposed ISGE, a novel iterative framework. 
Our geometric analysis proves that scarcity provides a robust outward impulse, executing a successful guidance handoff when the return signal fails. 
By iteratively exploring and aggregating high-return ``gift samples'', ISGE continuously repairs the optimization landscape. 
Extensive experiments on AuctionNet demonstrate that ISGE efficiently navigates the OOD vacuum, achieving transformative performance growth beyond cold-start initialization. 

While highly effective for bootstrapping, ISGE assumes local landscape continuity, and its outward exploration may become less effective near optimality (see Appendix~\ref{app:limitations} for detailed discussions).
Furthermore, our framework is currently designed for diffusion planners. 
Future work will focus on extending ISGE's core philosophy of trajectory-level scarcity guidance to other generative decision-making architectures, such as autoregressive Transformers.

\begin{ack}
This work is supported by the National Natural Science Foundation of China (Grant No. 62572010), Alibaba Group through the Alibaba Innovative Research Program, and Peking University-Alimama Joint Laboratory of AI Innovation. We thank all anonymous reviewers for their helpful feedback.
\end{ack}

\bibliographystyle{unsrtnat}
\bibliography{references}
\newpage
\input{checklist.tex}
\newpage
\setcounter{tocdepth}{2}
\tableofcontents


\appendix
\newpage
\section{Detailed Formulation of Auto-Bidding and MDP Modeling}
\label{app:detailed_formulation}

Following DiffBid~\cite{aigb2024}, we provide a comprehensive formalization of the auto-bidding problem and its translation into a sequential decision-making process, following established settings in the literature.

\subsection{Problem Formulation}
During a specific advertising period, suppose there are $N$ impression opportunities arriving sequentially, indexed by $i$. Advertisers submit bids to compete for each impression opportunity. An advertiser wins the impression if its bid $b_i$ surpasses all others, subsequently incurring a cost $c_i$ for winning and obtaining the corresponding value.

The fundamental mission of an advertiser is to maximize the total received value $\sum_i o_i v_i$, where $v_i$ is the value of impression $i$ and $o_i \in \{0, 1\}$ is a binary variable indicating whether the advertiser wins impression $i$. Furthermore, the ad delivery is subject to budget and various performance constraints. The budget constraint is simply formulated as $\sum_i o_i c_i \le B$, where $B$ is the total budget. To unify other complex key performance indicator (KPI) constraints, we adopt the generalized formulation $\frac{\sum_i c_{ij} o_i}{\sum_i p_{ij} o_i} \le C_j$, where $C_j$ is the upper bound of the $j$-th constraint, $p_{ij}$ represents any performance indicator (e.g., conversions), and $c_{ij}$ represents the cost incurred for impression $i$ under the $j$-th constraint (e.g., the cost used to compute the CPA for that constraint).

Given $J$ constraints, the Multi-constrained Bidding (MCB) problem is formally defined as:
\begin{equation}
\label{eq:mcb_problem}
\begin{aligned}
    \mathop{\text{maximize}}_{o_i} \quad & \sum\nolimits_{i=1}^N o_i v_i \\
    \text{s.t.} \quad & \sum\nolimits_{i=1}^N o_i c_i \le B \\
    & \frac{\sum_{i=1}^N c_{ij} o_i}{\sum_{i=1}^N p_{ij} o_i} \le C_j, \quad \forall j \in \{1, \dots, J\} \\
    & o_i \in \{0, 1\}, \quad \forall i
\end{aligned}
\end{equation}
Previous studies~\cite{USCB} have shown that the optimal solution to this constrained optimization takes the form:
\begin{equation}
    b_i^* = \lambda_0 v_i + C_j \sum\nolimits_{j=1}^J \lambda_j p_{ij},
    \label{eq:bid_formula}
\end{equation}
where $b_i^*$ is the predicted optimal bid for impression $i$, and $\lambda_j$ ($j \in \{0, \dots, J\}$) are the optimal bidding parameters. For instance, optimizing with only the budget constraint corresponds to Max Return bidding, whereas incorporating the CPA constraint forms Target-CPA bidding.

\subsection{Auto-Bidding as Decision-Making}
While Eq.~(\ref{eq:bid_formula}) provides the theoretical formulation of the optimal bid, the highly random and complex nature of real-world advertising environments prevents the direct calculation of the optimal bidding parameters $\lambda_j$. Instead, these parameters must be carefully calibrated to adapt to the environment and dynamically adjusted as it evolves. This dynamic adjustment inherently transforms auto-bidding into a sequential decision-making problem. 

To model this process, we define the key components of the automated bidding agent interacting with the industrial online advertising system:
\begin{itemize}[leftmargin=2em]
    \item \textbf{State $\mathbf{s}_t \in \mathcal{S}$}: Describes the real-time advertising status at time period $t$, which encapsulates features such as: 1) the remaining time of the advertiser; 2) the remaining budget; 3) the budget spend speed; 4) real-time cost-efficiency; and 5) average cost-efficiency.
    \item \textbf{Action $\mathbf{a}_t \in \mathcal{A}$}: Indicates the adjustment made to the bidding parameters at time period $t$. The action space has the dimension corresponding to the number of bidding parameters $\lambda_j$, modeled as a vector $(a_t^{\lambda_0}, \dots, a_t^{\lambda_J})$.
    \item \textbf{Reward $\mathbf{r}_t \in \mathcal{R}$}: The specific value contributed to the overall objective obtained within the time period $t$.
    \item \textbf{Trajectory $\tau$}: Represents the complete sequence of states, actions, and rewards within an episode. Note that when formulating this as a generative modeling problem in subsequent sections, we denote this trajectory sequence as $\bm{x}$ to align with standard diffusion formulation conventions.
\end{itemize}

In real-world online advertising systems, learning a policy through direct trial-and-error interaction with the live environment is unfeasible due to severe financial and safety concerns. Consequently, policies must be learned entirely from offline historical bidding logs, which contain trajectories $\tau$ generated by a variety of behavior strategies. This offline sequential decision-making formulation perfectly aligns with our generative paradigm, where we explicitly model and optimize the generation of these high-return trajectories (re-denoted as $\bm{x}$) via diffusion planners.

\section{Missing Proofs}
\label{missing_proofs}
\subsection{Derivation of the CFG Biased Langevin Update}
\label{proof:cfg_update}

Let $\epsilon_\theta(\bm{x}_t)$ denote the unconditional noise prediction. By leveraging the score matching property, we first reformulate the Classifier-Free Guidance (CFG) modified noise prediction $\hat{\epsilon}_\theta(\bm{x}_t)$ into a score-based gradient. Substituting this into the standard DDPM update reveals that the sampling process at step $t$ manifests as a biased Langevin update:
\begin{equation}
    \bm{x}_{t-1} \approx \mu_\theta(\bm{x}_t) + \lambda_t w \cdot \nabla_{\bm{x}_t} \log p(y \mid \bm{x}_t) + \sigma_t \mathbf{z},
\end{equation}
where $\lambda_t = \frac{1-\alpha_t}{\sqrt{\alpha_t}}$ is a positive coefficient derived from the noise schedule. \textbf{Under the idealized assumption of perfect density estimation (commonly adopted in theoretical analysis of diffusion models), the approximation becomes exact.}

\paragraph{Step 1: Reformulating CFG via Score Matching.}
Recall the definition of CFG. The modified noise prediction is a linear interpolation between the unconditional prediction $\epsilon_\theta(\bm{x}_t)$ and the conditional prediction $\epsilon_\theta(\bm{x}_t, y)$:
\begin{equation}
    \hat{\epsilon}_\theta(\bm{x}_t) = \epsilon_\theta(\bm{x}_t) + w \cdot \big( \epsilon_\theta(\bm{x}_t, y) - \epsilon_\theta(\bm{x}_t) \big).
    \label{eq:cfg_def}
\end{equation}

According to the denoising score matching objective~\cite{song2021score}, the optimal noise prediction network approximates the score of the data distribution (scaled by the noise level):
\begin{align}
    \epsilon_\theta(\bm{x}_t) &\approx -\sqrt{1-\bar{\alpha}_t} \, \nabla_{\bm{x}_t} \log p_t(\bm{x}_t), \label{eq:score_uncond} \\
    \epsilon_\theta(\bm{x}_t, y) &\approx -\sqrt{1-\bar{\alpha}_t} \, \nabla_{\bm{x}_t} \log p_t(\bm{x}_t \mid y). \label{eq:score_cond}
\end{align}
\textbf{Note on the approximation ($\approx$):} These relationships are approximate due to the finite capacity and training imperfections of the neural network; they become exact only in the limit of perfect density estimation.

Applying Bayes' rule to the score function yields $\nabla_{\bm{x}_t} \log p_t(\bm{x}_t \mid y) = \nabla_{\bm{x}_t} \log p_t(\bm{x}_t) + \nabla_{\bm{x}_t} \log p(y \mid \bm{x}_t)$. Substituting Eqs.~\eqref{eq:score_uncond} and~\eqref{eq:score_cond} into the CFG difference term cancels the unconditional score:
\begin{align}
    \epsilon_\theta(\bm{x}_t, y) - \epsilon_\theta(\bm{x}_t) 
    &\approx -\sqrt{1-\bar{\alpha}_t} \left( \nabla_{\bm{x}_t} \log p_t(\bm{x}_t \mid y) - \nabla_{\bm{x}_t} \log p_t(\bm{x}_t) \right) \nonumber \\
    &= -\sqrt{1-\bar{\alpha}_t} \cdot \nabla_{\bm{x}_t} \log p(y \mid \bm{x}_t).
\end{align}
Substituting this back into Eq.~\eqref{eq:cfg_def}, we obtain the effective noise prediction:
\begin{equation}
    \hat{\epsilon}_\theta(\bm{x}_t) \approx \epsilon_\theta(\bm{x}_t) - w \sqrt{1-\bar{\alpha}_t} \cdot \nabla_{\bm{x}_t} \log p(y \mid \bm{x}_t).
    \label{eq_app_noise}
\end{equation}

\paragraph{Step 2: Deriving the Biased Langevin Update.}
In the standard DDPM sampling process~\cite{ho2020denoising}, the transition from $\bm{x}_t$ to $\bm{x}_{t-1}$ is given by:
\begin{equation}
    \bm{x}_{t-1} = \frac{1}{\sqrt{\alpha_t}} \left( \bm{x}_t - \frac{1-\alpha_t}{\sqrt{1-\bar{\alpha}_t}} \hat{\epsilon}_\theta(\bm{x}_t) \right) + \sigma_t \mathbf{z},
    \label{eq:ddpm_update}
\end{equation}
where $\mathbf{z} \sim \mathcal{N}(0, \mathbf{I})$. Substituting Eq.~\eqref{eq_app_noise} into Eq.~\eqref{eq:ddpm_update}:
\begin{equation}
    \bm{x}_{t-1} \approx \frac{1}{\sqrt{\alpha_t}} \left( \bm{x}_t - \frac{1-\alpha_t}{\sqrt{1-\bar{\alpha}_t}} \left[ \epsilon_\theta(\bm{x}_t) - w \sqrt{1-\bar{\alpha}_t} \, \nabla_{\bm{x}_t} \log p(y \mid \bm{x}_t) \right] \right) + \sigma_t \mathbf{z}.
\end{equation}

We define $\mu_\theta(\bm{x}_t) = \frac{1}{\sqrt{\alpha_t}} \big( \bm{x}_t - \frac{1-\alpha_t}{\sqrt{1-\bar{\alpha}_t}} \epsilon_\theta(\bm{x}_t) \big)$ as the standard reconstruction mean toward the unconditional manifold. For the remaining guidance term, the coefficient $\sqrt{1-\bar{\alpha}_t}$ elegantly cancels out:
\begin{align}
    \text{Drift Term} 
    &= \frac{1}{\sqrt{\alpha_t}} \left( - \frac{1-\alpha_t}{\sqrt{1-\bar{\alpha}_t}} \right) \cdot \left( - w \sqrt{1-\bar{\alpha}_t} \, \nabla_{\bm{x}_t} \log p(y \mid \bm{x}_t) \right) \nonumber \\
    &= \frac{1-\alpha_t}{\sqrt{\alpha_t}} \, w \cdot \nabla_{\bm{x}_t} \log p(y \mid \bm{x}_t).
\end{align}

Defining $\lambda_t = \frac{1-\alpha_t}{\sqrt{\alpha_t}} > 0$, the update rule simplifies to the biased Langevin dynamic:
\begin{equation}
    \bm{x}_{t-1} \approx \mu_\theta(\bm{x}_t) + \lambda_t w \nabla_{\bm{x}_t} \log p(y \mid \bm{x}_t) + \sigma_t \mathbf{z}.
\end{equation}

\paragraph{Step 3: From Approximation to Equality.}
To decouple our subsequent geometric analysis and theorems from the empirical approximation errors of neural networks, we adopt the standard theoretical abstraction of perfect density estimation. Under this idealized setting, the score-matching mapping becomes exact, allowing us to drop the approximation ($\approx$) and rigorously formalize the sampling step as an exact equality:
\begin{equation}
    \bm{x}_{t-1} = \mu_\theta(\bm{x}_t) + \lambda_t w \nabla_{\bm{x}_t} \log p(y \mid \bm{x}_t) + \sigma_t \mathbf{z}.
\end{equation}
This exact formulation serves as the foundational equation for analysis of trajectory optimization in Section~\ref{sec:theory}.

\subsection{Proof of Geometric Barrier and Dynamics Failure (Section~\ref{subsec:self_locking})}
\label{app:proof_dynamics_failure}
In this section, we provide the proofs for Theorem \ref{thm:geometric_barrier} and Proposition \ref{prop:bottleneck}. 
We first establish the general geometric condition for distance reduction in high-dimensional space (Theorem \ref{thm:geometric_barrier}) and then apply this condition to the specific Langevin dynamics of diffusion planners to demonstrate the self-locking mechanism (Proposition \ref{prop:bottleneck}).

\subsubsection{Proof of Theorem~\ref{thm:geometric_barrier} (Geometric Barrier)}
\label{proof:geometric_barrier}
\begin{proof}
Without loss of generality, place the optimum at the origin ($\bm{x}^* = \mathbf{0}$), so that $d^2 = \|\bm{x}\|^2$ and the unit inward radial direction is $\mathbf{n}^* = -\bm{x} / d$. The update is $\Delta \bm{x} = \mathbf{v}_{\mathrm{det}} + \mathbf{v}_{\mathrm{noise}}$.
The expected squared distance of the new state $\bm{x}' = \bm{x} + \Delta \bm{x}$ is:
\begin{align}
    \mathbb{E}[\|\bm{x}'\|^2] 
    &= \mathbb{E}[\|\bm{x} + \mathbf{v}_{\mathrm{det}} + \mathbf{v}_{\mathrm{noise}}\|^2] \nonumber \\
    &= \|\bm{x}\|^2 + 2\langle \bm{x}, \mathbf{v}_{\mathrm{det}} \rangle + \|\mathbf{v}_{\mathrm{det}}\|^2 + \mathbb{E}[\|\mathbf{v}_{\mathrm{noise}}\|^2] + 2\mathbb{E}[\langle \bm{x} + \mathbf{v}_{\mathrm{det}}, \mathbf{v}_{\mathrm{noise}} \rangle].
\end{align}
Since $\mathbf{v}_{\mathrm{noise}} \sim \mathcal{N}(0, \sigma^2 \mathbf{I})$, the cross-term vanishes ($\mathbb{E}[\mathbf{v}_{\mathrm{noise}}] = 0$), and $\mathbb{E}[\|\mathbf{v}_{\mathrm{noise}}\|^2] = D\sigma^2$. Substituting $\bm{x} = -d \mathbf{n}^*$:
\begin{equation}
    \mathbb{E}[\Delta (d^2)] = \mathbb{E}[\|\bm{x}'\|^2] - d^2 = -2d \langle \mathbf{v}_{\mathrm{det}}, \mathbf{n}^* \rangle + \|\mathbf{v}_{\mathrm{det}}\|^2 + D\sigma^2.
\end{equation}
Letting $v_{\mathrm{rad}} = \langle \mathbf{v}_{\mathrm{det}}, \mathbf{n}^* \rangle$, the exact condition for distance reduction ($\mathbb{E}[\Delta (d^2)] < 0$) is $2d v_{\mathrm{rad}} > \|\mathbf{v}_{\mathrm{det}}\|^2 + D\sigma^2$, i.e., $\text{SNR} = \frac{2d v_{\mathrm{rad}}}{\|\mathbf{v}_{\mathrm{det}}\|^2 + D\sigma^2} >1$.
\end{proof}

\begin{figure}[h]
  \centering
  \includegraphics[width=0.4\linewidth]{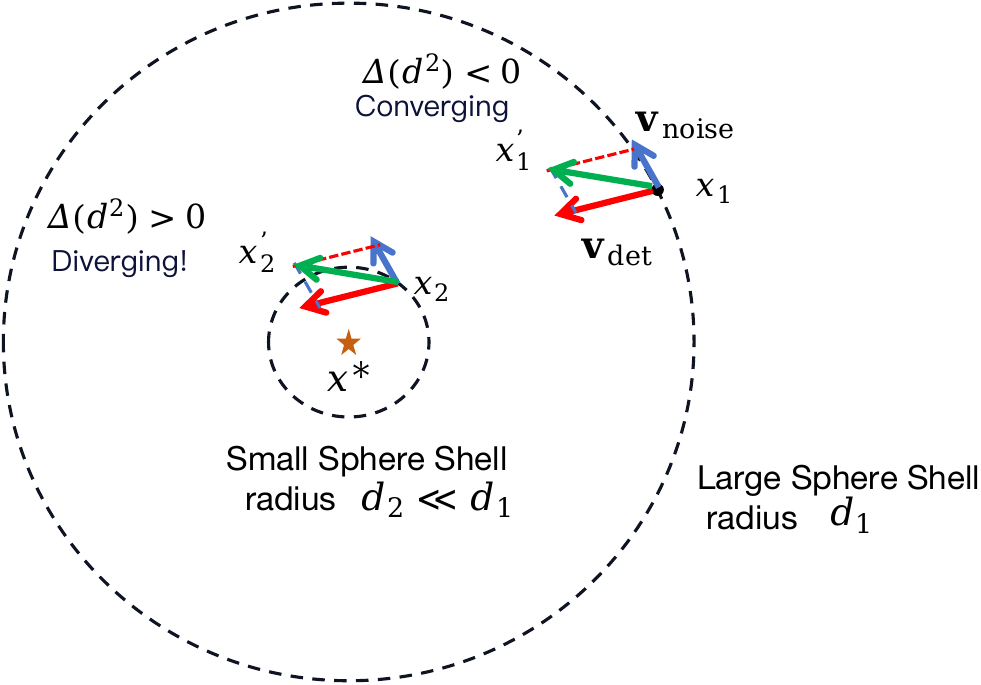}
  \caption{2D Geometric illustration of Theorem~\ref{thm:geometric_barrier}}
  \label{fig:demo}
\end{figure}

\paragraph{2D Geometric Intuition}
Figure \ref{fig:demo}  provides a geometric illustration of Theorem~\ref{thm:geometric_barrier}. Consider two points $\bm{x}_1$ and $\bm{x}_2$ located at distances $d_1$ and $d_2$ from the optimum, respectively, with $d_2 \ll d_1$. The update direction at any point is decomposed into a deterministic drift component $\mathbf{v}_{\text{det}}$ and an isotropic noise component $\mathbf{v}_{\text{noise}}$. In the outer region (Large Sphere Shell, radius $d_1$), the combined effect of drift and noise produces a gradient $\Delta \bm{x}_1 = \mathbf{v}_{\text{det}} + \mathbf{v}_{\text{noise}}$ that reduces the distance to the optimum ($\Delta (d^2) < 0$), resulting in convergent behavior. In contrast, in the inner region (Small Sphere Shell, radius $d_2$), the same decomposition produces $\Delta \bm{x}_2 = \mathbf{v}_{\text{det}} + \mathbf{v}_{\text{noise}}$ that increases the distance from the optimum ($\Delta (d^2) > 0$), leading to divergent behavior.

\subsubsection{Proof of Proposition~\ref{prop:bottleneck}}
\label{proof:bottleneck}
\begin{proof}
We isolate the return-guided step from the unconditional mean: $\mathbf{v}_{\mathrm{det}} = \lambda w_R \mathbf{g}_R$ and $\mathbf{v}_{\mathrm{noise}} = \sigma \mathbf{z}$.
We define the velocity of squared distance reduction as $V_{sq}(w_R) := -\mathbb{E}[\Delta (d^2)]_R$.
Substituting the terms into Theorem~\ref{thm:geometric_barrier}:
\begin{equation}
    V_{sq}(w_R) = 2d \lambda w_R \langle \mathbf{g}_R, \mathbf{n}^* \rangle - \lambda^2 w_R^2 \|\mathbf{g}_R\|^2 - D\sigma^2.
\end{equation}
To find the theoretical limit of convergence, we solve for the optimal scale $w_R^*$ that maximizes $V_{sq}(w_R)$:
\begin{equation}
    \frac{d V_{sq}}{d w_R} = 2d \lambda \langle \mathbf{g}_R, \mathbf{n}^* \rangle - 2\lambda^2 w_R \|\mathbf{g}_R\|^2 = 0 \implies w_R^* = \frac{d \langle \mathbf{g}_R, \mathbf{n}^* \rangle}{\lambda \|\mathbf{g}_R\|^2}.
\end{equation}
Substituting $w_R^*$ back into $V_{sq}(w_R)$ gives the maximum possible convergence velocity:
\begin{equation}
    V_{sq}(w_R^*) = \frac{d^2 \langle \mathbf{g}_R, \mathbf{n}^* \rangle^2}{\|\mathbf{g}_R\|^2} - D\sigma^2.
\end{equation}
Optimization stagnates when even this maximum velocity is non-positive ($V_{sq}(w_R^*) \leq 0$):
\begin{equation}
    \frac{d^2 \langle \mathbf{g}_R, \mathbf{n}^* \rangle^2}{\|\mathbf{g}_R\|^2} \leq D\sigma^2 \implies \langle \mathbf{g}_R, \mathbf{n}^* \rangle \leq \frac{\sigma \sqrt{D} \, \|\mathbf{g}_R\|}{d} (\text{i.e., }\text{SNR}\leq 1).
\end{equation}
\end{proof}

\subsubsection{Generalization to Local Quadratic Landscapes}
\label{app:geometry_generalization}

In Section~\ref{sec:theory}, we derived the geometric barrier under an idealized \textit{spherical} return landscape to geometrically define the Signal-to-Noise Ratio (SNR). Here, we justify its applicability to general smooth optimization landscapes by extending the analysis to \textit{general quadratic basins}. Since any smooth non-degenerate local optimum can be approximated by a quadratic form via Taylor expansion, this generalization mathematically bridges our theoretical abstraction with complex continuous environments.

\paragraph{1. Local Quadratic Model and Expected Return Dynamics.}
Let $\bm{x}^*$ be a strict local maximum of the return function $R(\bm{x})$. By Taylor expansion,
\begin{equation}
    R(\bm{x}) \approx R(\bm{x}^*) - \frac{1}{2} (\bm{x} - \bm{x}^*)^\top \mathbf{A} (\bm{x} - \bm{x}^*),
\end{equation}
where $\mathbf{A} = -\nabla^2 R(\bm{x}^*) \succ 0$ is the positive definite Hessian matrix. 

Consider an optimization update $\Delta \bm{x} = \mathbf{v}_{\mathrm{det}} + \mathbf{v}_{\mathrm{noise}}$, consistent with the biased Langevin dynamics of diffusion planners, where $\mathbf{v}_{\mathrm{noise}} \sim \mathcal{N}(0, \sigma^2 \mathbf{I})$. The expected change in the return is given by:
\begin{align}
    \mathbb{E}[\Delta R] &= \mathbb{E}\left[ R(\bm{x} + \Delta \bm{x}) - R(\bm{x}) \right] \nonumber \\
    &= - (\bm{x} - \bm{x}^*)^\top \mathbf{A} \mathbf{v}_{\mathrm{det}} - \frac{1}{2} \mathbf{v}_{\mathrm{det}}^\top \mathbf{A} \mathbf{v}_{\mathrm{det}} - \frac{1}{2} \mathbb{E}\left[ \mathbf{v}_{\mathrm{noise}}^\top \mathbf{A} \mathbf{v}_{\mathrm{noise}} \right].
\end{align}

\paragraph{2. The Universality of the $O(D)$ Noise Penalty.}
The critical term inhibiting optimization is the noise-induced penalty: $\mathcal{P}_{\text{noise}} = \frac{1}{2} \mathbb{E}\left[ \mathbf{v}_{\mathrm{noise}}^\top \mathbf{A} \mathbf{v}_{\mathrm{noise}} \right]$.
Using the properties of the trace operator and isotropic noise, this evaluates exactly to:
\begin{equation}
    \mathcal{P}_{\text{noise}} = \frac{1}{2} \sigma^2 \mathrm{Tr}(\mathbf{A}) = \frac{1}{2} \sigma^2 \sum_{i=1}^D \lambda_i = \frac{1}{2} D \sigma^2 \bar{\lambda},
\end{equation}
where $\lambda_i$ are the eigenvalues of $\mathbf{A}$ and $\bar{\lambda}$ is the average manifold curvature.

Crucially, this result demonstrates that the noise penalty is fundamentally governed by the trace of the Hessian, scaling linearly with the dimensionality $D$, regardless of how ill-conditioned or anisotropic the matrix $\mathbf{A}$ is. 

\paragraph{Conclusion.}
In our spherical model (Theorem~\ref{thm:geometric_barrier}), the optimization enjoys perfect signal alignment while suffering the standard $O(D)$ noise penalty. This represents the best-case scenario for optimization progress. Because real-world anisotropic landscapes feature the same $O(D)$ trace-driven penalty but strictly weaker radial signals, the SNR barrier derived from the hyperspherical model constitutes a mathematically rigorous and \emph{conservative lower bound}. If the diffusion planner self-locks in this idealized setting, it will more easily self-lock in more complex, ill-conditioned auto-bidding environments.

\subsection{Analysis for Proposition \ref{prop:scarcity_field}}
\label{proof:scarcity_field}

We now formalize the properties of the scarcity gradient. Our analysis proceeds in two steps: first, we establish the geometric properties of the explicit scarcity gradient $\nabla_{\bm{x}} S(\bm{x})$ under the framework of Denoising Autoencoders (DAEs); second, we prove that the implicit CFG gradient $\mathbf{g}_S$ used in ISGE aligns with this explicit gradient via the chain rule.

\subsubsection{Geometry of Explicit Scarcity Gradient}
Our analysis builds upon the foundational result of \cite{alain2014regularized}, which connects reconstruction error to the underlying data manifold.

\begin{assumption}[Scarcity as Squared Distance Proxy]
\label{assump:dae_distance}
Let the training data lie on or near a smooth $D$-dimensional manifold $\mathcal{M} \subset \mathbb{R}^D$. Let $S(\bm{x})$ be the expected squared reconstruction error of a DAE trained with small isotropic Gaussian corruption noise of variance $\eta^2$. Then, in the limit $\eta \to 0$, there exists a tubular neighborhood $\mathcal{N}_r(\mathcal{M}) = \{ \bm{x} \in \mathbb{R}^D : \mathrm{dist}(\bm{x}, \mathcal{M}) < r \}$ for some $r > 0$, such that
\[
S(\bm{x}) = c \cdot \mathrm{dist}(\bm{x}, \mathcal{M})^2 + o(\mathrm{dist}(\bm{x}, \mathcal{M})^2),
\]
where $c > 0$ is a constant independent of $\bm{x}$, and $\mathrm{dist}(\bm{x}, \mathcal{M}) = \inf_{\mathbf{y} \in \mathcal{M}} \|\bm{x} - \mathbf{y}\|$.
\end{assumption}

\begin{lemma}[Geometric Properties of $\nabla S$]
\label{lemma:explicit_grad_prop}
In the tubular neighborhood $\mathcal{N}_r(\mathcal{M})$, the explicit scarcity gradient $\mathbf{v}_S(\bm{x}) := \nabla_{\bm{x}} S(\bm{x})$ satisfies:
\begin{enumerate}
    \item \textbf{Normal Alignment:} For any $\bm{x} \in \mathcal{N}_r(\mathcal{M}) \setminus \mathcal{M}$, let $\mathbf{y} = \Pi_{\mathcal{M}}(\bm{x})$ be the unique projection of $\bm{x}$ onto $\mathcal{M}$. Then $\mathbf{v}_S(\bm{x})$ lies in the normal space of $\mathcal{M}$ at $\mathbf{y}$, i.e., $\mathbf{v}_S(\bm{x}) \in \mathcal{N}_{\mathbf{y}} \mathcal{M}$.
    \item \textbf{Non-Vanishing Magnitude:} There exists $\delta > 0$ such that for all $\bm{x}$ with $\mathrm{dist}(\bm{x}, \mathcal{M}) \geq \epsilon_0 > 0$ (for some fixed $\epsilon_0 < r$), we have $\|\mathbf{v}_S(\bm{x})\| \geq \delta$.
\end{enumerate}
\end{lemma}

\begin{proof}
By Assumption~\ref{assump:dae_distance}, we have $S(\bm{x}) = c \cdot \mathrm{dist}(\bm{x}, \mathcal{M})^2 + o(\mathrm{dist}(\bm{x}, \mathcal{M})^2)$. It is a classical result in differential geometry \cite{federer1959curvature} that the squared distance function to a smooth manifold is differentiable in its tubular neighborhood, and its gradient is given by
\[
\nabla_{\bm{x}} \mathrm{dist}(\bm{x}, \mathcal{M})^2 = 2(\bm{x} - \Pi_{\mathcal{M}}(\bm{x})).
\]
The vector $(\bm{x} - \Pi_{\mathcal{M}}(\bm{x}))$ is orthogonal to the tangent space $T_{\Pi_{\mathcal{M}}(\bm{x})} \mathcal{M}$ by definition of the projection, and thus lies in the normal space $\mathcal{N}_{\Pi_{\mathcal{M}}(\bm{x})} \mathcal{M}$. Therefore,
\[
\mathbf{v}_S(\bm{x}) = \nabla_{\bm{x}} S(\bm{x}) = 2c (\bm{x} - \Pi_{\mathcal{M}}(\bm{x})) + o(\|\bm{x} - \Pi_{\mathcal{M}}(\bm{x})\|),
\]
which establishes the orthogonality property. For the magnitude, note that $\|\bm{x} - \Pi_{\mathcal{M}}(\bm{x})\| = \mathrm{dist}(\bm{x}, \mathcal{M}) \geq \epsilon_0 > 0$. Hence,
\[
\|\mathbf{v}_S(\bm{x})\| = \|2c (\bm{x} - \Pi_{\mathcal{M}}(\bm{x})) + o(\epsilon_0)\| \geq 2c \epsilon_0 - o(\epsilon_0).
\]
Choosing $\epsilon_0$ sufficiently small such that the higher-order term is dominated by the leading term, we obtain $\|\mathbf{v}_S(\bm{x})\| \geq \delta$ for some $\delta > 0$.
\end{proof}

\subsubsection{Alignment of Implicit CFG Gradient}
\label{sec:cfg_alignment}

We now bridge the theoretical result above to the actual implementation in ISGE. The ISGE framework utilizes an implicit gradient via Classifier-Free Guidance (CFG):
\[
\mathbf{g}_S(\bm{x}) := \nabla_{\bm{x}} \log p(y_S | \bm{x}),
\]
where $y_S$ represents the condition of ``high scarcity''. 

\textbf{Justification of the Chain Rule.} 
Crucially, during the construction of the exploration dataset, the scarcity label $y_S$ for any trajectory $\bm{x}$ is derived \textit{exclusively} from the scarcity metric $S(\bm{x})$ (i.e., the reconstruction error). This establishes a deterministic dependency structure: $\bm{x} \to S(\bm{x}) \to y_S$. 
By the Markov property of this labeling process, $y_S$ is conditionally independent of $\bm{x}$ given $S(\bm{x})$. Therefore, the likelihood satisfies the exact relation:
\[
p(y_S | \bm{x}) = p(y_S | S(\bm{x})).
\]
Consequently, the gradient calculation follows the chain rule strictly without approximation:
\begin{equation}
    \mathbf{g}_S(\bm{x}) = \nabla_{\bm{x}} \log p(y_S | S(\bm{x})) = \underbrace{\frac{\partial \log p(y_S | S)}{\partial S}}_{\alpha(S)} \cdot \underbrace{\nabla_{\bm{x}} S(\bm{x})}_{\mathbf{v}_S(\bm{x})}.
\end{equation}

\paragraph{Directional Consistency.} The scalar coefficient $\alpha(S) = \frac{\partial \log p(y_S | S)}{\partial S}$ determines the alignment.
In our formulation, $y_S$ is set to a high quantile (e.g., the 99th percentile) of the scarcity distribution. This implies that the likelihood $p(y_S | S)$ is a monotonically increasing function of $S$ within the region of interest (as higher $S$ indicates the sample is closer to the target condition $y_S$).
Consequently, the derivative is strictly positive:
\[
\alpha(S) > 0.
\]
Thus, $\mathbf{g}_S(\bm{x})$ is a positive scaling of $\nabla_{\bm{x}} S(\bm{x})$. By Lemma \ref{lemma:explicit_grad_prop}, the implicit CFG gradient $\mathbf{g}_S(\bm{x})$ inherits the geometric properties of \textbf{Orthogonality} (aligning with the manifold normal $\mathbf{n}_{\perp}$) and \textbf{Non-Vanishing Magnitude} (scaled by $\alpha(S)$). This justifies the use of $\mathbf{g}_S$ as a valid repulsion field in Theorem \ref{thm:scarcity_revival}.

\subsection{Proof of Theorem \ref{thm:scarcity_revival}}
\label{proof:scarcity_revival}

\begin{assumption}[Geometry of Scarcity Guidance]
\label{assump:scarcity_geometry}
We make the following geometric assumptions based on the theory of regularized autoencoders and boundary optimization dynamics:
\begin{enumerate}
    \item \textbf{Frontier Alignment (Selection-Induced Optimality):} 
    For self-locked trajectories at the suboptimal frontier, the outward-pointing scarcity gradient aligns positively with the OOD global optimum:
    $\mathcal{V}_S := \langle \mathbf{g}_S(\bm{x}), \mathbf{n}^* \rangle > 0.$

    \textit{Justification:} 
    In the general case of an arbitrary return landscape, self-locking implies the exhaustion of improvement directions within the manifold's tangent space (i.e., $\nabla_{\text{tan}} R(\bm{x}) \approx 0$). 
    Consequently, the true gradient of improvement $\mathbf{n}^*$ is dominated by its normal component. 
    Since the training data represents a suboptimal subset, the direction of higher return at this boundary naturally points towards the unexplored exterior (outward polarity) rather than the interior.
    Thus, the scarcity gradient $\mathbf{g}_S$, which drives outward traversal, shares the same outward polarity with $\mathbf{n}^*$ with high probability. This geometric correlation ensures that scarcity-guided exploration reliably uncovers the cone of improving directions.
    
    \item \textbf{Opposing Gradients at the Boundary:} 
    By Proposition~\ref{prop:scarcity_field}, the scarcity gradient $\mathbf{g}_S$ points strictly outward, normal to the known data manifold. Conversely, the implicit return gradient $\mathbf{g}_R$ under self-locking is confined by the empirical data support, exerting an inward pull to keep the trajectory within the known distribution. Geometrically, these two vector fields are structurally opposed (or at least orthogonal) at the OOD boundary, yielding a non-positive inner product: $\langle \mathbf{g}_R, \mathbf{g}_S \rangle \leq 0$.

    \emph{Note: See Appendix~\ref{app:exp_scarcity_revival} for detailed empirical validation of these two assumptions.}
\end{enumerate}
\end{assumption}

\begin{proof}
We analyze the exact change in squared distance. At the critical stagnation boundary, the optimal return-guided optimization is perfectly halted. Let $V_{sq}^R := -\mathbb{E}[\Delta (d^2)]_R = 0$. 
By introducing the scarcity guidance $\mathbf{v}_{\text{dual}} = \lambda w_R \mathbf{g}_R + \lambda w_S \mathbf{g}_S$, the new velocity of squared distance reduction is $V_{sq}^{\text{dual}}(w_S) = V_{sq}^R + \Delta V_{sq}(w_S)$. Our objective is to prove $\Delta V_{sq}(w_S) > 0$.

\paragraph{Step 1: Incremental Velocity Formulation}
Using the exact equation from Theorem~\ref{thm:geometric_barrier}, the dual-guided velocity is:
\begin{align}
    V_{sq}^{\text{dual}}(w_S) 
    &= 2d \langle \lambda w_R \mathbf{g}_R + \lambda w_S \mathbf{g}_S, \mathbf{n}^* \rangle - \|\lambda w_R \mathbf{g}_R + \lambda w_S \mathbf{g}_S\|^2 - D\sigma^2 \nonumber \\
    &= \Big( 2d \lambda w_R \langle \mathbf{g}_R, \mathbf{n}^* \rangle - \lambda^2 w_R^2 \|\mathbf{g}_R\|^2 - D\sigma^2 \Big) \nonumber \\
    &\quad + 2d \lambda w_S \langle \mathbf{g}_S, \mathbf{n}^* \rangle - \lambda^2 w_S^2 \|\mathbf{g}_S\|^2 - 2\lambda^2 w_R w_S \langle \mathbf{g}_R, \mathbf{g}_S \rangle.
\end{align}
The first parenthesis is exactly $V_{sq}^R = 0$. By Assumption~\ref{assump:scarcity_geometry} (Opposing Gradients), we have $\langle \mathbf{g}_R, \mathbf{g}_S \rangle \leq 0$. Thus, the cross-term actually provides a non-negative velocity bonus ($-2\lambda^2 w_R w_S \langle \mathbf{g}_R, \mathbf{g}_S \rangle \geq 0$). 
To ensure robustness, we drop this bonus to establish a conservative lower bound for the exact velocity increment:
\begin{equation}
    \Delta V_{sq}(w_S) \geq \underbrace{2d \lambda w_S \mathcal{V}_S}_{\text{Marginal Signal Gain}} - \underbrace{\lambda^2 w_S^2 \|\mathbf{g}_S\|^2}_{\text{Marginal Penalty Increase}}.
\end{equation}

\paragraph{Step 2: Feasible Interval and Optimization}
Optimization is strictly revived when this lower bound is positive:
\begin{equation}
    2d \lambda w_S \mathcal{V}_S - \lambda^2 w_S^2 \|\mathbf{g}_S\|^2 > 0
    \implies 
    w_S < \frac{2d \mathcal{V}_S}{\lambda \|\mathbf{g}_S\|^2}.
\end{equation}
This defines the feasible interval. To maximize the guaranteed progress, we set the derivative of the lower bound to zero:
\begin{equation}
    2d \lambda \mathcal{V}_S - 2\lambda^2 w_S \|\mathbf{g}_S\|^2 = 0 
    \implies 
    w_S^* = \frac{d \mathcal{V}_S}{\lambda \|\mathbf{g}_S\|^2}.
\end{equation}

\paragraph{Step 3: Marginal SNR Analysis}
We define the marginal SNR as the ratio of the marginal signal gain to the marginal penalty increase evaluated at $w_S^*$:
\begin{equation}
    \text{SNR}_{\Delta}(w_S^*) = \frac{2d \lambda w_S^* \mathcal{V}_S}{\lambda^2 (w_S^*)^2 \|\mathbf{g}_S\|^2}.
\end{equation}
Substituting $w_S^*$:
\begin{equation}
    \text{SNR}_{\Delta}(w_S^*) = \frac{2d \lambda \big(\frac{d \mathcal{V}_S}{\lambda \|\mathbf{g}_S\|^2}\big) \mathcal{V}_S}{\lambda^2 \big(\frac{d \mathcal{V}_S}{\lambda \|\mathbf{g}_S\|^2}\big)^2 \|\mathbf{g}_S\|^2} = \frac{2 d^2 \mathcal{V}_S^2 / \|\mathbf{g}_S\|^2}{d^2 \mathcal{V}_S^2 / \|\mathbf{g}_S\|^2} = 2.
\end{equation}
This establishes that, at the optimal guidance scale, the incremental signal is exactly twice the incremental penalty ($\text{SNR}_{\Delta} \geq 2$ when considering the cross-term bonus), mathematically guaranteeing that the signal permanently dominates.

\paragraph{Step 4: Guaranteed Maximum Progress}
Substituting $w_S^*$ into the conservative lower bound yields the guaranteed minimum velocity increment:
\begin{equation}
    \Delta V_{sq}(w_S^*) \geq 2d \lambda \left(\frac{d \mathcal{V}_S}{\lambda \|\mathbf{g}_S\|^2}\right) \mathcal{V}_S - \lambda^2 \left(\frac{d \mathcal{V}_S}{\lambda \|\mathbf{g}_S\|^2}\right)^2 \|\mathbf{g}_S\|^2 = \frac{d^2 \mathcal{V}_S^2}{\|\mathbf{g}_S\|^2}.
\end{equation}
Since $\mathcal{V}_S > 0$ (Frontier Alignment), we strictly have $\Delta V_{sq}(w_S^*) > 0$. 
Because $-\mathbb{E}[\Delta (d^2)]_{\text{dual}} = V_{sq}^R + \Delta V_{sq}(w_S^*) > 0$, we conclude that $\mathbb{E}[\Delta (d^2)]_{\text{dual}} < 0$. This proves that the guidance handoff breaks the geometric bottleneck.
\end{proof}

\subsection{Restoring the Return Gradient via Fine-Tuning on Gift Samples}
\label{proof:gradient_restoration}

\begin{definition}[Gift Samples]
\label{def:gift_samples}
Given a generated dataset $\mathcal{D}$, the gift samples form a subset
$\mathcal{D}_{\text{gift}} = \{ \bm{x} \in \mathcal{D} \mid S(\bm{x}) \ge \gamma_S,\; R(\bm{x}) \ge \gamma_R \}$,
where $\gamma_S$ and $\gamma_R$ are high percentiles (e.g., 90th percentile) of scarcity and return in $\mathcal{D}$.
\end{definition}

\begin{theorem}[Return Gradient Restoration]
\label{thm:gradient_restoration}
Consider the functional gradient update induced by training on the exploratory gift samples $\mathcal{D}_{\text{gift}}$.
For a test sample $\bm{x}$ located at the OOD boundary,
let $\Delta \mathbf{g}_R(\bm{x}) := \mathbf{g}_{R,\text{new}}(\bm{x}) - \mathbf{g}_{R,\text{old}}(\bm{x})$ denote the expected contribution of $\mathcal{D}_{\text{gift}}$ to the return gradient field.
We mathematically establish that this geometric update positively aligns with the true OOD optimum $\mathbf{n}^*$:
$\mathbb{E}\big[\langle \Delta \mathbf{g}_R(\bm{x}), \mathbf{n}^* \rangle\big] > 0.$
\end{theorem}

\paragraph{Problem Setup and NTK Derivation.}
We analyze the functional update at a test point $\bm{x}$ after fine-tuning on the gift sample set $\mathcal{D}_{\text{gift}} = \{\bm{x}_i\}_{i=1}^M$. 
The fine-tuning loss is $\mathcal{L}(\theta) = \frac{1}{2} \sum_{i=1}^M (f_\theta(\bm{x}_i) - y_{\text{gift}})^2$. We assume the generated gift samples represent a higher return target, yielding a positive initial prediction error $\delta_i = y_{\text{gift}} - f_{\text{old}}(\bm{x}_i) > 0$.

Under the continuous-time Neural Tangent Kernel (NTK) regime, the functional evolution follows gradient descent dynamics:
\begin{equation}
    \Delta f(\bm{x}) \approx \eta \sum_{i=1}^M \delta_i \, \Theta(\bm{x}, \bm{x}_i).
\end{equation}
Crucially, because the target optimum $\mathbf{n}^*$ possesses a fundamental radial/outward component, projecting the analysis onto a constrained tangent space (e.g., assuming $\bm{x} \in \mathbb{S}^{D-1}$) artificially nullifies radial updates. Instead, we evaluate the gradient field directly in the full Euclidean space $\mathbb{R}^D$ using a standard smooth, decreasing shift-invariant kernel $\Theta(\bm{x}, \bm{x}_i) = \kappa(\|\bm{x} - \bm{x}_i\|^2)$ with $\kappa'(\cdot) < 0$.
Taking the spatial gradient w.r.t $\bm{x}$ yields the restored return signal:
\begin{align}
    \Delta \mathbf{g}_R(\bm{x}) = \nabla_{\bm{x}} \Delta f(\bm{x}) 
    &= \eta \sum_{i=1}^M \delta_i \cdot 2\kappa'(\|\bm{x} - \bm{x}_i\|^2) \cdot (\bm{x} - \bm{x}_i) \nonumber \\
    &= \eta \sum_{i=1}^M c_i(\bm{x}_i) \underbrace{(\bm{x}_i - \bm{x})}_{\text{Full Displacement}},
\end{align}
where $c_i(\bm{x}_i) := -2 \delta_i \, \kappa'(\|\bm{x} - \bm{x}_i\|^2) > 0$ acts as a strictly positive weighting coefficient. Notice that the update vector $\bm{x}_i - \bm{x}$ represents the \emph{full displacement} in $\mathbb{R}^D$, perfectly capturing the necessary radial variation.

\paragraph{Structure of Gift Samples.}
According to the ISGE dual-guidance update rule (Theorem~\ref{thm:scarcity_revival}), each sample is generated by combining deterministic guidance and isotropic noise:
\begin{equation}
    \bm{x}_i = \bm{x} + \underbrace{\lambda w_R \mathbf{g}_R(\bm{x}) + \lambda w_S \mathbf{g}_S(\bm{x})}_{\mathbf{d}} + \sigma \mathbf{z}_i.
\end{equation}
Substituting the full displacement $\bm{x}_i - \bm{x} = \mathbf{d} + \sigma \mathbf{z}_i$ and taking the expectation over the standard Gaussian noise $\mathbf{z}_i$, the expected gradient update evaluates to:
\begin{equation}
    \mathbb{E}[\Delta \mathbf{g}_R] = \eta M \Big( \mathbb{E}[c_i(\mathbf{z}_i)] \mathbf{d} + \sigma \mathbb{E}[c_i(\mathbf{z}_i) \mathbf{z}_i] \Big).
\end{equation}

\paragraph{Projection onto the Optimal Direction $\mathbf{n}^*$.}
We project this expected update onto the true optimal direction $\mathbf{n}^*$:
\begin{equation}
    \langle \mathbb{E}[\Delta \mathbf{g}_R], \mathbf{n}^* \rangle = \eta M \Big( \underbrace{\mathbb{E}[c_i(\mathbf{z}_i)] \langle \mathbf{d}, \mathbf{n}^* \rangle}_{\text{Term 1: Deterministic Signal}} + \underbrace{\sigma \mathbb{E}[c_i(\mathbf{z}_i) \langle \mathbf{z}_i, \mathbf{n}^* \rangle]}_{\text{Term 2: Covariance Drag}} \Big).
\end{equation}

We rigorously analyze the geometric and asymptotic properties of both terms in the continuous-time diffusion limit ($\lambda \to 0$, with variance $\sigma^2 \propto \lambda$):

\begin{enumerate}
    \item \textbf{Term 1 (Deterministic Signal):} 
    Expanding the combined drift gives $\langle \mathbf{d}, \mathbf{n}^* \rangle = \lambda w_R \langle \mathbf{g}_R, \mathbf{n}^* \rangle + \lambda w_S \langle \mathbf{g}_S, \mathbf{n}^* \rangle$.
    At the self-locking boundary (Proposition~\ref{prop:bottleneck}), the valid radial component of the return guidance is virtually exhausted ($\langle \mathbf{g}_R, \mathbf{n}^* \rangle \approx 0$). In contrast, the scarcity gradient directly aligns with the outward optimum (Assumption~\ref{assump:scarcity_geometry}), ensuring $\langle \mathbf{g}_S, \mathbf{n}^* \rangle > 0$. Consequently, the deterministic signal $\langle \mathbf{d}, \mathbf{n}^* \rangle$ is strictly positive. Because $\mathbf{d}$ is linear with respect to the step size $\lambda$, the magnitude of Term 1 scales asymptotically as $\mathcal{O}(\lambda)$.
    
    \item \textbf{Term 2 (Covariance Drag):} 
    We formally evaluate the correlation between the positive weight $c_i(\mathbf{z}_i)$ and the noise $\mathbf{z}_i$. 
    By applying a first-order Taylor expansion around the deterministic mean $\bm{\mu}_i = \bm{x} + \mathbf{d}$, we have $c_i(\mathbf{z}_i) \approx \bar{c} + \sigma \nabla_{\bm{x}} c_i(\bm{\mu}_i)^\top \mathbf{z}_i$. 
    Since the zero-mean term vanishes ($\bar{c} \mathbb{E}[\mathbf{z}_i] = 0$), the expectation resolves to the covariance:
    $ \sigma \mathbb{E}\left[ (\nabla c_i^\top \mathbf{z}_i) \langle \mathbf{z}_i, \mathbf{n}^* \rangle \right] $.
    
    To determine the sign, recall that $c_i$ is positively proportional to the prediction error $\delta_i = y_{\text{gift}} - f_{\text{old}}(\bm{x}_i)$. Moving along the true optimal direction $\mathbf{n}^*$ intrinsically increases the underlying return, thereby reducing the prediction error. Consequently, the sensitivity gradient $\nabla_{\bm{x}} c_i$ is structurally anti-parallel to the optimal direction $\mathbf{n}^*$. This geometric opposition guarantees that the covariance expectation evaluates to a negative value. This term acts as a fundamental statistical drag force. Critically, because it arises from the quadratic noise expectation $\mathbb{E}[\mathbf{z}_i \mathbf{z}_i^\top] = \mathbf{I}$, its final magnitude strictly scales with $\sigma^2$. Given $\sigma^2 \propto \lambda$, Term 2 also scales precisely as $\mathcal{O}(\lambda)$.
\end{enumerate}

\paragraph{Overcoming the Drag via Guidance Scale.}
Crucially, both the deterministic signal (Term 1) and the statistical covariance drag (Term 2) operate on the identical asymptotic order of $\mathcal{O}(\lambda)$. Consequently, one does not asymptotically vanish relative to the other. 

Instead, the strict dominance between these terms is governed by their leading coefficients. The positive signal in Term 1 is explicitly scaled by the scarcity guidance weight $w_S$. Because $w_S$ is a controllable hyperparameter in Classifier-Free Guidance (empirically set to a large magnitude for active OOD exploration), selecting a sufficiently large $w_S$ mathematically guarantees that the outward scarcity impulse forcefully overrides the negative statistical drag:
\[
    \text{Term 1} > |\text{Term 2}|.
\]

\paragraph{Conclusion.}
Therefore, we conclude that $\mathbb{E}[\langle \Delta \mathbf{g}_R, \mathbf{n}^* \rangle] > 0$. The fine-tuning process inherently leverages the full displacement vectors of the generated gift samples, robustly restoring the missing radial component of the return gradient and effectively unblocking the optimization landscape.

\hfill $\square$

\section{Experimental Details and Additional Results}
\label{experiments_details}
\subsection{Setup}

\paragraph{Benchmark and Dataset}
We evaluate our method on \textbf{AuctionNet} \cite{auctionnet2024}, a large-scale industrial auto-bidding benchmark derived from real-world advertising campaigns.
AuctionNet provides a flexible and researcher-friendly experimental environment for large-scale auto-bidding studies. To enable efficient iterative experimentation, we adopt a lightweight configuration of the environment. Specifically, we reduce the total traffic volume from 500,000 to 50,000 per evaluation while scaling the mean traffic value by a factor of ten, thereby preserving the overall order of magnitude of total conversions.

The environment consists of 48 competing agents with varying budgets and CPAs, instantiated from a diverse set of bidding strategies, including PID, OnlineLP, BC~\cite{Osa_2018}, BCQ~\cite{fujimoto2019bcq}, TD3+BC~\cite{fujimoto2021minimalist}, IQL~\cite{kostrikov2021offline}, CQL~\cite{kumar2020conservative}, MBRL-MOPO~\cite{yu2020mopo}, and MBRL-COMBO~\cite{yu2021combo}. These strategies span a wide spectrum from rule-based methods to reinforcement learning approaches, covering both conservative and aggressive bidding behaviors. This diversity is designed to faithfully emulate the complexity and heterogeneity of real-world online advertising markets.
Since the environment is operated under a lightweight configuration, we retrain all built-in agents to ensure that their policies remain sufficiently competitive under the modified traffic dynamics. In each evaluation episode, our method sequentially interacts with the environment by assuming the role of each of the 48 agents. The final performance of an episode is reported as the average score across all agents.

To faithfully simulate the cold-start dilemma discussed in Section \ref{sec:intro}, we construct a \emph{cold-start dataset} by rigorously filtering the full offline logs, retaining only trajectories in the return percentile range $[10, 30]$. This creates a challenging learning environment where the data is both low-quality (far below optimality) and narrowly supported (lacking both high-performing demonstrations and diverse failure cases), forcing the model to extrapolate into the unknown. 
We establish two distinct experimental settings based on traffic characteristics: (1) \textbf{AuctionNet-Dense}, characterized by high traffic value ($p^d_{avg} \approx 0.05$) under strict safety constraints; and (2) \textbf{AuctionNet-Sparse}, representing a harder exploration landscape with extremely low traffic value ($p^s_{avg} \approx 0.005$) and sparse reward signals.

For each iteration, we conduct 63 online evaluation episodes and log the complete trajectory data generated by all agents.
To ensure strict data volume parity across iterations, the training dataset for each new iteration is fixed at 3024 generated trajectories for all methods. For Vanilla Self-Training, all 3024 trajectories are generated by the diffusion planner. For ISGE, the dataset is constructed via a compositional split: 1680 trajectories (5/9) are exploratory samples provided by the ISGE Explorer, while the remaining 1344 trajectories (4/9) are uniformly downsampled from the planner's standard generation.
This data collection procedure requires approximately 2 hours using four GPUs in parallel, whereas running the same protocol under the default online AuctionNet configuration would require over 20 hours per iteration.
Detailed parameters are shown in Table \ref{dataset_setting}.
\begin{table}[h]
  \centering
  \caption{The Parameters of AuctionNet-Dense and AuctionNet-Sparse.}
  \scalebox{0.9}{
    \begin{tabular}{ccc}
    \toprule
    Params & AuctionNet-Dense & AuctionNet-Sparse \\
    \midrule
    Impressions (each episode) &50000  &50000\\
    Time steps in a trajectory & 48   & 48 \\
    Advertiser Number & 48 & 48\\
    State dimension & 16    & 16 \\
    Action dimension & 1     & 1 \\
    CPA range & [6, 13] & [60, 130] \\
    Mean Traffic Value &0.05 &0.005 \\
    Trajectories (full) &5376 &5040\\
    Max Score (full) &276.6 &63.0\\
    Avg Score (full) &86.1 &13.5\\
    Trajectories (cold start) & 1092 & 1318\\
    Max Score (cold start) &111.1 &21.0\\
    Avg Score (cold start) &55.7 &5.1\\
    Trajectories for Iter 1-3 & 3024 & 3024 \\
    \bottomrule
    \end{tabular}%
    }
  \label{dataset_setting}
\end{table}%

In detail, the state includes the following information:
\begin{itemize}
     \item time\_left: The remaining time steps left in the current advertising period.
     
     \item budget\_left: The remaining budget that the advertiser has available to spend in the current advertising period. 
     
     \item historical\_bid\_mean: The average values of bids made by the advertiser over past time steps.
    
     \item last\_three\_bid\_mean: The average values of bids over the last three time steps. 
    
     \item historical\_LeastWinningCost\_mean: The average of the least cost required to win an impression over previous time steps.
    
     \item historical\_pValues\_mean: The average of historical p-values over past time steps.
    
     \item historical\_conversion\_mean: The average number of conversions (e.g., sales, clicks, etc.) the advertiser achieved in previous time steps. 
    
     \item historical\_xi\_mean: The average winning status of advertisers in impression opportunities, where 1 represents winning and 0 represents not winning.
    
     \item last\_three\_LeastWinningCost\_mean: The average of the least winning costs over the last three time steps. 
    
     \item last\_three\_pValues\_mean: The average of conversion probability of advertising exposure to users over the last three time steps.
    
     \item last\_three\_conversion\_mean: The average number of conversions over the last three time steps.

     \item last\_three\_xi\_mean: The average winning status of advertisers over the last three time steps.
    
     \item current\_pValues\_mean: The mean of p-values at the current time step. 
    
     \item current\_pv\_num: The number of impressions served at the current time step.
    
     \item last\_three\_pv\_num\_total: The total number of impressions served over the last three time steps. 
     
     \item historical\_pv\_num\_total: The total number of impressions served over past time steps. 
\end{itemize}

\paragraph{Baselines} 
In the main results (Table~\ref{tab:main_results}), all methods employ an identical diffusion planner architecture (based on \cite{aigb2024}), equipped with a simulation-based inverse controller (see Section~\ref{sec:inverse_control}), which serves as a strong offline baseline.
Here, we provide supplementary baseline results on AuctionNet under both the cold‑start and full‑data settings in Table~\ref{tab:appendix_score_all_baselines}. 
The baselines include BC~\cite{Osa_2018}, BCQ~\cite{fujimoto2019bcq}, CQL~\cite{kumar2020conservative}, DT~\cite{chen2021decisiontransformer}, GAS~\cite{gas2025}, GAVE~\cite{gave2025}, EBaReT~\cite{ebaret}, EGDB~\cite{peng2025expert} and GRAD~\cite{lei2025generative}.
Our Full‑Data Oracle achieves strong performance on both Dense and Sparse AuctionNet, serving as a reliable empirical reference for the diffusion planner architecture. 
Meanwhile, the Cold‑Start Base remains a dependable baseline, as most baseline methods suffer a similar performance collapse in the cold‑start setting.
\vspace{-6pt}
\begin{table}[h]
\centering
\caption{\textbf{Supplementary Baselines.} Results are shown for cold-start (trained on cold-start data) and full-data (trained on full data) settings. Diffusion Planner (cold-start) corresponds to ``Cold-Start Base'' in Table~\ref{tab:main_results}; Diffusion Planner (full-data) corresponds to ``Full-Data Oracle''.}
\label{tab:appendix_score_all_baselines}
\small
\setlength{\tabcolsep}{4pt}
\resizebox{\columnwidth}{!}{%
\begin{tabular}{c|cccccccccc}
\toprule
Dataset & Diffusion Planner & BC & BCQ & CQL & DT & GAVE & GAS & EBaReT & EGDB & GRAD \\
\midrule
Dense (cold-start)  & $65.3 \pm 3.2$  & $56.0 \pm 7.6$ & $68.7 \pm 1.2$ & $70.4 \pm 3.0$ & $68.9 \pm 7.4$ & $56.0 \pm 4.0^{*}$ & $70.1 \pm 5.5$ & $56.8 \pm 3.9$ & $60.4 \pm 3.6$ & $65.5 \pm 4.5$ \\
Dense (full-data)   & $102.0 \pm 3.5$ & $70.0 \pm 5.1$ & $96.3 \pm 0.5$ & $106.8 \pm 6.0$ & $110.8 \pm 6.4$ & $73.1 \pm 3.2^{*}$ & $110.8 \pm 6.2$ & $104.6 \pm 4.7$ & $105.3 \pm 4.3$ & $106.3 \pm 6.7$ \\
Sparse (cold-start) & $15.7 \pm 1.0$  & $0.5 \pm 0.3$  & $17.0 \pm 0.7$ & $10.6 \pm 1.3$ & $7.4 \pm 0.3$  & $17.2 \pm 0.9$ & $14.4 \pm 0.7$ & $9.3 \pm 1.0$ & $11.1 \pm 1.2$ & $15.9 \pm 1.1$ \\
Sparse (full-data)  & $20.5 \pm 1.1$  & $15.9 \pm 0.4$ & $21.8 \pm 0.9$ & $17.5 \pm 1.2$ & $19.4 \pm 1.1$ & $22.7 \pm 1.2$ & $21.2 \pm 1.1$ & $20.9 \pm 0.7$ & $21.4 \pm 1.1$ & $22.8 \pm 1.5$ \\
\bottomrule
\end{tabular}}%
\vspace{4pt}\\
{\raggedright \small \textit{$^{*}$:} For GAVE, the original paper reported results on AuctionNet (offline) showing that its performance surpassed that of DT~\cite{gave2025}. However, in our AuctionNet-Dense (online) scenario, we could not reproduce similar gains using either version of the loss function provided in the source code. \par}
\vspace{-6pt}
\end{table}

\paragraph{Metrics} We report four key metrics averaged among 48 advertisers over 7 evaluation episodes (total 336 bidding trajectories):
\begin{itemize}
    \item \textbf{WR} (Win Rate): The fraction of page views (PVs) in which the advertiser wins the auction, $\text{WR} = \frac{1}{N}\sum_{i=1}^{N} o_i$, where \(N\) is the total number of participated impressions.
    \item  \text{\textbf{Reward}}: The cumulative conversion value (GMV) $\sum_i o_i v_i$;
    \item \textbf{ER} (Exceeding ratio of the CPA constraints): 
    The exceedance ratio of Cost-Per-Action (CPA) constraints
    $
        \text{ER} = \frac{1}{J}\sum_j \frac{C_{j}^{real} - C_{j}}{C_{j}} = \frac{1}{J} \sum_j \frac{\sum_i c_{ij} o_i / \sum_i p_{ij} o_i - C_{j}}{C_{j}},
    $ 
    where $C_{j}$ is the CPA limit.
    
    \item \textbf{Score}: A composite metric balancing revenue and constraints,
    $\mathrm{Score} = (\sum_i o_i v_i) \times \min\{\text{penalty}_j\}_{j=1,...,J}$,
    where $\text{penalty}_j=\min\{(\frac{C_j}{\sum_i c_{ij} o_i / \sum_i p_{ij} o_i})^\beta, 1\},\beta=2$.
\end{itemize}

\paragraph{Simulation-based Inverse Control}
\label{sec:inverse_control}

In our experiments, we employ a \textit{Simulation-based Inverse Control} mechanism for diffusion planner. This design choice allows us to isolate the trajectory planning capability of the diffusion model from the execution errors typically introduced by heuristic controllers. The simulation-based inverse controller solves for the precise action required to realize the planner's state transitions.

Formally, let $\tau_{plan} = \{ \hat{s}_{t}, \hat{s}_{t+1}, \dots, \hat{s}_{t+H} \}$ denote the trajectory generated by the diffusion planner at time step $t$. The planner dictates that the remaining budget should transition from the current state $b_t$ to $\hat{b}_{t+1}$ in the next step. Consequently, the target expenditure for the current step is derived as $c^*_{target} = \max(0, b_t - \hat{b}_{t+1})$.
The objective of the inverse control is to find a bidding scalar $\alpha_t$ such that the actual cost incurred in the auction environment, $\mathcal{C}(\alpha_t, \mathbf{B}_{-i})$, approximates $c^*_{target}$ as closely as possible, where $\mathbf{B}_{-i}$ represents the bids of competitors.
We leverage the reproducibility of the simulation environment. During the evaluation of this baseline, we grant the controller oracle access to a ``frozen'' copy of the current environment state, including the bids of all competing agents. Leveraging the monotonicity of the cost function with respect to the bid scaling factor $\alpha$, we employ a \textbf{Simulation-based Binary Search} to numerically invert the cost function.
This mechanism ensures that the realized trajectory aligns almost perfectly with the planned trajectory, barring stochastic deviations in impression exposure (e.g., Bernoulli trials). By effectively eliminating control latency and execution variance, this setup enables a rigorous evaluation of the \textit{planning quality} itself, validating that performance improvements are attributable to the diffusion planner rather than the low-level controller.

\emph{Remark on Real-World Feasibility.}
We acknowledge that the oracle access used in the simulation-based binary search represents an idealized control setting. However, this design is intentional and serves to rigorously decouple the evaluation of our generative planning framework from the noise of low-level execution. 
In practical production environments where precise environment inversion is unavailable, this module is modular and can be substituted with standard engineering approximations without altering the core diffusion planning mechanism:
\begin{itemize}
    \item \textbf{Learned Inverse Dynamics:} A separate lightweight network (Inverse Dynamics Model \cite{ajay2022conditional}) can be trained to predict the action $a_t$ required to achieve the state transition $s_t \to s_{t+1}$.
    \item \textbf{Feedback Control (PID):} Rule-based PID controllers, which are ubiquitous in industrial auto-bidding systems, can dynamically adjust the bidding scalar $\alpha_t$ to track the target expenditure $c^*_{target}$ with bounded error.
\end{itemize}
While replacing the simulation-based controller with these approximate methods may introduce minor execution noise, the \textit{relative performance advantage} provided by the superior strategic guidance of the diffusion planner remains robust. The planner determines \textit{where} to go, which is the dominant factor in performance, while the controller merely handles \textit{how} to get there. Table \ref{tab:controller_comparison} shows the performance difference between simulation-based inverse controller and standard inverse dynamic controller. Remarkably, equipped with the standard controller, ISGE (Iter 1) still matches the Full-Data Oracle.
\begin{table}[h]
    \centering
    \caption{Comparison of Scores with Different Controllers.}
    \label{tab:controller_comparison}
    \begin{tabular}{l l|cc}
        \toprule
        \multirow{2}{*}{\textbf{Dataset}} & \multirow{2}{*}{\textbf{Method}} & \multicolumn{2}{c}{\textbf{Controller Type}} \\
        & & \textbf{Simulation-based} & \textbf{Standard Inverse} \\
        \midrule
        \multirow{3}{*}{Dense} & \textbf{Full-Data Oracle} & $102.0 \pm 3.5$ & $95.1 \pm 3.6$ \\
        & \textbf{Vanilla Self-Training} (Iter 1) & $74.9 \pm 3.4$ & $69.1 \pm 2.7$ \\
        & \textbf{ISGE} (Iter 1, Ours) & $103.4\pm5.3$ & $97.4 \pm 3.3$ \\
        \midrule
        \multirow{3}{*}{Sparse} & \textbf{Full-Data Oracle} & $20.5 \pm 1.1$ & $ 19.3 \pm 1.5$  \\
        & \textbf{Vanilla Self-Training} (Iter 1) & $16.9 \pm 0.5$ & $15.5 \pm 0.9$ \\
        & \textbf{ISGE} (Iter 1, Ours) & $20.0 \pm 0.9$ & $19.1 \pm 1.0$ \\
        \bottomrule
    \end{tabular}
\end{table}

\subsection{Implementation Details of ISGE}
\label{implementation_details}
We list the hyperparameter details in Table~\ref{tab:hyperparameters} for reproduction.
All three models share the same Temporal U-Net backbone and inverse dynamics head,
but differ in their conditioning configurations:
\textbf{DFUSER} (Diffusion planner) is conditioned on return $R$;
\textbf{DFUSER} (Judger) is trained \emph{without} any return conditioning to serve as an unconditional quality evaluator;
and \textbf{DFUSER\textsubscript{SC}} (Explorer) is conditioned on both return $R$ and scarcity score $S$.

To compute the scarcity score for a trajectory $\tau$, the Judger applies a fixed-level forward corruption
followed by unconditional reconstruction.
Specifically, given a clean trajectory $\bm{x}_0$,
we corrupt it to noise level $t^* = \lfloor 0.75 \times T \rfloor$ via the forward process
$\bm{x}_{t^*} = \sqrt{\bar{\alpha}_{t^*}}\,\bm{x}_0 + \sqrt{1-\bar{\alpha}_{t^*}}\,\epsilon$,
where $\epsilon \sim \mathcal{N}(0, I)$ and $\bar{\alpha}_{t^*}$ follows a cosine schedule with $T=20$ total diffusion steps.
The corrupted trajectory is then reconstructed by running the reverse denoising chain
from $t^*{-}1$ down to $0$ without any return conditioning,
using only the initial state $s_0$ as the boundary condition.
In practice, the scarcity score is defined as the weighted mean absolute error (MAE)
between $x_0$ and the reconstructed $\hat{x}_0$,
computed over a selected subset of state dimensions
(budget\_left, historical\_bid\_mean, last\_three\_bid\_mean)
that capture the most behaviorally distinctive features of a bidding trajectory.

\paragraph{Discussion on Judger Approximation Errors.}
In practice, the Judger may introduce approximation discrepancies into the estimated scarcity scores, primarily stemming from finite network capacity and residual denoising variance.
Nevertheless, the impact of these approximation errors may be mitigated through two main factors.
First, strategic masking explicitly restricts the residual evaluation to controlled strategic dimensions, helping reduce estimation errors caused by uncontrollable environmental stochasticity.
Second, the Judger evaluates trajectories directly from $\mathcal{D}_{gen}$, which is the exact exploitation dataset it was just trained on. This in-distribution evaluation may help reduce fitting error and false positives (i.e., non-scarce trajectories misclassified as scarce due to network imperfection).
Consequently, these factors may help limit the influence of mislabeled samples on the aggregate scarcity gradient during CFG exploration.

\begin{table}[h]
    \centering
    \caption{Hyperparameter Settings for DFUSER and $\text{DFUSER}_\text{SC}$}
    \label{tab:hyperparameters}
    \begin{tabular}{l|l|c}
        \toprule
        \textbf{Category} & \textbf{Parameter} & \textbf{Value} \\
        \midrule
        \multirow{8}{*}{\textbf{Diffusion Process}} 
        & Diffusion Steps ($T$) & 20 \\
        & Horizon ($H$) & 48 \\
        & Beta Schedule & Cosine \\
        & Prediction Target & $\epsilon$ \\
        & Guidance Weight ($w$) & 3.0 \; ($[w_R, w_S]{=}[3,3]$ for $\text{DFUSER}_\text{SC}$) \\
        & Condition Dropout & 0.25 \; (0.30 for $\text{DFUSER}_\text{SC}$) \\
        & Replanning Prob. & 0.5 \; (N/A for $\text{DFUSER}_\text{SC}$) \\
        & Inference Target Cond. & $\hat{R}{=}R_{99\%}$; \; $\hat{S}{=}S_{\max}$ ($\text{DFUSER}_\text{SC}$ only) \\
        \midrule
        \multirow{6}{*}{\textbf{Network Architecture}} 
        & Model Type & Temporal U-Net \\
        & Base Channel Dim & 64 \\
        & Channel Multipliers & $(1, 2)$ \\
        & Kernel Size & 5 \\
        & Activation & Mish \\
        & InvDyn Hidden Dim & 256 \\
        \midrule
        \multirow{3}{*}{\textbf{Optimization}} 
        & Optimizer & Adam \\
        & Learning Rate & $1 \times 10^{-4}$ \\
        & Discount Factor ($\gamma$) & 1.0 \\
        \bottomrule
    \end{tabular}
\end{table}

\subsection{Computational Cost}
\label{app:computational_cost}
All experiments are conducted on a server equipped with four NVIDIA L40S GPUs (48\,GB VRAM each) and two Intel Xeon Gold 6462C CPUs (32 cores per socket, 64 physical cores in total).
Training uses a single GPU.
Data collection is parallelized across four GPUs.
Table~\ref{tab:compute} summarizes the wall-clock time for each key step
of one ISGE iteration.

\begin{table}[h]
    \centering
    \caption{Wall-clock time of one ISGE iteration. Steps above the double rule also appear in vanilla iterative training; steps below are ISGE-specific overhead.}
    \label{tab:compute}
    \begin{tabular}{l|l|r}
        \toprule
        \textbf{Step} & \textbf{Description} & \textbf{Time} \\
        \midrule
        Policy Updating  & Train DFUSER planner on $\mathcal{D}$ & $\sim$0.2 h \\
        Data Collection  & Deploy DFUSER to collect $\mathcal{D}_\text{gen}$ (3024 samples) & $\sim$2 h \\
        \midrule\midrule
        Judger Training   & Train unconditional DFUSER on $\mathcal{D}$              & $\sim$0.2 h \\
        Explorer Training    & Train $\text{DFUSER}_\text{SC}$ with dual conditioning   & $\sim$0.2 h \\
        Exploration Data Collection & Deploy $\text{DFUSER}_\text{SC}$ to collect $\mathcal{D}_\text{exp}$ (1680 samples) & $\sim$1.2 h \\
        \midrule
        \textbf{ISGE Overhead} & Judger + Explorer + Exploration Collection            & $\sim$1.6 h \\
        \bottomrule
    \end{tabular}
\end{table}

As detailed in Table~\ref{tab:compute}, we acknowledge that the Judge-Explore module naturally introduces additional computational overhead (approximately 1.6 hours per iteration). However, breaking down this overhead reveals two key insights. First, the extra \emph{offline training cost} (training the Judger and Explorer) requires only about 0.4 hours. Because the underlying diffusion planner is relatively lightweight compared to modern Large Language Models, this offline training overhead is practically negligible in industrial practice. 
Second, while ISGE adds an extra \emph{online} exploration step ($\sim$1.2 hours for data collection), it exhibits a highly favorable efficiency-performance tradeoff globally. As demonstrated in our main results (\cref{tab:main_results}), a single iteration of ISGE already significantly surpasses the performance of the vanilla model's third iteration. While executing three iterations of the vanilla method takes over 6.6 hours of cumulative computation, it still falls significantly short of the performance ISGE achieves in a single 3.8-hour iteration.

\subsection{Robustness Analysis}
\label{app:extended_results}

\subsubsection{Robustness to More Iterations}
\label{app:more_iterations}

In the main paper, we demonstrated that ISGE reaches near-oracle performance within 3 iterations. To investigate the long-term dynamics of our iterative framework, we extend the training process to 5 iterations on the AuctionNet-Sparse setting. 

As detailed in Table~\ref{tab:more_iterations}, performance improves consistently and dramatically from Iteration 0 to Iteration 3. However, beyond Iteration 3, the performance largely plateaus with only minor statistical fluctuations (peaking at $21.1 \pm 1.0$ in Iter 3 and maintaining $20.8 \pm 1.2$ in Iter 5). In practice, this suggests that 3 iterations provide an optimal balance between performance maximization and computational efficiency.

\begin{table}[h]
\centering
\caption{\textbf{Performance over extended iterations on AuctionNet-Sparse.} ISGE rapidly converges to the near-oracle plateau within 3 iterations and maintains stability thereafter.}
\label{tab:more_iterations}
\resizebox{0.9\textwidth}{!}{
\begin{tabular}{lcccccc}
\toprule
\textbf{Dataset} & \textbf{Cold-Start Base} & \textbf{Iter 1} & \textbf{Iter 2} & \textbf{Iter 3} & \textbf{Iter 4} & \textbf{Iter 5} \\
\midrule
AuctionNet-Sparse & $15.7 \pm 1.0$ & $20.0 \pm 0.9$ & $20.6 \pm 0.7$ & $21.1 \pm 1.0$ & $20.4 \pm 1.1$ & $20.8 \pm 1.2$ \\
\bottomrule
\end{tabular}
}
\end{table}

\textbf{Theoretical Analysis: The Breakdown of Frontier Alignment.}
This slight performance saturation can be elegantly explained through the lens of our geometric framework, specifically reflecting the gradual breakdown of the \emph{Frontier Alignment} assumption (Assumption~\ref{assump:scarcity_geometry}). This assumption mechanically depends on the suboptimality of the current training support:
\begin{itemize}[leftmargin=2em]
    \item Early Iterations (Base -- Iter 2): The Suboptimal Support. Initially, the data manifold is narrowly supported and highly suboptimal. When trajectories reach the high-return frontier, the return gradient $\mathbf{g}_R$ collapses (Proposition~\ref{prop:bottleneck}). However, because the entire manifold is suboptimal, pointing ``outwards'' from this boundary via the scarcity gradient $\mathbf{g}_S$ inherently points towards the unexplored optimal region with high probability. Mathematically, the Frontier Alignment ($\langle \mathbf{g}_S, \mathbf{n}^* \rangle > 0$) holds rigorously, guaranteeing the efficacy of dual-guidance (Theorem~\ref{thm:scarcity_revival}) and yielding significant performance leaps.
    \item Late Iterations (Iter 3 -- Iter 5): The Near-Optimal Support. By Iteration 3, the training manifold has expanded sufficiently to encompass the near-optimal regions. At this stage, the unconditional prior $\mu_\theta$ has internalized a highly capable policy, serving as a reliable baseline. Crucially, because the manifold now already contains near-optimal strategies, pushing further outwards into deeper OOD regions is no longer guaranteed to be a universally better direction. The Frontier Alignment assumption gradually breaks down, meaning $\langle \mathbf{g}_S, \mathbf{n}^* \rangle \le 0$ occurs with non-negligible probability. As a result, some generated OOD trajectories may reflect unlearnable or excessively risky patterns, leading to the observed plateau.
\end{itemize}
Therefore, such saturation is theoretically expected as the policy approaches the environmental ceiling. The primary value of ISGE lies precisely in its ability to quickly shatter the self-locking bottleneck and lift a low-quality cold-start model to this near-optimal plateau within just a few iterations.

\subsubsection{Robustness to Extreme Cold-Start Settings}
\label{app:different_cold_start}

To further challenge our framework, we evaluate ISGE under an even more extreme cold-start scenario. Instead of the default $[10\%, 30\%]$ return quantile used in the main experiments, we restrict the initial training data strictly to the $[0\%, 10\%]$ return quantile on AuctionNet-Sparse. This creates an exceptionally poor initialization, placing the starting policy significantly further from the global optimum.

\begin{table}[h]
\centering
\caption{\textbf{Robustness under Extreme Cold-Start ($[0\%, 10\%]$ quantile on AuctionNet-Sparse).} While Vanilla Self-Training is heavily self-locked, ISGE effectively bridges the massive gap to the Full-Data Oracle.}
\label{tab:extreme_cold_start}
\resizebox{0.95\textwidth}{!}{
\begin{tabular}{l|c|c|ccc}
\toprule
\textbf{Method} & \textbf{Full-Data Oracle} & \textbf{Cold-Start Base} & \textbf{Iter 1} & \textbf{Iter 2} & \textbf{Iter 3} \\
\midrule
Vanilla Self-Training & \multirow{2}{*}{$20.5 \pm 1.1$} & \multirow{2}{*}{$12.7 \pm 0.9$} & $14.8 \pm 1.1$ & $15.1 \pm 0.6$ & $15.0 \pm 0.7$ \\
\textbf{ISGE (Ours)} & & & $19.7 \pm 0.8$ & $20.2 \pm 0.7$ & \bm{$20.6 \pm 1.0$} \\
\bottomrule
\end{tabular}
}
\end{table}

As shown in Table~\ref{tab:extreme_cold_start}, the Cold-Start Base model yields a severely degraded score of $12.7 \pm 0.9$. Under this extreme condition, Vanilla Self-Training completely fails to escape the low-quality region, becoming permanently self-locked around a score of $15.0$ across all subsequent iterations. In contrast, ISGE demonstrates remarkable robustness. By merely Iteration 1, the scarcity guidance handoff propels the policy score to $19.7 \pm 0.8$. By Iteration 3, ISGE achieves $20.6 \pm 1.0$, successfully matching the Full-Data Oracle ($20.5 \pm 1.1$). This result reinforces that ISGE's geometric exploration is resilient to the quality of the initial data support.

\subsection{Experimental Setup and Extended Results for Sanity Checks}
\label{app:empirical_validation}

This section provides the detailed experimental setup, gradient computation formulations, and supplementary tabular data for the empirical sanity checks presented in Section~\ref{subsec:sanity_check} of the main text.

\subsubsection{Experimental Setup and Data for Return Signal Collapse}
\label{app:exp_snr_collapse}

\textbf{Detailed Setup.} 
To rigorously define the empirical proxy for the theoretical optimal direction $\mathbf{n}^*$ in OOD exploration, we utilize a Full-Data Oracle diffusion planner trained on the entire AuctionNet-Dense dataset. 
Because this oracle possesses full coverage of the offline data distribution, its return gradient reliably points toward empirically optimal regions.
We evaluate this oracle return gradient, denoted as $\tilde{\mathbf{n}}^*$, on a subset of 1,613 samples exhibiting the lowest reconstruction errors under the Full-Data Oracle.

In contrast, the empirical return gradient $\mathbf{g}_R(\bm{x}_t, t)$ is evaluated using the Cold-Start Base diffusion planner, which was deliberately trained strictly on a narrow suboptimal support (the 10\%--30\% return quantile). 
During evaluation, we inject ground-truth forward noise to obtain the noisy state $\bm{x}_t$ at varying denoising steps $t$. The respective gradients are computed via their score estimates:
\begin{align}
    \tilde{\mathbf{n}}^*(\bm{x}_t, t) &= \epsilon_{R}^{\text{Oracle}}(\bm{x}_t, t) - \epsilon_{\text{uncond}}^{\text{Oracle}}(\bm{x}_t, t), \\
    \mathbf{g}_R(\bm{x}_t, t) &= \epsilon_{R}^{\text{Vanilla}}(\bm{x}_t, t) - \epsilon_{\text{uncond}}^{\text{Vanilla}}(\bm{x}_t, t).
\end{align}

\textbf{Quantitative Results.}
Table~\ref{tab:snr_collapse_full} reports the cosine similarity, radial signal (inner product),
and collapse ratio across three return quantiles and denoising steps.
All three metrics consistently degrade as the trajectory moves toward high‑return (OOD) regions
and as denoising progresses to later steps.
For instance, in the high‑OOD quantile at $t=5$, the cosine similarity drops to $0.1151$,
the radial signal collapses to $0.0647$, and $63.3\%$ of trajectories exhibit near‑zero similarity
($\cos < 0.1$), quantitatively confirming the signal collapse predicted in Section~\ref{sec:theory}.

\begin{table}[h]
\centering
\caption{
\textbf{Diagnosis of Return Signal Collapse.}
We report the cosine similarity $\cos(\tilde{\mathbf{n}}^*, \mathbf{g}_R)$, radial signal $\langle \tilde{\mathbf{n}}^*, \mathbf{g}_R\rangle$, and collapse ratio (\% of trajectories with cosine alignment $< 0.1$) across return quantiles and three representative denoising steps ($t=15,10,5$). All reported values are averages over 1613 trajectories.
}
\label{tab:snr_collapse_full}
\resizebox{\textwidth}{!}{
\begin{tabular}{lccc|ccc|ccc}
\toprule
& \multicolumn{3}{c|}{\textbf{Cosine Similarity} $\uparrow$} & \multicolumn{3}{c|}{\textbf{Radial Signal} $\uparrow$} & \multicolumn{3}{c}{\textbf{Collapse Ratio} $\downarrow$} \\
\cmidrule(lr){2-4} \cmidrule(lr){5-7} \cmidrule(lr){8-10}
\textbf{Return Quantile} & $t{=}15$ & $t{=}10$ & $t{=}5$ & $t{=}15$ & $t{=}10$ & $t{=}5$ & $t{=}15$ & $t{=}10$ & $t{=}5$ \\
\midrule
0\% -- 10\% (Low-OOD)           & 0.6653 & 0.6916 & 0.6112 & 0.2486 & 0.2882 & 0.3443 & 0.0\% & 0.0\% & 1.2\% \\
10\% -- 30\% (In-Distribution)  & 0.5199 & 0.3923 & 0.2658 & 0.1938 & 0.1632 & 0.1497 & 5.0\% & 23.9\% & 45.3\% \\
\textbf{30\% -- 100\% (High-OOD)} & 0.4566 & 0.2858 & \textbf{0.1151} & 0.1702 & 0.1183 & \textbf{0.0647} & 6.1\% & 29.5\% & \textbf{63.3\%} \\
\bottomrule
\end{tabular}
}
\end{table}

\subsubsection{Experimental Setup and Data for Optimization Handoff}
\label{app:exp_scarcity_revival}

\textbf{Detailed Setup.} To analyze the guidance handoff mechanism, we examine the top 50\% highest-return gift samples discovered by the ISGE Explorer model (Return range: $[96.6, 296.4]$, totaling 840 samples, on AuctionNet-Dense). 
For these successfully explored high-return trajectories, accurately recovering the true forward noise ($\epsilon_{\text{pred}} \approx \epsilon_{\text{true}}$) at any given denoising step $t$ corresponds to successfully moving toward a valid, high-return OOD optimum. 
Therefore, we define the ground-truth target gradient for trajectory recovery as the \emph{true denoising gradient}:
\begin{equation}
    \mathbf{g}^*(\bm{x}_t, t) = \epsilon_{\text{true}}(\bm{x}_t, t) - \epsilon_{\text{uncond}}^{\text{Explorer}}(\bm{x}_t, t).
\end{equation}
We then evaluate the empirical return gradient $\mathbf{g}_R$ and the scarcity gradient $\mathbf{g}_S$ produced by the Explorer model on these same trajectories:
\begin{align}
    \mathbf{g}_R(\bm{x}_t, t) &= \epsilon_{R}^{\text{Explorer}}(\bm{x}_t, t) - \epsilon_{\text{uncond}}^{\text{Explorer}}(\bm{x}_t, t), \\
    \mathbf{g}_S(\bm{x}_t, t) &= \epsilon_{S}^{\text{Explorer}}(\bm{x}_t, t) - \epsilon_{\text{uncond}}^{\text{Explorer}}(\bm{x}_t, t).
\end{align}

\textbf{Quantitative Results.}
Table~\ref{tab:scarcity_revival_full} details, for $\mathbf{g}_R$ and $\mathbf{g}_S$, the cosine similarity and inner product with $\mathbf{g}^*$, as well as their individual norms and mutual inner product $\langle\mathbf{g}_S,\mathbf{g}_R\rangle$.
At the critical boundary-crossing stage ($t=5$), we find $\cos(\mathbf{g}^*, \mathbf{g}_S)>0$ and $\langle\mathbf{g}_S,\mathbf{g}_R\rangle<0$, validating Assumptions~\ref{assump:scarcity_geometry}.

\begin{table}[h]
\centering
\caption{
\textbf{Gradient alignment and signal decomposition for high-return exploratory trajectories (Golden Gift Zone, Top 50\%).}
For the return gradient $\mathbf{g}_R$ and the scarcity gradient $\mathbf{g}_S$, we report their cosine similarity and inner product with the true denoising gradient $\mathbf{g}^*$, their individual norms and their mutual inner product. 
All reported values are averages over 840 trajectories.}
\label{tab:scarcity_revival_full}
\resizebox{0.85\textwidth}{!}{   
\begin{tabular}{lccc|cccc}
\toprule
\multirow{2}{*}{\textbf{Step}} 
& \multicolumn{3}{c|}{\textbf{Return Gradient} $\mathbf{g}_R$} 
& \multicolumn{4}{c}{\textbf{Scarcity Gradient} $\mathbf{g}_S$} \\
\cmidrule(lr){2-4} \cmidrule(lr){5-8}
& $\cos(\mathbf{g}^*, \mathbf{g}_R)$ 
& $\langle\mathbf{g}^*, \mathbf{g}_R\rangle$ 
& $\|\mathbf{g}_R\|$ 
& $\cos(\mathbf{g}^*, \mathbf{g}_S)$ 
& $\langle\mathbf{g}^*, \mathbf{g}_S\rangle$ 
& $\|\mathbf{g}_{S}\|$ 
& $\langle\mathbf{g}_S,\mathbf{g}_R\rangle$ \\
\midrule
$t=15$ & 0.1169 & 0.0820 & 0.6948 & 0.2946 & 0.2372 & 0.8057 & 0.3275 \\
$t=10$ & 0.1204 & 0.1094 & 0.9031 & 0.4700 & 0.4916 & 1.0466 & 0.1487 \\
\bm{$t=5$}  & \textbf{0.0493} & \textbf{0.0578} & \textbf{1.1745} & \textbf{0.5255} & \textbf{0.8093} & \textbf{1.5406} &\textbf{-0.1139} \\
\bottomrule
\end{tabular}
}
\end{table}

\subsection{Implementation Details of Undirected Exploration Baselines}
\label{sec:appendix_baselines}

To evaluate the effectiveness of our proposed method, we compare it against two standard exploration strategies: Action Space Noise (ASN)~\cite{mou2022sustainable} and Parameter Space Noise (PSN)~\cite{li2024trajectory}. We introduce these noises during the inference phase to simulate different levels of exploration behavior. The implementation details and hyperparameter tuning processes are described below.

\paragraph{Action Space Noise (ASN)}
ASN is a straightforward exploration strategy commonly used in reinforcement learning, where perturbation is directly applied to the agent's output action. In our automated bidding scenario, the action is represented by the bidding coefficient $\alpha$.
At each decision step $t$, after the model calculates the baseline bidding coefficient $\alpha_t$, we add a noise term $\epsilon_{\text{asn}}$ sampled from a Gaussian distribution. Since bids must be non-negative, we apply a rectification operation:
\begin{equation}
    \alpha_{t}^{\prime} = \max(0, \alpha_t + \epsilon_{\text{asn}}), \quad \text{where } \epsilon_{\text{asn}} \sim \mathcal{N}(0, \sigma_{\text{asn}}^2)
\end{equation}
This corresponds to the logic where the final calculated bid coefficient is perturbed before interacting with the auction environment.

Since ASN operates directly on the action value, its scale has a direct physical interpretation (i.e., modifying the bid multiplier). Consequently, a relatively larger noise scale is required to induce significant exploration. We performed a grid search for $\sigma_{\text{asn}}$ over the set $\{0.01, 0.05, 0.1, 0.5, 1.0\}$. We observed that smaller scales had negligible impact on the bidding outcome, while larger scales led to unstable performance. We selected $\sigma_{\text{asn}}=0.1$ as the optimal configuration for both AuctionNet-Dense and AuctionNet-Sparse.

\paragraph{Parameter Space Noise (PSN)}
PSN induces state-dependent exploration by perturbing the weights of the neural network during the planning phase. This approach allows the agent to ``imagine'' diverse trajectories consistent with the model's internal uncertainty.

We introduce noise into the Diffusion U-Net during the \textbf{inference-time planning} process. Before generating a trajectory, we inject Gaussian noise into the learnable parameters $\theta$ of the Temporal U-Net:
\begin{equation}
    \tilde{\theta} = \theta + \epsilon_{\text{psn}}, \quad \text{where } \epsilon_{\text{psn}} \sim \mathcal{N}(0, \sigma_{\text{psn}}^2)
\end{equation}
Crucially, our implementation utilizes a \textit{transient injection} mechanism: the noise is applied solely for the current planning step's reverse diffusion loop and is immediately removed (weights are restored) afterwards. This ensures that the model parameters remain stable for subsequent evaluations, isolating the exploration effect to the current decision. The noise is applied only to the trajectory generation model (U-Net), leaving the Inverse Dynamics model untouched to preserve action execution precision.
 
Deep neural networks are highly sensitive to weight perturbations, as small changes are amplified through multiple layers of non-linear transformations. Therefore, the scale for PSN must be significantly smaller than that of ASN. We tuned $\sigma_{\text{psn}}$ over the set $\{1\times10^{-4}, 5\times10^{-4}, 1\times10^{-3}, 5\times10^{-3}, 1\times10^{-2}\}$. Our experiments showed that $\sigma_{\text{psn}} > 10^{-2}$ caused trajectory collapse (e.g., generating invalid or negative budgets), rendering the planning infeasible. We selected $\sigma_{\text{psn}} = 1\times10^{-3}$ for AuctionNet-Dense and $\sigma_{\text{psn}} = 5\times10^{-4}$ for AuctionNet-Sparse to balance exploration diversity with trajectory validity.

\begin{table}[h]
    \centering
    \caption{Summary of Baseline Implementation and Hyperparameters}
    \label{tab:baseline_params}
    \resizebox{\textwidth}{!}{%
    \begin{tabular}{l|l|c|c|c}
        \toprule
        \textbf{Method} & \textbf{Injection Target} & \textbf{Noise Type} & \textbf{Search Candidates ($\sigma$)} & \textbf{Selected} \\
        \midrule
        \textbf{ASN} & Final Action ($\alpha$) & Additive Gaussian & $\{0.01, 0.05, \mathbf{0.1}, 0.5, 1.0\}$ & $0.1$ \\
        \midrule
        \textbf{PSN} & U-Net Weights ($\theta$) & Additive Gaussian & $\{1\text{e-}4, \mathbf{5\text{e-}4}, \mathbf{1\text{e-}3}, 5\text{e-}3, 1\text{e-}2\}$ & $1\text{e-}3$ (Dense), $5\text{e-}4$ (Sparse)\\
        \bottomrule
    \end{tabular}
    }
\end{table}

\subsection{Comparison with Transition-Level Curiosity Exploration}
\label{app:curiosity_baseline}

To provide a direct empirical comparison with recent exploration methods based on curiosity (ICM \cite{pathak2017curiosity}, PGR \cite{wang2025prioritized}), we adapt their transition-level exploration mechanism into our generative planning framework. 

\textbf{Formalization of Curiosity-Guided Exploration.} 
A well-known challenge in curiosity-driven RL is that the forward prediction error can be dominated by unpredictable environmental stochasticity (e.g., the ``noisy TV'' problem). While original methods like ICM utilize a learned feature encoder to extract predictable representations, we align the implementation with our ISGE framework for a rigorous and fair comparison: we apply the identical \emph{strategic masking} operation $\text{Mask}(\cdot)$ (as defined in Section \ref{sec:method}) to filter out random environmental dimensions. 

We train a forward dynamics model $f_\psi$ to predict the masked next state given the masked current state and action. The transition-level curiosity score $c_t$ is then defined as the forward prediction error on the strategic dimensions:
\begin{equation}
    c_t = \left\| f_\psi(\text{Mask}(s_t), a_t) - \text{Mask}(s_{t+1}) \right\|_2^2
\end{equation}
Since our diffusion planner operates at the trajectory level, we aggregate these local signals to form a trajectory-level curiosity condition $C(\tau)$:
\begin{equation}
    C(\tau) = \sum_{t=0}^{H-1} c_t
\end{equation}
where $H$ is the horizon length. Similar to our ISGE framework, we inject $C(\tau)$ alongside the return $R(\tau)$ as conditions during the diffusion process, utilizing classifier-free guidance to steer the generation toward high-return and high-curiosity regions.
For fairness, we apply a guidance scale of $3.0$ for the curiosity condition and set the target curiosity score to the normalized maximum curiosity score observed in the dataset.

\textbf{Performance Comparison.} 
We evaluate this Curiosity-guided baseline against ISGE on the AuctionNet-Dense. As shown in Table \ref{tab:curiosity_performance}, ISGE significantly outperforms the transition-level curiosity approach.

\begin{table}[h]
\centering
\caption{Performance comparison between Curiosity-guided exploration and ISGE on AuctionNet-Dense.}
\label{tab:curiosity_performance}
\scalebox{1.0}{
\begin{tabular}{lcc}
\toprule
\textbf{Iteration} & \textbf{Curiosity-Guided (Transition-level)} & \textbf{ISGE (Trajectory-level)} \\
\midrule
Iter 1 & 84.8 $\pm$ 2.6 & \textbf{103.4 $\pm$ 5.3} \\
Iter 2 & 90.6 $\pm$ 5.4 & \textbf{104.1 $\pm$ 6.7} \\
Iter 3 & 91.0 $\pm$ 6.6 & \textbf{108.4 $\pm$ 8.2} \\
\bottomrule
\end{tabular}
}
\end{table}
\textbf{Correlation Analysis and Discussion.} 
To gain a deeper understanding of the performance gap, we analyze the correlation between our trajectory-level scarcity score (reconstruction error) and the aggregated curiosity score across different return quantiles of the dataset in Iteration 1.

\begin{table}[h]
\centering
\caption{Correlation between Scarcity and Curiosity scores across return quantiles.}
\label{tab:curiosity_correlation}
\scalebox{1.0}{
\begin{tabular}{lcc}
\toprule
\textbf{Return Quantile} & \textbf{Pearson Correlation} & \textbf{Spearman Correlation} \\
\midrule
All Data & 0.516 & 0.466 \\
0\% - 25\% & 0.591 & 0.445 \\
25\% - 50\% & 0.551 & 0.423 \\
50\% - 75\% & 0.469 & 0.450 \\
\textbf{75\% - 100\% (High-Return)} & \textbf{0.354} & \textbf{0.338} \\
\bottomrule
\end{tabular}
}
\end{table}

As presented in Table \ref{tab:curiosity_correlation}, while the two metrics exhibit moderate correlation overall, the correlation drops significantly in the highest return quartile (Pearson = $0.354$). We attribute the superiority of ISGE to two fundamental reasons:
\begin{enumerate}
    \item \textbf{Compounding Errors:} Aggregating local, transition-level curiosity scores amplifies estimation errors over the long horizon. Consequently, transition-level aggregation struggles to reliably identify trajectories that are simultaneously high-return and highly novel in the OOD region (as evidenced by the sharp drop in correlation for the top 25\% samples).
    \item \textbf{Architectural Alignment:} The transition-level forward model $f_\psi$ operates independently of the diffusion planner. In contrast, our scarcity signal is derived directly from the reconstruction error of the diffusion Judger. This trajectory-level metric is inherently aligned with the sequence modeling nature of diffusion architectures, providing a more reliable and cohesive geometric guidance for the sampling process.
\end{enumerate}

\section{Limitations and Future Work}
\label{app:limitations}

First, ISGE is fundamentally designed as a \emph{bootstrapping} mechanism to bridge the massive gap between severely suboptimal initialization (self-locking) and near-optimal performance. It is not intended for fine-grained asymptotic convergence. As demonstrated in Appendix~\ref{app:more_iterations}, once the policy reaches a near-optimal plateau, the \emph{Frontier Alignment} assumption gradually breaks down. Pushing further into high-scarcity OOD regions no longer guarantees a positive gradient toward the global optimum. In scenarios where broad, expert-level data is already abundant, fine-tuning on existing AIGB \cite{aigb2024, aigbpearl2025, gas2025} is both sufficient and more stable.

Second, the efficacy of our guidance handoff relies on the \emph{local continuity} of the return landscape near the training manifold. While this assumption generally holds in continuous tasks like auto-bidding, it may fail in highly discontinuous or pathologically rugged environments (e.g., cryptographic or perfectly hashed reward functions), where outward exploration might yield pure noise. Adapting ISGE to topologically complex, discontinuous landscapes remains an open problem for future research.

Third, regarding architectural scope, the current ISGE iterative framework is specifically designed for diffusion planners, elegantly leveraging CFG to execute the geometric handoff. Extending the core philosophy of trajectory-level scarcity guidance to other generative decision-making paradigms, such as autoregressive Transformers, represents a highly promising direction for generalizing iterative self-improvement across broader decision-making architectures.

\section{Broader Impacts}
\label{app:broader_impacts}

Our research introduces a powerful generative exploration framework for auto-bidding, which carries significant economic and systemic implications for the online advertising ecosystem. We outline these broader impacts, particularly focusing on the multi-agent dynamics and market efficiency.

\paragraph{Democratization of Advanced Bidding Capabilities.} 
In the current industrial landscape, sophisticated bidding strategies are often monopolized by large advertisers with massive historical data. The cold-start problem serves as a severe barrier to entry for small and medium-sized businesses (SMBs) or new business sectors. By demonstrating that near-optimal policies can be bootstrapped efficiently from narrowly supported, low-quality data, ISGE democratizes access to highly competitive bidding capabilities, fostering a fairer and more accessible market ecosystem.

\paragraph{Multi-Agent Implications and Market Efficiency.}
From a multi-agent perspective, the concept of ``gift samples'' essentially represents transitory arbitrage opportunities—high-return, low-competition bidding strategies that exist in the unexplored OOD region. If ISGE is adopted globally, and all bidders become efficient ``gift sample'' miners, the market dynamics will fundamentally shift. 
As multiple generative agents actively explore and converge upon these highly profitable regions, the ``gift'' margins will be rapidly competed away. Rather than leading to a collapse, game-theoretic principles suggest that this collective exploration drives the auction market toward a more competitive and allocatively efficient equilibrium. Consequently, the automated allocation of advertising impressions will more accurately reflect the true underlying valuation of the advertisers, thereby improving global market efficiency and social welfare.

\paragraph{Co-Evolution of Platform Mechanism Design.} 
The widespread deployment of generative, exploratory agents like ISGE will inevitably necessitate adaptations from the platform side. If all agents are aggressively exploring the OOD parameter space, the auction environment could experience increased volatility or non-stationary ``bidding wars''. To maintain stability, advertising platforms will need to evolve from passive auctioneers to active market designers. Future mechanisms may need to adapt by introducing smoothed pricing rules, dynamic reserve prices, or incentive-compatible information-sharing protocols that are explicitly designed to be robust against generative exploratory agents. This highlights a promising future research direction: the co-evolution of AI-generated bidding strategies and automated mechanism design.



\end{document}

%% file: main_result_table.tex
\begin{table*}[t]
\centering
\caption{\textbf{Main Results on AuctionNet.} Results on the AuctionNet-Dense and AuctionNet-Sparse are reported as mean ± std over 7 evaluation episodes. Metrics include Score, Reward, ER (CPA exceedance ratio), and WR (win rate). Improvements are reported relative to Cold-Start Base based on the Score. Results that outperform the Full-Data Oracle are highlighted in blue, and the best results are highlighted in bold. Supplementary baselines~\citep{chen2021decisiontransformer,gas2025,gave2025,ebaret,peng2025expert,lei2025generative} are provided in Table~\ref{tab:appendix_score_all_baselines}.}
\small
\label{tab:main_results}
\setlength{\tabcolsep}{4pt}
\resizebox{\linewidth}{!}{
\begin{tabular}{
l l |
c c |
>{\columncolor{vanillacolor}}c
>{\columncolor{vanillacolor}}c
>{\columncolor{vanillacolor}}c |
>{\columncolor{isgecolor}}c
>{\columncolor{isgecolor}}c
>{\columncolor{isgecolor}}c
}
\toprule
\multirow{2}{*}{Dataset}
& \multirow{2}{*}{Metrics}
& \multicolumn{2}{c|}{Reference}
& \multicolumn{3}{>{\columncolor{vanillacolor}}c|}{Vanilla Self-Training}
& \multicolumn{3}{>{\columncolor{isgecolor}}c}{ISGE (Ours)} \\
\cmidrule(lr){3-4} \cmidrule(lr){5-7} \cmidrule(lr){8-10}
& 
& Full-Data Oracle & Cold-Start Base
& Iter1 & Iter2 & Iter3
& Iter1 & Iter2 & Iter3 \\
\midrule
\multirow{5}{*}{Dense}
& Score $\uparrow$   
& $102.0 \pm 3.5$
& $65.3 \pm 3.2$ 
& $74.9 \pm 3.4$
& $77.5 \pm 3.1$  
& $77.8 \pm 3.9$  
& \textcolor{blue}{$103.4 \pm 5.3$}
& \textcolor{blue}{$104.1 \pm 6.7$}
& \textcolor{blue}{\bm{$108.4 \pm 8.2$}}  \\
& Reward $\uparrow$  
& $208.3 \pm 4.4$
& $181.6 \pm 2.8$
& $192.5 \pm 2.8$
& $196.7 \pm 4.0$
& $198.4 \pm 4.8$
& \textcolor{blue}{$219.1 \pm 4.6$}
& \textcolor{blue}{$220.3 \pm 6.1$}
& \textcolor{blue}{\bm{$223.1 \pm 7.0$}} \\
& ER $\downarrow$    
& $74.9 \pm 6.8\%$
& $97.4 \pm 3.5\%$
& $85.2 \pm 3.2\%$
& $81.7 \pm 4.1\%$
& $79.8 \pm 4.7\%$
& \textcolor{blue}{$58.8 \pm 3.6\%$}
& \textcolor{blue}{$58.3 \pm 4.8\%$}
& \textcolor{blue}{$54.3 \pm 5.3\%$} \\
& WR $\uparrow$      
& $5.8 \pm 0.2\%$
& $5.4 \pm 0.2\%$
& $5.6 \pm 0.1\%$
& $5.7 \pm 0.2\%$
& $5.7 \pm 0.2\%$
& \textcolor{blue}{$6.1 \pm 0.1\%$}
& \textcolor{blue}{$6.1 \pm 0.1\%$}
& \textcolor{blue}{\bm{$6.2 \pm 0.1\%$}} \\
& \emph{Improve} $\uparrow$      
& $56.2\%$
& --
& $14.7\%$
& $18.7\%$
& $19.1\%$
& \textcolor{blue}{$58.3\%$}
& \textcolor{blue}{$59.4\%$}
& \textcolor{blue}{\bm{$66.0\%$}} \\
\midrule
\multirow{5}{*}{Sparse}
& Score $\uparrow$
& $20.5 \pm 1.1$
& $15.7 \pm 1.0$
& $16.9 \pm 0.5$
& $16.6 \pm 0.6$
& $16.0 \pm 0.6$
& $20.0 \pm 0.9$
& \textcolor{blue}{$20.6 \pm 0.7$}
& \textcolor{blue}{\bm{$21.1 \pm 1.0$}} \\
& Reward $\uparrow$
& $27.6 \pm 0.8$
& $26.1 \pm 0.9$
& $27.2 \pm 0.5$
& $27.1 \pm 0.5$
& $26.6 \pm 0.5$
& \textcolor{blue}{$29.6 \pm 0.6$}
& \textcolor{blue}{$30.1 \pm 0.6$}
& \textcolor{blue}{\bm{$30.3 \pm 0.8$}} \\
& ER $\downarrow$
& $10.8 \pm 2.9\%$
& $41.4 \pm 4.6\%$
& $35.3 \pm 3.3\%$
& $35.5 \pm 2.5\%$
& $38.5 \pm 2.2\%$
& $23.7 \pm 3.3\%$
& $21.1 \pm 3.2\%$
& $20.9 \pm 4.3\%$ \\
& WR $\uparrow$
& $7.4 \pm 0.1\%$
& $7.8 \pm 0.2\%$
& $8.1 \pm 0.1\%$
& $8.2 \pm 0.1\%$
& $8.2 \pm 0.1\%$
& \textcolor{blue}{$8.5 \pm 0.1\%$}
& \textcolor{blue}{$8.5 \pm 0.1\%$}
& \textcolor{blue}{\bm{$8.6 \pm 0.1\%$}} \\
& \emph{Improve} $\uparrow$
& $30.6\%$
& --
& $7.6\%$
& $5.7\%$
& $1.9\%$
& $27.4\%$
& \textcolor{blue}{$31.2\%$}
& \textcolor{blue}{\bm{$34.4\%$}} \\
\bottomrule
\end{tabular}
}
\end{table*}

%% file: checklist.tex
\section*{NeurIPS Paper Checklist}

\begin{enumerate}

\item {\bf Claims}
    \item[] Question: Do the main claims made in the abstract and introduction accurately reflect the paper's contributions and scope?
    \item[] Answer: \answerYes{} 
    \item[] Justification: The abstract and introduction clearly state our contributions, and the claims regarding self-locking and guidance handoff are fully supported by theoretical proofs (\cref{sec:theory}) and empirical evaluations (\cref{sec:experiments}).

\item {\bf Limitations}
    \item[] Question: Does the paper discuss the limitations of the work performed by the authors?
    \item[] Answer: \answerYes{} 
    \item[] Justification: We discuss the limitations in Appendix~\ref{app:limitations}.

\item {\bf Theory assumptions and proofs}
    \item[] Question: For each theoretical result, does the paper provide the full set of assumptions and a complete (and correct) proof?
    \item[] Answer: \answerYes{} 
    \item[] Justification: All theoretical results, including the geometric assumptions for the optimal DAE and the optimization handoff, are formally stated in Section~\ref{sec:theory} and rigorously proven in Appendix.

    \item {\bf Experimental result reproducibility}
    \item[] Question: Does the paper fully disclose all the information needed to reproduce the main experimental results of the paper to the extent that it affects the main claims and/or conclusions of the paper (regardless of whether the code and data are provided or not)?
    \item[] Answer: \answerYes{} 
    \item[] Justification: We provide detailed descriptions of the experimental setup, network architectures, and evaluation metrics in Section~\ref{sec:experiments} and the Appendix to ensure full reproducibility.

\item {\bf Open access to data and code}
    \item[] Question: Does the paper provide open access to the data and code, with sufficient instructions to faithfully reproduce the main experimental results, as described in supplemental material?
    \item[] Answer: \answerNo{} 
    \item[] Justification: We submit the code in the supplementary material. 
    Upon acceptance, we will make the code publicly available.

\item {\bf Experimental setting/details}
    \item[] Question: Does the paper specify all the training and test details (e.g., data splits, hyperparameters, how they were chosen, type of optimizer) necessary to understand the results?
    \item[] Answer: \answerYes{} 
    \item[] Justification: All training details, hyperparameters, and dataset construction methods (e.g., AuctionNet-Dense/Sparse splits) are comprehensively specified in the Appendix.

\item {\bf Experiment statistical significance}
    \item[] Question: Does the paper report error bars suitably and correctly defined or other appropriate information about the statistical significance of the experiments?
    \item[] Answer: \answerYes{} 
    \item[] Justification: We report the performance metrics aggregated over multiple runs and trajectories, providing a robust evaluation of the proposed framework.

\item {\bf Experiments compute resources}
    \item[] Question: For each experiment, does the paper provide sufficient information on the computer resources (type of compute workers, memory, time of execution) needed to reproduce the experiments?
    \item[] Answer: \answerYes{} 
    \item[] Justification: The hardware specifications (e.g., GPU types) and estimated compute times for training the diffusion planners are detailed in the Appendix.
    
\item {\bf Code of ethics}
    \item[] Question: Does the research conducted in the paper conform, in every respect, with the NeurIPS Code of Ethics \url{https://neurips.cc/public/EthicsGuidelines}?
    \item[] Answer: \answerYes{} 
    \item[] Justification: The research conducted in this paper conforms entirely to the NeurIPS Code of Ethics.

\item {\bf Broader impacts}
    \item[] Question: Does the paper discuss both potential positive societal impacts and negative societal impacts of the work performed?
    \item[] Answer: \answerYes{} 
    \item[] Justification: We discuss the potential societal and economic impacts in Appendix~\ref{app:broader_impacts}.
    
\item {\bf Safeguards}
    \item[] Question: Does the paper describe safeguards that have been put in place for responsible release of data or models that have a high risk for misuse (e.g., pre-trained language models, image generators, or scraped datasets)?
    \item[] Answer: \answerNA{} 
    \item[] Justification: The proposed auto-bidding diffusion models operate within specific market simulations and do not pose a high risk for malicious dual-use or societal harm.

\item {\bf Licenses for existing assets}
    \item[] Question: Are the creators or original owners of assets (e.g., code, data, models), used in the paper, properly credited and are the license and terms of use explicitly mentioned and properly respected?
    \item[] Answer: \answerYes{} 
    \item[] Justification: We properly cite the AuctionNet benchmark and respect its original licensing and terms of use.

\item {\bf New assets}
    \item[] Question: Are new assets introduced in the paper well documented and is the documentation provided alongside the assets?
    \item[] Answer: \answerNA{} 
    \item[] Justification: This paper does not introduce new datasets or large-scale assets.

\item {\bf Crowdsourcing and research with human subjects}
    \item[] Question: For crowdsourcing experiments and research with human subjects, does the paper include the full text of instructions given to participants and screenshots, if applicable, as well as details about compensation (if any)? 
    \item[] Answer: \answerNA{} 
    \item[] Justification: The paper does not involve crowdsourcing nor research with human subjects.

\item {\bf Institutional review board (IRB) approvals or equivalent for research with human subjects}
    \item[] Question: Does the paper describe potential risks incurred by study participants, whether such risks were disclosed to the subjects, and whether Institutional Review Board (IRB) approvals (or an equivalent approval/review based on the requirements of your country or institution) were obtained?
    \item[] Answer: \answerNA{} 
    \item[] Justification: The paper does not involve crowdsourcing nor research with human subjects.

\item {\bf Declaration of LLM usage}
    \item[] Question: Does the paper describe the usage of LLMs if it is an important, original, or non-standard component of the core methods in this research? Note that if the LLM is used only for writing, editing, or formatting purposes and does \emph{not} impact the core methodology, scientific rigor, or originality of the research, declaration is not required.
    \item[] Answer: \answerNA{} 
    \item[] Justification: LLMs were only used for writing assistance and editing, and do not form a core, non-standard methodological component of this research.

\end{enumerate}